\documentclass[manuscript,screen]{acmart}

\usepackage{todonotes}
\usepackage{caption}
\usepackage{subcaption}
\usepackage{multicol}
\usepackage{multirow}
\usepackage{comment}

\usepackage{pifont}
\usepackage{newunicodechar}
\newunicodechar{✓}{\ding{51}}
\newunicodechar{✗}{\ding{55}}

\usepackage{listings}
\usepackage[most]{tcolorbox}

\usepackage{array}
\usepackage{makecell}

\AtBeginDocument{%
  }

\begin{document}

\title{Adversarial Robustness in Unsupervised Machine Learning: A Systematic Review}

\author{Mathias Lundteigen Mohus}
\email{mathias.l.mohus@ntnu.no}
\orcid{0000-0002-1105-6515}
\author{Jingyue Li}
\email{jingyue.li@ntnu.no}
\orcid{0000-0002-7958-391X}
\affiliation{%
  \institution{NTNU Department of Computer Science}
  \streetaddress{Sem Sælandsvei 9}
  \city{Trondheim}
  \state{Trøndelag}
  \country{Norway}
  \postcode{NO-7491}
}

\renewcommand{\shortauthors}{Mohus et. al.}

\begin{abstract}
As the adoption of machine learning models increases, ensuring robust models against adversarial attacks is increasingly important. With unsupervised machine learning gaining more attention, ensuring it is robust against attacks is vital. This paper conducts a systematic literature review on the robustness of unsupervised learning, collecting 86 papers. Our results show that most research focuses on privacy attacks, which have effective defenses; however, many attacks lack effective and general defensive measures. Based on the results, we formulate a model on the properties of an attack on unsupervised learning, contributing to future research by providing a model to use.
\end{abstract}

\ccsdesc[500]{Computing methodologies~Unsupervised learning}
\ccsdesc[500]{Security and privacy~Software security engineering}

\keywords{unsupervised learning, machine learning, adversarial robustness, adversarial attack, systematic literature review}


\maketitle

\section{Introduction}
As technologies mature, hardware capabilities increase, and legislation is introduced, adopting machine learning technologies in various applications seems inevitable, with the technology generally divided into three types: Supervised Learning (SL), Reinforcement Learning (RL), and Unsupervised Learning. SL focuses on approaches where the training data is labeled, i.e., a data sample maps to a correct value. A variety of cases use this, e.g., object detection \cite{ali2005supervised}, image classification \cite{pena2014object}, and voice recognition \cite{ali2021voice}. RL is an approach where the model trains in the working environment, rewarding or punishing the model based on its performance. Use cases for this are bug detection in software \cite{rani2023deep}, computing resource optimization \cite{shi2020reinforcement}, and control of cyber-physical systems \cite{liu2019reinforcement}. UL performs training of its models without guidance; pure data are used as the basis for training. Use cases for this is data generation \cite{devaranjan2020meta}, e.g., \textit{GANs}; encoding/decoding \cite{srivastava2015unsupervised}, e.g., \textit{autoencoders} , and data clustering \cite{chander2021unsupervised}.

By its lack of need for guided information, UL can enable more data to be usable. In a world that generates a large amount of data (around 2.5 quintillion bytes each day \cite{bytesnum}), methods for using unsupervised data would be invaluable. However, where technology is used, adversaries inevitably attempt to exploit vulnerabilities in these technologies. With the increased use of ML in various fields, it is crucial to explore the adversarial robustness of ML models, particularly the use of UL, as this field has received less focus than supervised and reinforcement learning.

\citet{liu2018survey} summarized attacks on machine learning models broadly, while \citet{guan2018machine} cover machine learning within cybersecurity. \citet{zhang2020privacy} outline privacy vulnerabilities in machine learning, which could reveal sensitive information used to train synthetic data generators in the context of UL. At the same time, \citet{martins2020adversarial}cover \textit{adversarial examples} relating to detection techniques, where the input to an ML model is perturbed in a specific way in order to disrupt the functionality of the model.

In contrast with \citet{liu2018survey}, \citet{guan2018machine} and \citet{zhang2020privacy}, where parts of UL adversarial robustness are covered, we aim to conduct a Systematic Literature Review (SLR) on the topic of adversarial robustness within the whole field of unsupervised machine learning. The papers focus on answering the three research questions below:

\begin{itemize}
    \item \textbf{RQ1}: What types of attacks exist against UL-based systems, and what are the common attributes of the attacks? 
    \item \textbf{RQ2}: What defensive countermeasures exist for attacks against UL-based systems? 
    \item \textbf{RQ3}: What are the challenges to defending against possible attacks targeting UL-based systems? 
\end{itemize}

The SLR follows the guidelines outlined by \citet{keele2007guidelines} and identified 86 papers relevant to adversarial robustness in unsupervised learning. Thematic analysis was then used on these papers to summarize the papers according to the RQs systematically. 

The results of SLR reveal that:
\begin{itemize}
    \item \textit{GANs} are mostly targeted by \textit{privacy} attacks.
    \item \textit{AutoEncoders} are mostly targeted by \textit{adversarial examples}.
    \item \textit{Poisoning} attacks are used against most types of UL.
    \item Most attacks utilize \textit{distance metrics} in limiting how much the attack alters samples for the attack.
    \item Several defensive methods that work in a supervised or reinforcement context are not applicable in an unsupervised context. For example, \textit{fine pruning} and \textit{output activation clustering} are not suitable defenses for generative models.
    \item The adversary's knowledge of the target system greatly affects the attack's performance.
    \item There is a noticeable lack of defensive methods used in UL, except for \textit{differential privacy}.
\end{itemize}

The contributions of the study are:
\begin{itemize}
    \item We provide a complete and novel summary of the state-of-the-art adversarial robustness of unsupervised learning relating to attributes of the attack, target, and defense.
    \item We have summarized and outlined the main challenges facing the adversarial robustness of unsupervised learning.
    \item We create a model of the attributes associated with attacks against unsupervised machine learning to analyze and compare adversarial attacks against UL.
\end{itemize}

The rest of the paper is organized as follows. Section 2 covers the basics of unsupervised learning. Section 3 covers surveys related to the security and robustness of machine learning. Section 4 explains our research method. Sections 5 to 7 present the results. Section 8 discusses the results. Section 9 concludes and proposes future work.

\section{Background}

Within UL, two types of technologies generally exist: discriminatory and generative. \textit{Discriminatory} covers models where data are input into the model, producing a specific output based on its training. Examples of these are \textit{clustering} \cite{grira2004unsupervised, hahne2008unsupervised}, \textit{AutoEncoders} \cite{bank2020autoencoders, pinaya2020autoencoders} and \textit{super-resolution networks} \cite{anwar2020deep}. 

\begin{itemize}
    \item \textit{Clustering} refers to a broad family of methods that defines clusters of data points based on certain heuristics. Unsupervised clustering finds these clusters based on distance metrics between data points to find which data points are mathematically close enough to belong to the same cluster. Clustering is often used for data analysis in data-heavy fields, such as gene information \cite{sumedha2008unsupervised}.

    \item \textit{AutoEncoders} is a method for training a neural network to create an inner representation of the training data, since the model trains to output the same data as the input. In most cases, AutoEncoders force a dimensionality reduction on the inner representation by limiting the number of hidden nodes in the network, where the first part of the network training is to create the encoding, and the last part of the network training is to decode the representation into the original data. One application of AutoEncoders is denoising data \cite{bank2020autoencoders} where data is passed through the AutoEncoder to remove noise. 

    \item \textit{Super-resolution networks} are neural networks that work to increase the dimensionality of the data, by training the model to output a higher-dimensional piece of data, based on a lower-dimensional piece of data, e.g. a low-resolution image. Super-resolution is often used in image technologies to upscale the resolution of an image \cite{anwar2020deep}.
\end{itemize}

\textit{Generative} models cover models which do not directly use input data but are used to create new data with properties tied to the properties of the training set. Examples of this are \textit{Variational AutoEncoders} \cite{kingma2019introduction} and \textit{Generative Adversarial Networks} \cite{creswell2018generative, wang2017generative, goodfellow2020generative}. 

\begin{itemize}
    \item \textit{Variational AutoEncoder} is a method to use AutoEncoder to generate data by forcing constraints on the inner representation of the AutoEncoder, in turn making it possible to input random noise into the decoder part of the AutoEncoder and producing an output that is statistically similar to the training set used to train the AutoEncoder. This model is one of the main components of generating \textit{deepfake} images by supplanting some of the encoded features of an image with another, generating an image with features from both images \citet{rocco2021deepfake}.
    
    \item \textit{Generative adversarial networks} consist of two models, the generator and the discriminator. The generator is a neural network with random noise as input, while the discriminator is a classifier trained to differentiate between the output from the generator and the training data. The generator and the discriminator are, in turn, trained against each other based on the performance of the discriminator. GANs are used for various synthetic data generation, e.g., medical data generation \cite{han2018gan}. 
\end{itemize}

\section{Related Work}
As the field of machine learning security has matured, several papers have been published that survey the state of the art within both broad and narrow fields within ML security. 

\citet{liu2018survey} use the taxonomy from \citet{fredrikson2015model} to analyze 42 papers on attacks against machine learning models. The survey group the papers based on 1) The targeted algorithm, with the types being: Naïve Bayes/Logistic Regression/Decision Tree, Support Vector Machines (SVM), Principal Component Analysis (PCA)/Least Absolute Shrinkage and Selection Operator (LASSO), Clustering, and Deep Neural Network (DNN), and 2) The type of attack, with the types being: Poisoning Attack, Evasion Attack, Impersonation Attack, and Inversion Attack. \citet{liu2018survey} also focus explicitly on comparing types of Adversarial Examples attacks, comparing the techniques, advantages, disadvantages, and formations of the adversarial examples. Additionally, \citet{liu2018survey} summarize defensive techniques for ML, outlining how its surveyed papers assess the security of their ML models and which defensive mechanisms are put in place: proactive or reactive. Defenses in training are also discussed, such as Data Sanitation, Bootstrap Aggregating, Random Subspace Method, and designing more robust algorithms. For defenses in the testing/inferring phase, \citet{liu2018survey} state that “countermeasures in the testing/inferring phase mainly focus on the improvement of learning algorithms’ robustness,” describing techniques like Game Theory, Adversarial Training, Defensive Distillation, Dimension Reduction, as well as ensemble methods. Lastly, \citet{liu2018survey} also cover the privacy of data used in ML, including Differential Privacy and Homomorphic Encryption. 

\citet{martins2020adversarial} perform a survey on the use of adversarial examples targeted towards detection techniques, particularly malware (11 papers) and intrusion detection (nine papers), surveyed from mainly IEEE Xplore, Springer, arXiv, ScienceDirect, Research Gate, and Google, using the keywords “Intrusion Detection,” “Malware Detection” and “Adversarial Machine Learning.” The paper compares several features recurring in its literature: Attack Techniques, Classifiers, Datasets, Metrics, and Results and Conclusions. \citet{martins2020adversarial} also cover defensive techniques against adversarial examples: Adversarial Training, Gradient Masking, Defensive Distillation, Feature Squeezing, Universal Perturbation Defense Method, and MagNet. 

\citet{guan2018machine} cover two related topics: ML’s impact on cybersecurity and AI systems’ robustness, presenting 14 papers. The paper labels attributes covering the Research Direction, whether AI for cybersecurity or security of AI, points of innovation in the paper, and a summary of the idea for each paper. The paper also synthesizes common threats and solutions in AI security, such as Data Sanitation and Learning Algorithm Improvement being the solution to Poisoning attacks, Adversarial Training and Output Smoothing being solutions to Evasion and Impersonation Attacks, and Differential Privacy and Homomorphic Encryption being the solution to Inversion Attacks. 

\citet{zhang2020privacy} present a survey on privacy threats and mitigations within ML. The paper outlines a taxonomy for attacks, covering the Stage (where the attack happens), Goal, Knowledge, Capability, and Strategy of the attack. The authors summarize privacy threats against models as Parameter/Function, Hyperparameter, and Architecture, with defensive measures being Confidence Perturbation and Malicious Query Behaviour Detection. Privacy threats against data are summarized as Data Memorization attacks, Model Inversion Attacks, and Membership Inference attacks, with defensive measures being Hierarchical Model Stacking Techniques, Attribute Correlation with Adversarial Perturbations, Meta-Classifier, Collaborative Learning, Homomorphic Encryption, and Differential Privacy.

Table \ref{tab:comparison_gap} outlines the contribution of the existing surveys. Only one survey covers GAN \cite{zhang2020privacy}, two surveys cover clustering \cite{liu2018survey, martins2020adversarial}, and none covers attacks on AutoEndoers. The systematic description of attack goals is also only present in two papers \cite{liu2018survey, guan2018machine}. There is also a need for systematic metrics descriptions to determine attack success, with only one paper  \cite{martins2020adversarial} summarizing the metrics. Two papers cover the attacker’s knowledge, while the other two somewhat systematically cover the knowledge. Additionally, three papers cover attacks in training and production systems, while \citet{martins2020adversarial} do not cover attacks in training scenarios. Finally, all the papers cover defensive measures against attacks. 

\begin{table}[ht]
    \centering
    \caption{Comparison between the four surveys and this paper regarding categories covered and whether the survey is performed systematically. ✓ is when the paper covers the category, ✗ is when the paper does not cover the category, and * is when the category is partly covered by the paper}
    \label{tab:comparison_gap}
    \begin{tabular}{|p{0.04\textwidth}|p{0.34\textwidth}|p{0.03\textwidth}|p{0.03\textwidth}|p{0.03\textwidth}|p{0.03\textwidth}|p{0.03\textwidth}|p{0.03\textwidth}|p{0.03\textwidth}|p{0.03\textwidth}|p{0.03\textwidth}|p{0.03\textwidth}|p{0.03\textwidth}|p{0.03\textwidth}|}
    \hline
    \multirow{2}{*}{Ref.} & \multirow{2}{*}{Short Description} & \multirow{2}{*}{\rotatebox[origin=r]{90}{Systematic}} & \multirow{2}{*}{\rotatebox[origin=r]{90}{Attack Goal}} & \multirow{2}{*}{\rotatebox[origin=r]{90}{Attack Knowledge}} \newline \newline & \multicolumn{2}{c|}{Attack Focus} & \multirow{2}{*}{\rotatebox[origin=r]{90}{Metrics}} & \multicolumn{5}{c|}{Target Types} & \multirow{2}{*}{\rotatebox[origin=r]{90}{Defenses}} \\ \cline{6-7} \cline{9-13}
    & & & & & \rotatebox[origin=r]{90}{ Production} & \rotatebox[origin=r]{90}{ Training} & & \rotatebox[origin=r]{90}{GAN} & \rotatebox[origin=r]{90}{AutoEncoder} \newline & \rotatebox[origin=r]{90}{Clustering} & \rotatebox[origin=r]{90}{Classification} & \rotatebox[origin=r]{90}{Super-resolution} & \\ \hline
    \cite{liu2018survey} & Adversarial Examples attacks against Machine Learning & ✗ & ✓ & * & \multicolumn{1}{l|}{✓} & ✓ & ✗ & \multicolumn{1}{l|}{✗} & \multicolumn{1}{l|}{✗} & ✓ & ✗ & ✗ & ✓ \\ \hline
    \cite{martins2020adversarial} & Adversarial Examples against detection techniques & * & ✗ & ✓ & \multicolumn{1}{l|}{✓} & ✗ & ✓ & \multicolumn{1}{l|}{✗} & \multicolumn{1}{l|}{✗} & ✓ & ✗ & ✗ & ✓ \\ \hline
    \cite{guan2018machine} & Impact of ML on security, and robustness of AI systems & ✗ & ✓ & * & \multicolumn{1}{l|}{✓} & ✓ & ✗ & \multicolumn{1}{l|}{✗} & \multicolumn{1}{l|}{✗} & ✗ & ✗ & ✗ & ✓ \\ \hline
    \cite{zhang2020privacy} & Privacy threats and mitigations within ML & ✗ & ✗ & ✓ & \multicolumn{1}{l|}{✓} & ✓ & ✗ & \multicolumn{1}{l|}{✓} & \multicolumn{1}{l|}{✗} & ✗  & ✗ & ✗ & ✓ \\ \hline
    This work & A systematic review on adversarial robustness of unsupervised machine learning & ✓ & ✓ & ✓ & \multicolumn{1}{l|}{✓} & ✓ & ✓ & \multicolumn{1}{l|}{✓} & \multicolumn{1}{l|}{✓} & ✓  & ✓ & ✓ & ✓ \\ \hline
    \end{tabular}
    
\end{table}

In summary, there is an evident lack of a systematic and comprehensive survey focusing on attacks targeting all kinds of ULs and corresponding defenses. Considering the potential importance of unsupervised learning in environments with a large amount of unlabeled data, a survey to cover the adversarial robustness of unsupervised machine learning would provide the industry with insight into the shortcomings of UL technologies and academia with knowledge on the state-of-the-art attacks and defenses that are covered in the field of adversarial robustness within UL.

\section{Research Design and Implementation}

\subsection{Systematic Literature Review outline}

In computer science, \citet{keele2007guidelines} show one framework for conducting an SLR within software engineering and related fields. This SLR uses the method outlined in \citet{keele2007guidelines} as a base, which includes 1) Creating the search query based on inclusion criteria, 2) Filtering based on exclusion criteria, and 3) Validating the quality of papers. 

\subsection{Creating the search query}
Based on the research questions and research focus, we generate the search query using a modified version of PICOC (Population, Intervention, Comparison, Outcome, Context) from \citet{petticrew2008systematic}, without the \textit{Comparison} section, as it was considered not applicable to the context of this SLR. While the PIOC framework groupings guide the query’s creation, the specific interpretation in \citet{keele2007guidelines} do not map perfectly to the chosen tokens, as described below. 

\begin{itemize}
    \item \textit{Population:} Relating to the security of unsupervised machine learning, the population will cover general areas of technologies that are related to or specifically UL. 
    \item \textit{Intervention:} The intervention relates here to general descriptions of how an attacker might exploit vulnerabilities in UL systems. 
    \item \textit{Outcome:} Outcome relates to the desired data or information that a piece of research produces through their papers, specifically finding the vulnerabilities and defensive methods of UL methods.  
    \item \textit{Context:} The context relates to the context in which the research is explored, specifically which part the UL is considered, e.g., the technology versus human error.
\end{itemize}

As \citet{keele2007guidelines} do not outline a framework for evaluating the quality of the search query, other than following the PICOC framework, while \citet{denyer2009sage} only state “it is often wise to spend significant time on constructing the search strings, for this investment can improve the efficiency of the search”, we formulate here a method for evaluating the query. First, a list of 13 papers relating to security in UL was created, based on an ad-hoc search, in addition to previous knowledge of specific papers. We then search the IEEE Xplore, ACM Digital Library, arXiv, and Scopus databases using the search query and note how many of the papers in the ad-hoc list are present in the resulting list. While the ad-hoc nature of the list is not ideal, we consider this method better than the alternative, i.e., not having a baseline of comparison, as we know the papers in the ad-hoc list should ideally be present in the resulting list. The second consideration metric was the number of papers produced by the search query. For this, we would only accept queries that produced between 1000 and 10 000 papers to ensure enough relevant papers were included and ensure the following filtration process would not take too long. The results of the query testing are summarized in Table \ref{tab:eval_queries}. Rounds 1 and 2 were good examples of casting a too-narrow and too-wide net, highlighted by the matching percentage to the ad-hoc list and the total number of papers being very low and very high, respectively. Round 3 produced the correct number of papers but was considered not precise enough, with a matching score of 15\%. Round 4 also produced the correct number of papers, increased the matching rate to the ad-hoc list, and was therefore chosen as the query for the SLR. 


\begin{table}
  \caption{Evaluation of different queries}
  \label{tab:eval_queries}
  \begin{tabular}{c|c|c|c|c}
    \toprule
    Search query evaluation round&Peer-reviewed&Pre-published&Total&Match with ad-hoc list\\
    \midrule
    1 & 529 & 39 & 568 & 8\%\\
    2 & 120 241 & 763 & 121 004 & \textbf{85\%}\\
    3 & 4623 & 26 & \textbf{4649} & 15\%\\
    4 & 2647 & 57 & \textbf{2704} & 46\%\\
  \bottomrule
\end{tabular}
\end{table}

The resulting query based on the PIOC categories produces the following search query, modified to fit each database search engine: \\
    
(“Clustering” OR “cluster” OR “AutoEncoder” OR “GAN” OR “Generative Adversarial Network” OR “Unsupervised Learning” OR “UL”) AND (“vulnerability” OR “exploitation” OR “Poisoning” OR “extraction” OR “obfuscation” OR “Adversarial attack” OR “cyber-attack” OR “cyber attack”) AND (“Increase robustness” OR “defend against” OR “increase defence” OR “Mitigate attack” OR “Attack success” OR “attack against” OR “attack vector”) AND (“model” OR “models” OR “system” OR “systems”) 

\subsection{Paper filtering}
The initial list of papers was generated using the query in IEEE Xplore, ACM Digital Library, arXiv, and Scopus. Including unpublished papers in this review follows from the recommendation from \citet{denyer2009sage} “extensive searches are conducted to include both published and unpublished studies.” The initial list was imported into EndNote, automatically removing any duplicates. We used the inclusion criterion for filtering the papers: the paper must cover attacks against UL technologies. We show the exclusion criteria in Table \ref{tab:exclusion_criteria}. We specify criteria E3 to 1) avoid including outdated research and 2) limit the number of papers that would have to be filtered.

\begin{table}[ht]
  \caption{The exclusion criteria for the systematic review}
  \label{tab:exclusion_criteria}
  \begin{tabular}{p{0.2\textwidth}|p{0.6\textwidth}}
    \toprule
    Index & Criteria \\
    \midrule
    E1 & The paper covers attack against general technologies which is not currently used in an unsupervised manner \\
    E2 & Not written in English \\
    E3 & Written before 2011 \\
    E4 & Paper is not available \\
    E5 & Not relevant to UL \\
  \bottomrule
\end{tabular}
\end{table}

\begin{figure}[ht]
  \centering
  \includegraphics[width=0.5\linewidth]{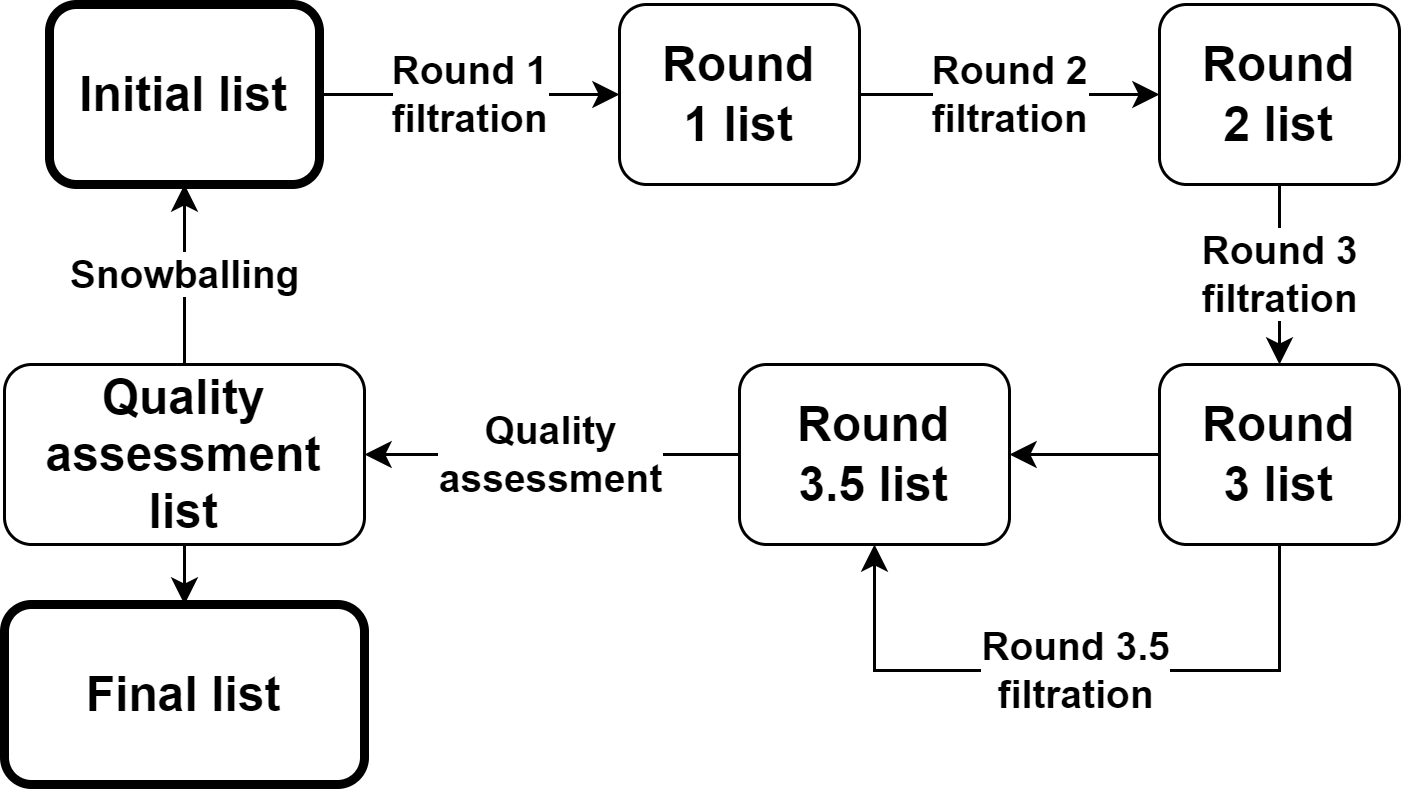}
  \caption{Workflow for the filtration rounds and quality assessment}
  \Description{Outline of the steps taken to produce the final list of papers from the initial results from the query. Filtration rounds 1, 2, 3, and 3.5, and quality assessment, after which two rounds of forward and backward snowballing produced another list of papers to be filtered.}
  \label{fig:slr_workflow}
\end{figure}

We performed several rounds of filtering. The first round focused on removing irrelevant papers from the list. In this round, the papers were considered only based on the title and evaluated on whether 1) It does not relate to security, 2) It does not relate to UL, or 3) The title singularly refers to supervised or reinforcement learning. 
The second round focused on further limiting the number of irrelevant papers from the list. The evaluation criteria remain the same as for filtration round 1. Still, we give the evaluation for each paper more time, and we document each decision to filter in or out. 
The third round limits the number of irrelevant papers even further, where we use the same criteria as filtration rounds 1 and 2, use more time for each evaluation and consider the paper’s abstract. Here, the evaluator can also consider a paper in its entirety to confidently ascertain whether it is relevant or not, according to the criteria referred to here as round 3.5. 
The workflow for the filtration steps and quality assessment is outlined in Figure \ref{fig:slr_workflow}.

\subsection{Quality Assessment}
To ensure the papers contributing to the SLR have a certain quality, \citet{keele2007guidelines} also outline the importance of quality assessment (QA). Our QA consists of 11 questions based on the approach presented by \citet{keele2007guidelines}, scoring 0, 0.5, or 1 point per question. For any paper scoring lower than 5.5 points, we exclude the paper. Additionally, to ensure internal validity, a random sample of papers will be analyzed using the Cohen Kappa statistic \cite{warrens2015five}, measured between the primary researcher doing the QA, and a separate researcher, to ensure the QA scores are accurately applied. The questions used for the QA are in Appendix \ref{sec:slr_details}.

\subsection{Snowballing}
After the initial round of search, filtration, and QA, we also perform two backward and forward snowballing rounds to ensure we include as many relevant papers as possible in this SLR. For the first round of snowballing, the final list of papers after the QA is used as a base, and for the second round, we use the list of papers after QA from the first round of snowballing. \textit{Backward snowballing} (BS) is performed by compiling a list of all the references used in each paper. In contrast, \textit{forward snowballing} (FS) is performed using \textit{Semantic Scholar} to find papers referencing each paper. After one round of initial search and two rounds of forward and backward snowballing, we combine the papers into a single list of papers to be analyzed.

\subsection{Identified papers}
For each round of the systematic literature review, the number of papers for consideration is summarized in Table \ref{tab:summary_slr}. After the snowballing steps, we further reduced the list of 114 papers to 86 by finding duplicates, updated versions, or published duplicates of unpublished papers, with the last forward snowballing round performed in August 2022. 

\begin{table}[ht]
  \caption{Summary of the number of papers after each round of filtration and quality assurance. We base each snowballing round on the list of papers from the quality assurance in the previous round.}
  \label{tab:summary_slr}
  \begin{tabular}{p{0.25\textwidth}|p{0.06\textwidth}|p{0.12\textwidth}|p{0.12\textwidth}|p{0.12\textwidth}|p{0.12\textwidth}||p{0.05\textwidth}}
    \toprule
    &Initial Round & Forward Snowballing 1 & Backward Snowballing 1 & Forward Snowballing 2 & Backward Snowballing 2 & Sum\\
    \midrule
    Initial List & 5825 & 996 & 365 & 683 & 1260 & 9129\\
    Filtration 1 & 1002 & 651 & 160 & 95 & 171 & 2079\\
    Filtration 2 & 478 & 330 & 75 & 53 & 36 & 972\\ 
    Filtration 3 & 28 & 54 & 15 & 37 & 11 & 145\\
    After Quality Assurance & 24 & 46 & 11 & 27 & 6 & 114\\
  \bottomrule
\end{tabular}
\end{table}

\subsection{Data analysis and synthesis}
Thematic analysis (TA) is a widely used method for analyzing qualitative data \cite{braun2006using}, as it “offers an accessible and theoretically flexible approach to analyzing qualitative data.” Performing a TA consists of coding sections of the qualitative data based on its content and categorizing the codes into different themes, which in turn is used for describing and comparing data. Two main ways of performing the TA are either by grounding it in the content or using predetermined codes. In grounding the TA in the content, the codes and themes emerge from the data itself instead of being based on a theory of which codes should exist \cite{walker2006grounded}. The grounded codes can, in turn, be used for creating hypotheses or theories based on the data. Using predetermined codes revolves around codes and themes emerging from predetermined assumptions, hypotheses, and knowledge, on the subject, and the codes and themes used within this for research based on these preconceptions. Grounded and predetermined codes can be combined into a hybrid process, with specific predetermined codes and themes complemented by grounded codes to generate the finished set of codes and themes.

\begin{figure}[ht]
  \centering
  \includegraphics[width=0.5\linewidth]{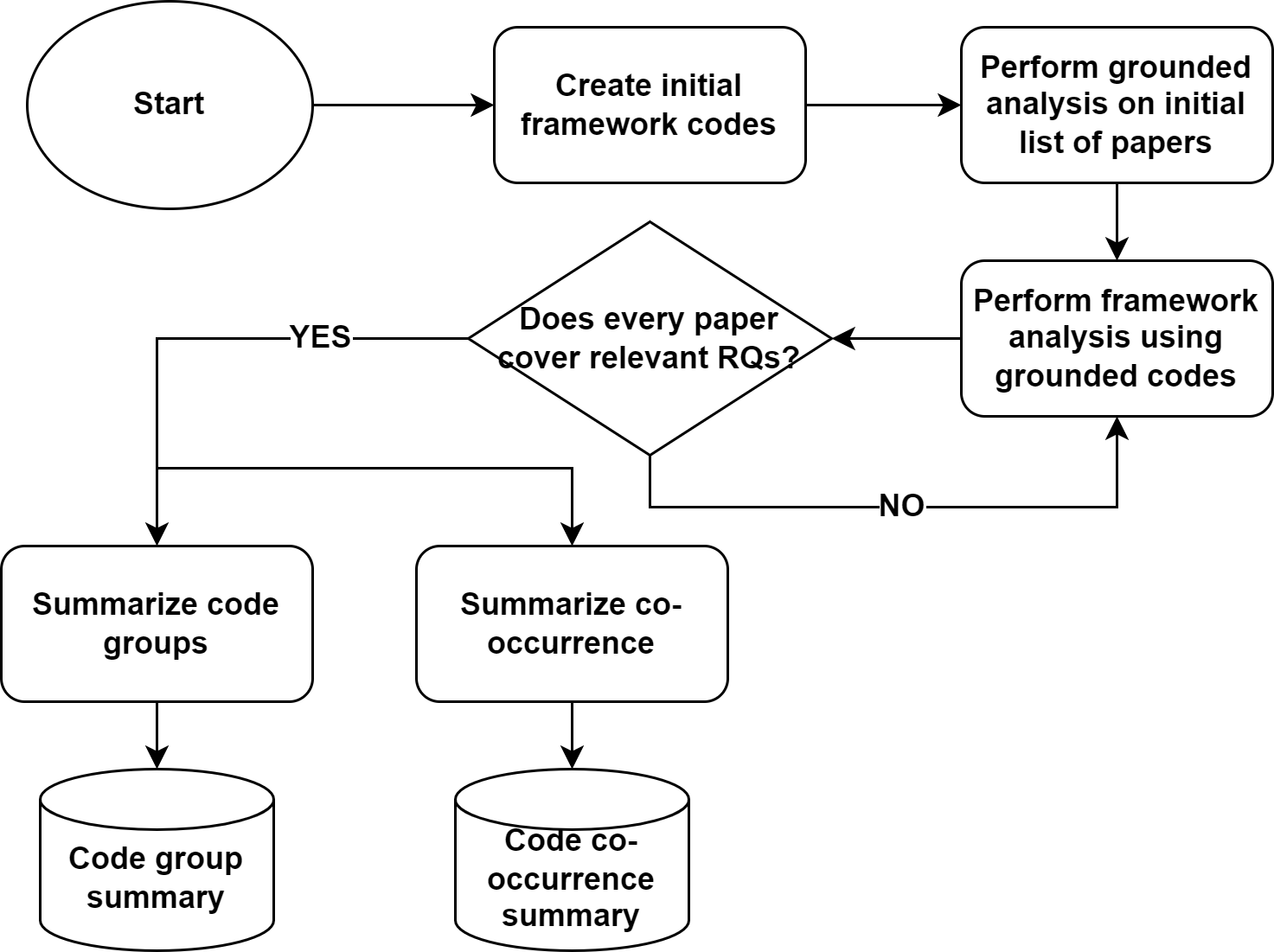}
  \caption{Workflow for the thematic analysis of the produced papers}
  \Description{Outline of the stages in the thematic analysis 1) Creating framework codes, 2) Grounded analysis on the initial list of papers, 3) Framework analysis on the rest of papers using the grounded codes, 4) Summarize code groups and co-occurrence.}
  \label{fig:ta_workflow}
\end{figure}

The workflow for our applied thematic analysis is as described in Figure \ref{fig:ta_workflow}. The predetermined list of codes relating to the RQs,  derived from our initial understanding of UL and its security risks and defenses, are as follows:
\begin{multicols}{4}
\begin{itemize}
    \item Clustering
    \item GAN
    \item AutoEncoder 
    \item Training 
    \item Production  
    \item Model
    \item Confidentiality  
    \item Privacy attacks 
    \item Integrity 
    \item Poisoning 
    \item Adversarial \newline Examples 
    \item Disruption 
    \item Privacy 
    \item Self-driving cars 
    \item Intrusion detection 
\end{itemize}
\end{multicols}

For the analysis, we also split the coding into two sections, \textit{grounded} and \textit{predetermined} coding. The \textit{grounded} coding covers the initial set of papers and is used to establish the primary set of grounded codes. Here, we code each paper thoroughly and in-depth. The \textit{predetermined} coding is performed on the two rounds of snowballing papers and is used to cover the general content of the papers using the established codes. For each paper, the codes relating to the RQs are guaranteed to be coded in this process. While this produces less detail because of the scope of the paper focusing on the big picture of robustness in UL, we consider this a worthy tradeoff to speed up the process of coding all the papers. 

\begin{figure}
    \centering
    \includegraphics[width=0.9\textwidth]{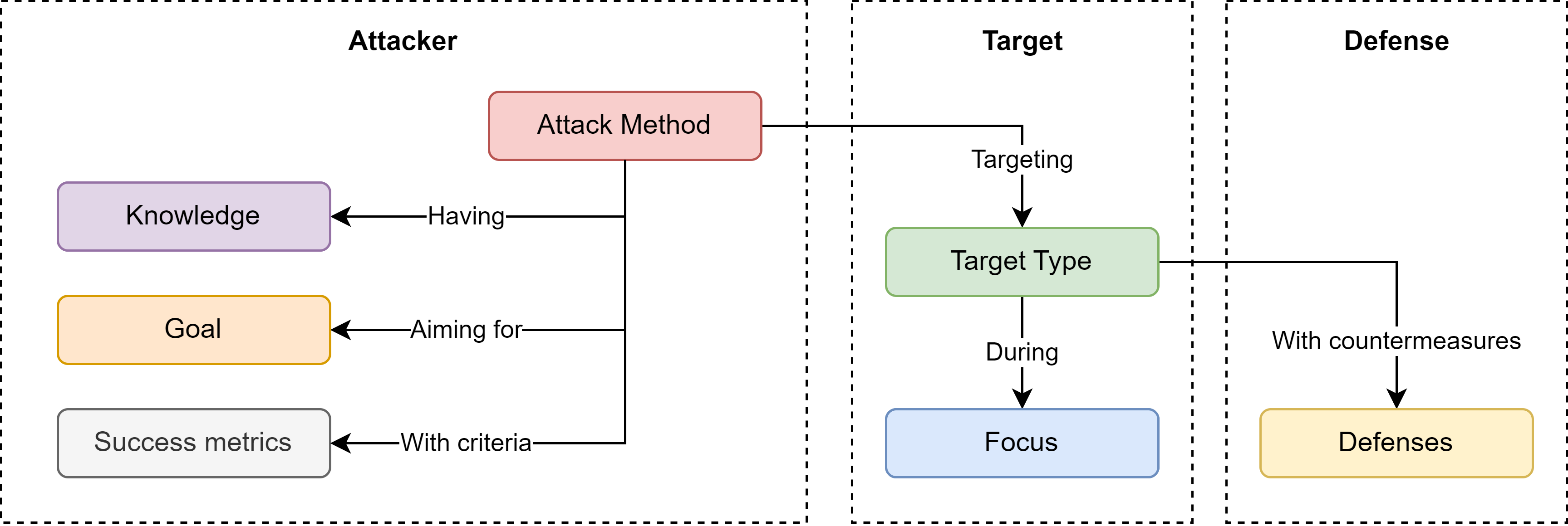}
    \caption{A 3-part model describing the relationship between the different aspects of an adversarial attack}
    \label{fig:model_3part}
\end{figure}

After the thematic analysis, we extract the data to answer each research question using the relevant codes. Additionally, we also summarize the co-occurrence of all the code types, i.e., \textit{attack types}, \textit{target models}, \textit{attack goals}, \textit{attack knowledge} \textit{metrics}, \textit{target types}, and \textit{defenses}. Based on a collection of existing taxonomies and models found in \cite{liu2018survey, martins2020adversarial, guan2018machine, zhang2020privacy} describing adversarial attacks against machine learning, as well as this review, we construct a general model of the attributes that are associated with the attack in Figure \ref{fig:model_3part}. The model follows the seven coded groups from the thematic analysis. 

The model could outline the aspects of adversarial attacks that should be considered in any research into adversarial attacks on ML. The replication package of the study is publicly available at \hyperlink{https://github.com/mathialm/Adversarial-Robustness-in-Unsupervied-Machine-Learning-A-Systematic-Review}{https://github.com/mathialm/Adversarial-Robustness-in-Unsupervied-Machine-Learning-A-Systematic-Review}.

\begin{figure}
     \centering
     \begin{minipage}[b]{0.48\textwidth}
        \centering
        \includegraphics[width=\linewidth]{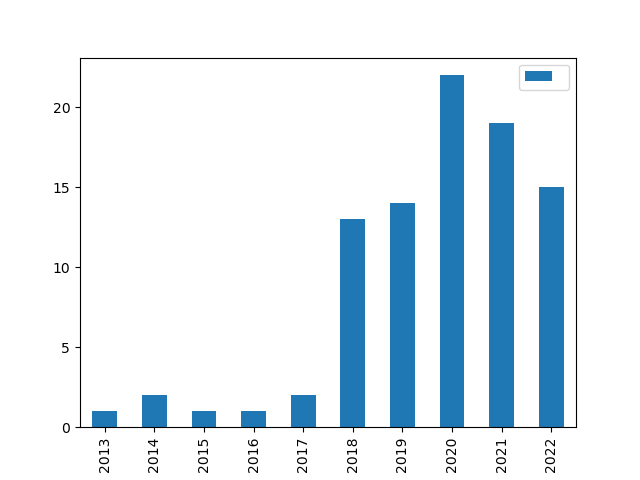}
        \caption{Number of papers published in a specific year, from 2013 to 2022}
        \label{fig:years_num}
     \end{minipage}
     \hfill
     \begin{minipage}[b]{0.48\textwidth}
          \centering
          \includegraphics[width=\linewidth]{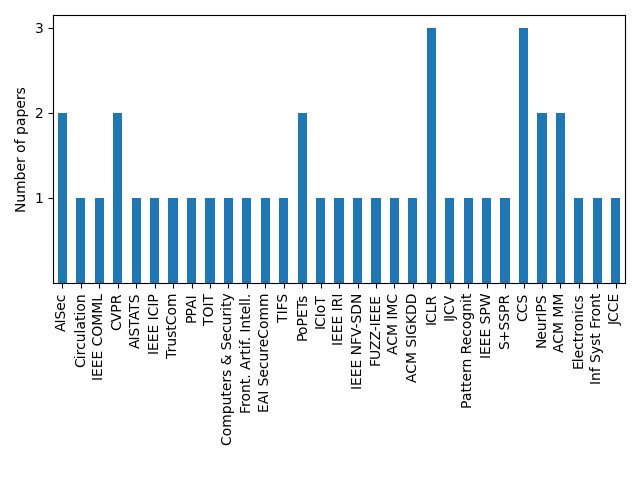}
          \caption{Number of papers in Journals and Conferences, with the long name available in Appendix  \ref{list:journal_conference_full_names}}
          \label{fig:publishers_num}
     \end{minipage}
\end{figure}

Of the final 86 papers for consideration, 39 are published as conference or journal papers, and 47 are unpublished. It can be noted that even though several of the unpublished papers also have published versions, only the papers found through the SLR directly are included. The years of publication for all the papers are found in Figure \ref{fig:years_num}, and the number of papers published in different journals and conferences are found in Figure \ref{fig:publishers_num}. Most papers on this topic were published around 2020, with relatively few papers before 2018. As expected, the journals and conferences on this topic are mainly from the databases selected to be directly used, i.e., \textit{ACM}, and \textit{IEEE}.

\section{Results of RQ1: attacks targeting UL-based systems}
\begin{figure}[ht]
    \centering
    \begin{minipage}{0.48\textwidth}
        \centering
        \includegraphics[width=\linewidth]{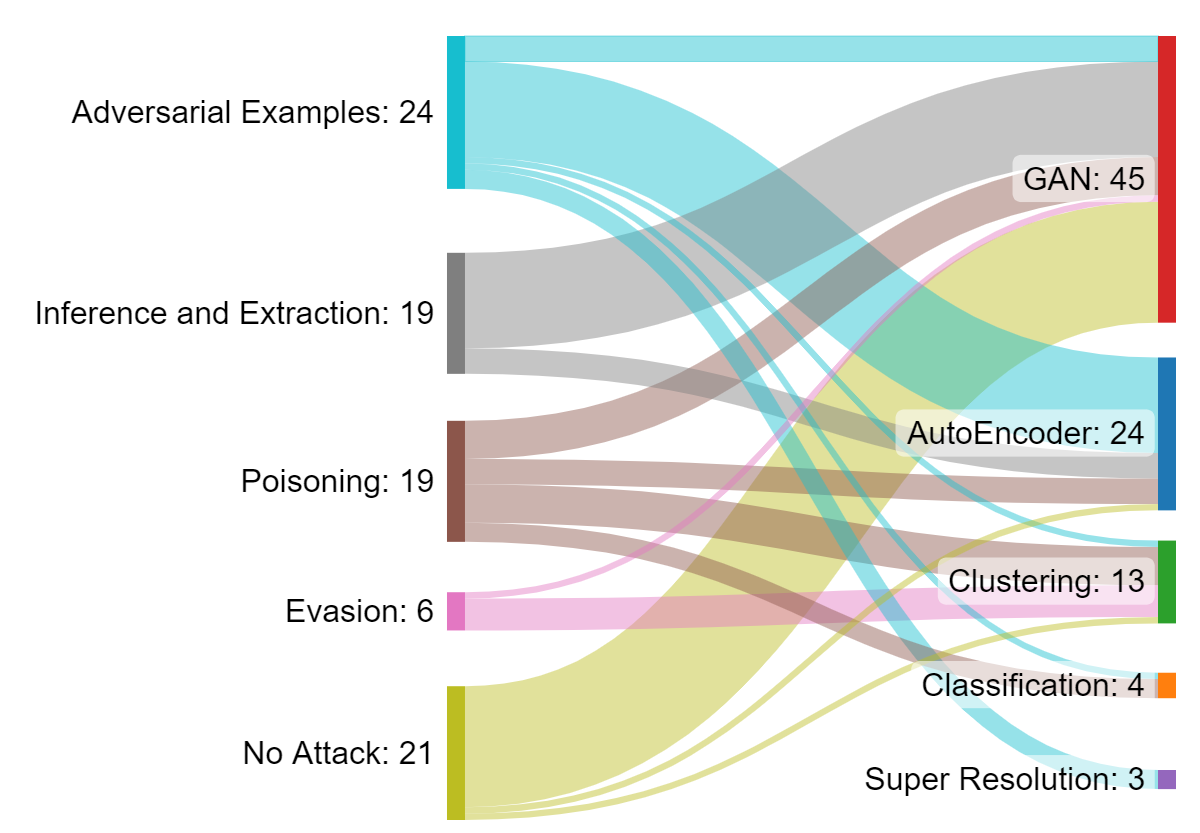}
        \caption{Sankey diagram on the number of papers covering different \textit{attack methods} against different \textit{target types}, showing the degree to which papers in this field cover different attacks against different technologies}
        \label{fig:sankey_attack_target}
    \end{minipage}
    \hfill
    \begin{minipage}{0.48\textwidth}
        \centering
        \includegraphics[width=\textwidth]{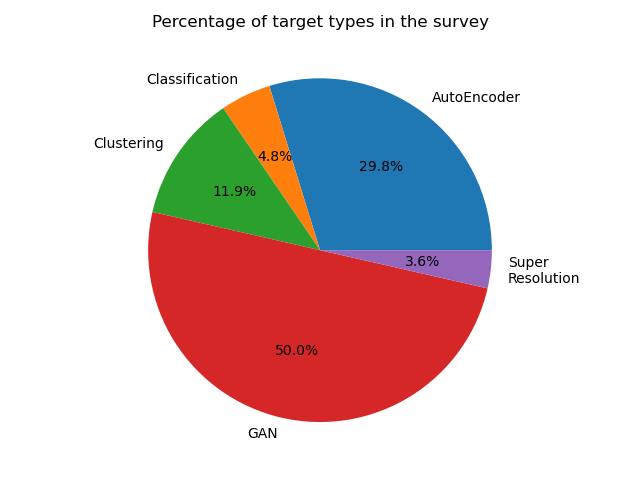}
        \caption{Distribution of the different types of UL technologies found in the review}
        \label{fig:targettypes_pie}
    \end{minipage}
\end{figure}

Below is a summary of the different kinds of \textit{attack methods}, \textit{attack knowledge}, \textit{attack goals}, \textit{metrics}, and \textit{attack focus}, for each of the UL types. A summary of the relations between kinds of attacks and targets is found in Figure \ref{fig:sankey_attack_target}, showing that most target types have a specific attack mainly associated with it.

The type of unsupervised machine learning technologies targeted by attacks is summarized in Figure \ref{fig:targettypes_pie}, showing that most of the target types covered consist of \textit{GAN}, \textit{autoencoder}, or \textit{clustering}.

In the following sections, the total percentage would, in some cases, be above 100\%, meaning some papers would be covered in multiple groups and counted more than once.

Details on \textit{target types} can be found in Appendix \ref{sec:description_target_types}; \textit{attack types} in Appendix \ref{sec:description_attack_types}; and \textit{metrics} in Appendix \ref{sec:description_metrics}.

\subsection{Attacks targeting AutoEncoder-based systems}
We found 25 papers (29\%) that cover \textit{AutoEncoder} models in the SLR.

\subsubsection{Attack Method:}
Attacks against \textit{autoencoders} include \textit{adversarial examples}, \textit{inference and extraction}, and \textit{poisoning}, which are covered by 60\%, 16\%, and 16\% of the papers on \textit{autoencoders}, respectively, and one paper \cite{FW1_51} cover the implementation of a differentially private \textit{autoencoder}-\textit{GAN} hybrid generator, aimed to protect against \textit{privacy} attacks generally.

\textit{Adversarial Examples} can work at two points, i.e.,  perturbations of the actual input \cite{17, BW1_2, FW1_3, FW1_15, FW1_30, FW1_45, FW1_46, FW1_50, FW1_52, FW2_14, FW2_15, FW2_16, FW2_26}, or on the latent code passed to the decoder \cite{10, 16, 21, FW1_15, FW2_14, FW2_15}. The attacks appear to rely on where an attacker could have reasonable access to the input data. In contexts where the entire AutoEncoder is used as a single step, e.g., removing noise, attacks do not focus directly on the latent code but rather construct an attack that tricks the encoder into creating different latent codes, changing the output. Similarly, in contexts where the adversary has direct access to the latent code, e.g., physical transmission or generation, the focus is to create perturbations that trick the decoder.

\textit{Inference and extraction} attacks are limited to \textit{membership inference} \cite{02, FW1_53, FW1_25, BW1_10}, where the adversary aims to extract information on whether the training used a specific sample. 

\textit{Poisoning} attacks can be placed into three categories, depending on how and what the adversary does. \textit{Injection} attacks \cite{FW1_15, FW2_24} aims to disrupt a model by introducing new samples into the training set, \textit{manipulation} attacks \cite{FW1_11, FW1_15, FW1_33} changes existing samples, and \textit{logic corruption} \cite{FW2_25, FW1_15} aims to introduce a backdoor into the model, which can later be abused.

\subsubsection{Attack Knowledge:}
We separate an adversary's knowledge into \textit{white box}, \textit{grey box}, and \textit{black box}, which are covered by 60\%, 40\%, and 28\% of the papers on \textit{autoencoders}, respectively.

\textit{White box} enable the attacker to craft their attack specifically to the model to enable their objective. For certain attacks like \textit{membership inference} \cite{02, FW1_15}, \textit{adversarial examples} \cite{BW1_2, FW1_3, FW1_30, FW1_46, FW2_14, FW2_15, FW2_16}, \textit{poisoning} \cite{FW1_33, FW2_24, FW2_25}, and \textit{PGD} \cite{FW1_46}, white box knowledge enables a more optimized attack. In a physical system, knowledge of the AutoEncoder enables the adversary to perform a general attack \cite{10, 21}. Additionally, white box attacks appear easier and more successful than scenarios of less knowledge \cite{02, 10, 17, 21}.

\textit{Grey box} attacks mean the adversary has incomplete knowledge of the AutoEncoder and has to make a sub-optimal attack. One such knowledge is the mapping between input and output values, whether the input is the latent code \cite{02, FW1_15, FW1_52} or the actual data \cite{FW1_11, FW1_15, FW1_45, FW1_50, FW1_52, FW2_26}. Other attacks, like \textit{adversarial examples} \cite{16}, create substitute networks from the available information to craft universal perturbations, enabling attacks against any AutoEncoder. For attacks like \textit{membership inference} \cite{FW1_25, FW1_53}, the adversary utilizes available training and test data to infer whether the training used specific samples. Grey box attacks also appear more successful than a black box and less than white box attacks \cite{02}.

\textit{Black box} attacks rely on the known vulnerabilities of AutoEncoders and make assumptions on the model and data to create the attack. For generative AutoEncoders like VAE, the adversary would only have direct access to the generated samples \cite{02, FW1_15, FW2_15}. In other systems like physical communication \cite{10}, \textit{adversarial examples} \cite{17, 21}, and \textit{membership inference} \cite{BW1_10}, the adversary uses an existing model as a substitute for the actual model, which enables the adversary to turn a \textit{black box} attack into a kind of \textit{white box} attack against the target AutoEncoder.

One paper \cite{FW1_51} focuses on implementing \textit{differential privacy}, which is commonly used for \textit{membership inference} attacks and would be most helpful in the context of limited knowledge.

\subsubsection{Attack Goal:}
We separate an adversary's goal into \textit{integrity}, \textit{privacy}, and \textit{availability}, which are covered by 68\%, 20\%, and 12\% of the papers on \textit{autoencoders}, respectively.

\textit{Integrity} attacks aim to introduce new functionality, change existing functionality, or avoid the existing functionality of an AutoEncoder. Attacks often focus on maximization of distortion of the model's output, on all, or subsets, of the input data \cite{FW1_3, FW1_15}. For attacks like \textit{adversarial examples} \cite{17, BW1_2, FW1_30, FW1_33, FW1_45, FW1_46, FW1_50, FW1_52, FW2_14, FW2_15, FW2_16, FW2_26}, the adversary aim to force a change in the latent representation of the \textit{autoencoder}, in turn producing different output from what the input data intended to produce. Similarly, \textit{poisoning} attacks focus on introducing unintended functionality into the model, e.g., backdoors on specific input \cite{FW1_11, FW2_25}, or prevent intended functionality of the model \cite{FW2_24}.

\textit{Privacy} attacks have as a goal to extract some private information from the \textit{autoencoder} \cite{FW1_51}. \textit{Membership inference} \cite{02, BW1_10, FW1_25, FW1_53} is the primary type of privacy attack against AutoEncoders, aiming to determine if a specific sample was used in training by analyzing the output from the model.

\textit{Availability} attacks aim to prevent the intended functionality of the model by performing the attack. In physical communication \cite{10, 16, 21}, using \textit{adversarial examples} causes a jamming attack on the decoder, preventing the model from decoding the encoded message.

\subsubsection{Metrics to measure attack success:}
We separate the metrics used by an adversary to determine attack success into \textit{distance metrics}, and \textit{true/false positive/negative} based, which are covered by 60\%, and 56\% of the papers on \textit{autoencoders}, respectively.

\textit{Distance metrics} refer to using a mathematical formula comparing two or two sets of samples to determine the amount of difference between them. Some metrics, like \textit{Structured Similarity Index Measure} (SSIM), \textit{Peak Signal-to-Noise Ratio} (PSNR), \textit{Mean Square Deviation} (MSD), and \textit{Root Mean Square Deviation} (RMSD) are used for comparing whether there is a noticeable difference between malicious and clean samples \cite{02, BW1_2, BW1_10, FW1_3, FW1_11, FW1_45, FW2_14, FW2_15, FW2_16, FW2_25}, with \textit{Evidence Lower BOund} (ELBO) \cite{FW1_30} denoting the performance of the model in regards to the quality. \textit{Area Under Distortion-Distortion Curve} (AUDDC) is used to measure how well an attack can influence the output of the model \cite{FW2_26}. Other metrics, like \textit{$L_2$}, are used as a threshold of perturbations in the input \cite{FW1_15, FW1_33, FW1_50, FW2_14}.

\textit{True/False Positive/Negative based metrics} are used to determine the rate of success and failure by the attackers and determine how many measured instances are successes (true positive and true negative) and how many are failures (false positive and false negative). One way of using these metrics is to measure the performance of the attacked system with and without the influence of the attacker, e.g., detection systems \cite{FW2_24}, classifiers using output from UL systems \cite{10, 17, FW1_3, FW2_16}, and physical communication \cite{10, 16, 21}. In \textit{adversarial examples}, \textit{success rate} is used for determining if the attack works, both in a targeted and untargeted manner \cite{FW2_14}. For other attacks like \textit{membership inference} \cite{02, BW1_10, FW1_25, FW1_46, FW1_53}, the success is measured by how well their attack can differentiate if a sample was used for training by the \textit{autoencoder}. 

Additionally, two papers do not contribute with metrics to determine attack success, as they cover implementation of \textit{differential privacy} \cite{FW1_51}, or an analysis on adversarial robustness to \textit{adversarial examples} \cite{FW1_52}.

\subsubsection{Attack focus:}
We separate where the adversary focuses into \textit{production} and \textit{training}, which are covered by 92\%, and 12\% of the papers on \textit{autoencoders}, respectively.

\textit{Production} models exist within their respective working environments, exposing them somewhat to the outside world. \textit{Privacy} attacks like \textit{membership inference} \cite{02, BW1_10, FW1_25, FW1_53} mainly only focus on production models, as the data pursued are no longer directly used by the model. Defenses like \textit{differential privacy} \cite{FW1_51} also would only be applicable for protecting a production model. \textit{Adversarial examples} \cite{10, 16, 17, 21, BW1_2, FW1_3, FW1_15, FW1_30, FW1_33, FW1_45, FW1_46, FW2_14, FW2_15, FW2_16, FW2_26} is another type of attack mainly used on production models, where the adversary somehow has direct access to the input to the model, and can introduce perturbations to the input samples. Some papers cover generalized increases of \textit{autoencoder} robustness to perturbations \cite{FW1_50, FW1_52}, while others cover uses of \textit{differential privacy}  to enable privacy guarantees on the training data \cite{FW1_51, FW1_53}. 

\textit{Training} models are models in the context of training for a specific purpose. \textit{Poisoning} attacks \cite{FW1_11, FW1_33, FW2_24, FW2_25} is the only attack targeting \textit{autoencoder} in training by modifying the training data itself to modify the training of the model in a specific direction.

\begin{tcolorbox}[colback=blue!5!white,colframe=blue!50!black,
  colbacktitle=blue!75!black,title=Summary of attacks on \textit{AutoEncoders}:]
  For \textit{autoencoders}, the main source of attacks stems from \textit{adversarial examples}, being usable against both \textit{discriminatory} and \textit{generative} models, often in a limited knowledge scenario in order to disrupt the output from the model. There are also attacks focusing on extracting information about the training data, which often are performed in minimum knowledge scenarios. \textit{Poisoning} attacks are also used to disrupt or introduce new functionality to the model. An adversary targeting \textit{autoencoders} also often relies on \textit{distance metrics} to determine the difference between clean and malicious data in order to avoid simple detection techniques, while \textit{true/false positive/negative} metrics are used for determining how successful the attack is.
\end{tcolorbox}

\subsection{Attacks targeting UL supported classification}
We found 14 (16\%) papers that cover UL-supported classification. However, not every paper covered focuses on the adversarial robustness of the classifier itself, and instead focuses on the use of classifiers to increase robustness \cite{18}, or the use of classifiers as a baseline for determining whether attacks were successful \cite{12, 24, FW1_29, FW1_40, FW1_42, FW1_53, FW2_8, FW2_11, FW2_14}. In other words, only four (5\%) papers cover the adversarial robustness of classifiers in conjunction with unsupervised learning.

\subsubsection{Attack Method:}
Attacks against \textit{UL supported classification} include \textit{poisoning} and \textit{adversarial examples}, which are covered by 75\%, and 25\% of the papers on \textit{UL supported classification}, respectively.

\textit{Poisoning} attacks \cite{19, FW1_35} in the classifier's training are particularly effective in disrupting its performance. Similarly, poisoning attacks are also effective at disrupting the performance of a classifier based on \textit{domain adaption} \cite{FW1_47} by affecting the training of the unsupervised model. 

\textit{Adversarial examples} against UL-supported classifiers means that the adversary crafts perturbations, which affect the UL system to the degree that the classification is also affected. An \textit{AutoEncoder}  is sometimes used before a classifier to denoise the input before the classifier. Attacks on this often consist of \textit{adversarial examples}, constructed to disrupt the classifier performance \cite{17}.

\subsubsection{Attack Knowledge:}
We separate an adversary's knowledge into \textit{white box} and \textit{black box}, which are covered by 75\% and 75\% of the papers on \textit{UL supported classification}, respectively.

\textit{White box} attacks, in the use of \textit{AutoEncoder}, corresponds to improved accuracy under \textit{adversarial examples} \cite{17}. In \textit{semi-supervised} \cite{19} and \textit{domain adaption} \cite{FW1_47} scenarios, poisoning with white-box knowledge can significantly reduce the classifier performance.

\textit{Black box} attacks against \textit{UL supported classification} with an \textit{AutoEncoder} has a decreased performance with \textit{adversarial examples} \cite{17}. Additionally, \textit{semi-supervised} \cite{19, FW1_35} and black box \textit{poisoning} attacks appear to be marginally worse than the white box, but are still able to significantly reduce classifier performance.  

\subsubsection{Attack Goal:}
An adversary's goal against \textit{UL supported classification} is \textit{integrity} in 100\% of the papers.

\textit{Integrity} attacks like \textit{poisoning} against \textit{domain adaption} \cite{FW1_47} and \textit{semi-supervised} models \cite{19, FW1_35} can construct poisoned samples that decrease the performance of the resulting classifier, while \textit{adversarial examples} \cite{17} can manipulate the classification results by perturbing the input. 

\subsubsection{Metrics to measure attack success:}
We separate the metrics used by an adversary to determine attack success into \textit{distance metrics}, and \textit{true/false positive/negative} based, which is covered by 50\% and 75\% of the papers on \textit{UL supported classification}, respectively.

\textit{Distance metrics} have several different ways of measurement. One measurement of success is the degree to which the attacker can influence the results of the classifier through an unsupervised model, like with \textit{semi-supervised} \cite{19}, where the \textit{RMSE} of the classification is used to determine success. When joined with an \textit{autoencoder}, classification \textit{accuracy} is used to determine the success of an attack \cite{17}. Another measurement is the degree to which the attacker perturbs the input, as unrestricted perturbation would be visible to statistical analysis or human evaluation. For this, the \textit{$L_2$} norm \cite{19} can be used to determine perturbations in a \textit{poisoning} attack. 

\textit{True/False Positive/Negative}-based metrics, like how well the classifier can function under attack, is one way of measuring attack success, e.g., \textit{adversarial examples} \cite{17} and \textit{poisoning} \cite{FW1_35, FW1_47} both utilizing the \textit{accuracy} of the classifier.

\subsubsection{Attack focus:}
Where the adversary focuses their attack is separated into \textit{training} and \textit{production}, which are covered by 75\% and 25\% of the papers on \textit{UL supported classification}, respectively.

\textit{Training} systems have adversaries that focus on disrupting classifier functionality by performing a \textit{poisoning} attack on \textit{semi-supervision} \cite{19, FW1_35} and \textit{domain adaption} \cite{FW1_47} models.

\textit{Production} systems have adversaries where the focus is on disrupting the functionality of a classifier in a real-world environment, specifically using an \textit{autoencoder} before the classifier itself \cite{17}, as it is targeted by \textit{adversarial examples}. 

\begin{tcolorbox}[colback=blue!5!white,colframe=blue!50!black,
  colbacktitle=blue!75!black,title=Summary of attacks on \textit{UL supported classification}:]
Attacks on \textit{UL supported classification} are mainly \textit{adversarial examples}, where the adversary perturbs data to influence the UL part, in turn influencing the classification; and \textit{poisoning}, where the adversary introduces poisoned samples into the training of \textit{domain adaption} which influence the performance of the resulting classifier. Attacks appear to be most successful in a high-knowledge scenario. The measurements of success mainly focus on the classifier's performance, e.g., \textit{accuracy}, with \textit{distance metrics} used to measure how much the samples are perturbed to avoid simple detection.
\end{tcolorbox}

\subsection{Attacks targeting clustering}
We found ten papers (12\%) covering unsupervised clustering. 

\subsubsection{Attack Method:}
Attacks against \textit{clustering} include \textit{poisoning}, \textit{evasion}, and \textit{adversarial examples}, which are covered by 60\%, 50\%, and 10\% of the papers on \textit{clustering}, respectively.

\textit{Poisoning} attacks work by inserting crafted samples that compromise the cluster's functionality. One such attack is the \textit{bridging} attack \cite{13, 14, 15, FW1_9}, where the adversary crafts samples that "bridges" gaps between clusters, in turn altering the resulting clusters. Other approaches involve manipulating nodes and edges of \textit{graph-based clustering} \cite{FW1_23} to disrupt the system or to alter existing samples from the training set according to a specified maximum total perturbation \cite{FW1_6}. 

\textit{Evasion} attacks consist of crafting samples with features overlapping with the actual sample set to conceal the sample and avoid detection without significantly altering the clustering functionality \cite{12, 13, 14, 15, FW1_4}. 

\textit{Adversarial examples} work by optimizing the perturbation relating to the maximization of the clustering results \cite{FW1_10}. 

One paper \cite{BW2_9} covers the construction of a privacy constraint enforced in training to prevent private sample information from being inferred to protect against privacy attacks.

\subsubsection{Attack Knowledge:}
We separate an adversary's knowledge into \textit{white box}, \textit{grey box}, and \textit{black box}, which are covered by 50\%, 50\%, and 10\% of the papers on \textit{clustering}, respectively.

\textit{White box} attacks can optimize \textit{poisoning} \cite{13, 14, 15, FW1_9, FW1_23} attacks, manipulate the resulting cluster, and optimize \textit{obfuscation} \cite{13} attacks by crafting adversarial samples to hide samples in existing clusters. 

\textit{Grey box} means adversaries would have access to a limited data mapping with its resulting cluster. \textit{Bridging} attacks \cite{13} can be used to gradually discover samples that disrupt the function of the resulting cluster, or for \textit{evasion} \cite{FW1_4} and \textit{adversarial examples} \cite{FW1_10}, use the mapping between samples and resulting clusters to optimize perturbation needed to avoid detection by the clustering. Some methods of \textit{poisoning} attacks \cite{FW1_6} rely on constructing a general perturbation based on the mapping between samples and resulting clusters by using a \textit{genetic algorithm} to discover optimizations iteratively. Other limited knowledge attacks rely on using approximate models based on open-sourced data and performing \textit{white box} attacks against these approximate models to construct their attack \cite{FW1_23}. 

\textit{Black box} attacks include adversaries who can employ an \textit{evasion} attack \cite{12} by slowly enforcing drift in the clustering towards a specific point, which does not require any information of the model.

One paper \cite{BW2_9} enforces privacy constraints in training and would likely be most applicable in limited knowledge scenarios.

\subsubsection{Attack Goal:}
We separate an adversary's goal into \textit{integrity}, \textit{privacy}, and \textit{availability}, which are covered by 68\%, 20\%, and 12\% of the papers on \textit{clustering}, respectively.

\textit{Availability} attacks work by preventing or disrupting the general functionality of the clustering. \textit{Poisoning} attacks \cite{13, 14, 15, FW1_6, FW1_9, FW1_23} work by introducing specific samples to the training to influence the shape of the resulting clusters, to the degree that prevents the clustering from working as intended.

\textit{Integrity} attacks focus on influencing the clustering results regarding specific samples. \textit{Evasion} attacks aim to hide specific samples in existing clusters by overlapping their features with those of actual samples \cite{12, 13} or by slowly introducing modified samples to drift the clustering in a specific direction \cite{FW1_4}. \textit{Adversarial examples} \cite{FW1_10} aim to modify samples in a way that results in the misclustering of samples.  

\textit{Privacy} attacks cover theoretical constraints on clusters to ensure the privacy of the samples used to train the model \cite{BW2_9}.

\subsubsection{Metrics to measure attack success:}
We separate the metrics used by an adversary to determine attack success into \textit{true/false positive/negative}, \textit{distance metrics},  \textit{number of clusters}, and \textit{objective function}, which are covered by 50\%, 40\%, 40\%, and 30\% of the papers on \textit{clustering}, respectively.

\textit{True/false positive/negative} metrics, like \textit{accuracy} used in detection \cite{12, FW1_4} to determine its performance against \textit{evasion} attacks. In \textit{poisoning} attacks, \textit{F-measure} \cite{15} and \textit{success rate} \cite{FW1_23} are used for determining the quality of the clusters produced under attack. In \textit{adversarial examples} \cite{FW1_10}, \textit{accuracy} measures how well the clusters fit the labels.

\textit{Distance metrics} are used in \textit{poisoning} attacks \cite{13, FW1_6} to craft samples to avoid simple detection. Metrics like \textit{$L_0$}, \textit{$L_2$}, and \textit{$L_\infty$} norms are used to limit the number of perturbations added to the samples. In contrast, others use a \textit{feature modification threshold} \cite{FW1_4} to limit the amount any single feature can be modified. For \textit{adversarial examples}, \textit{Normalized Mutual Information} (NMI) is used to compare actual and predicted cluster labels, and \textit{distortion} measures the perturbation of the sample \cite{FW1_10}. 

\textit{Number of clusters} is used to compare the performance of attacks and their capability of reducing or increasing the number of clusters the model produces. \textit{Poisoning} attacks \cite{13, 14, 15, FW1_9} use this metric to determine how well the attack works.

\textit{Objective function} based metrics are used in \cite{13, 14, 15} and are crafted for the specific attack and model.

One paper \cite{BW2_9} does not describe concrete attacks and, as such, does not outline any metrics for the success of an attack.

\subsubsection{Attack focus:}
We separate the adversary focus into \textit{production} and \textit{training}, which are covered by 50\%, and 50\% of the papers on \textit{clustering}, respectively.

\textit{Production} systems are targeted by attacks like \textit{evasion} \cite{12, 13, FW1_4} and \textit{adversarial examples} \cite{FW1_10}, where the aim is to reduce the functionality of specific samples while keeping the general clustering functionality intact, while other attacks like \textit{targeted noise injection} and \textit{small community} attacks \cite{FW1_23} aim to disrupt the functionality of existing \textit{graph-based clusters}. Additionally, papers, like \cite{BW2_9}, aim to ensure clusters' privacy guarantees to prevent privacy attacks, such as \textit{membership inference}. 

\textit{Training} systems are targeted by \textit{poisoning} attacks \cite{13, 14, 15, FW1_6, FW1_9}, modifying or injecting malicious samples to the training set, which can significantly reduce the functionality of the clustering systems.

\begin{tcolorbox}[colback=blue!5!white,colframe=blue!50!black,
  colbacktitle=blue!75!black,title=Summary of attacks on \textit{clustering}:]
\textit{Poisoning} attacks on clustering involve introducing samples to disrupt the cluster functionality, merge, or increase the number of clusters found. Evasion attacks work by hiding specific samples within clusters to avoid detection without altering the cluster's functionality. Most attacks on clustering rely on increased knowledge, as the adversary has a hard time accomplishing their goals without knowing the data samples used for training. \textit{True/False Positive/Negative} metrics are used by the adversary to determine how good the resulting cluster is, measuring how well the cluster fits the labels. In contrast, \textit{distance metrics} limit the amount of perturbation a sample can have. The \textit{number of clusters} is also used to determine the success of the attacks, particularly when the attack's goal revolves around changing the cluster.
\end{tcolorbox}

\subsection{Attacks targeting GAN}
We found 42 papers (48\%) covering GAN models.

\subsubsection{Attack Method:}
Attacks against \textit{GAN} include \textit{inference and extraction}, \textit{poisoning}, \textit{adversarial examples}, and \textit{evasion}, which are covered by 33\%, 14\%, 10\%, and 2\% of the papers on \textit{GAN}, respectively. Additionally, 45\% of the papers cover implementations of privacy measures to prevent \textit{privacy} attacks without specifying concrete attacks. These include implementations of \textit{differential privacy} for \textit{GANs} in order to facilitate sharing models and datasets without risk of privacy violations \cite{BW1_4, BW1_5, BW1_6, BW1_11, BW1_14, BW1_15, FW1_17, FW1_29, FW1_38, FW1_51, BW2_3, BW2_7, FW2_3, FW2_8, FW2_11, FW2_19, FW2_30}, context-dependent implementations of privacy guarantees to enable private sharing \cite{FW1_40}, and a survey of privacy challenges in \textit{GANs} \cite{FW2_29}.

\textit{Inference and extraction} attacks, like \textit{membership inference} attacks, are methods of attacks to determine whether specific samples were used during the training of the GAN model \cite{BW1_8, FW1_43}. Methods for this include using distance metrics between the sample and generated samples \cite{02, BW1_10, FW1_2, FW1_15, FW1_16}, using discriminator confidence \cite{BW1_9, FW1_15, FW1_16}, or training a classifier \cite{03, FW1_15, FW1_42, FW1_54}, in order to determine whether the training used a sample. Other methods determine membership by controlling the latent code of the generator and determining membership based on the reconstruction error of the generator when trying to generate specific samples \cite{FW2_17}. \textit{Attribute inference} is another way of extracting private attributes from GANs by looking at publicly known features, collecting generated samples, and using regression to predict the private features \cite{FW1_15, FW1_42}. \textit{Model extraction} is a method for cloning an existing model based on the generated samples of a target GAN, like \textit{accuracy extraction} \cite{FW1_15, FW1_27}, where the adversary measures the difference in the distribution of the cloned and actual model; and \textit{fidelity extraction} \cite{FW1_15, FW1_27} compares the distribution of the cloned model to the actual training data set. The cloned model can subsequently be used as the substitute model in a \textit{white-box} \textit{membership inference} attack \cite{BW1_9, FW1_20}. 

\textit{Poisoning} attacks is, among other, used against federated GAN training, where the adversary can contribute to the global model as one or multiple participant(s) and subsequently modify the information that it contributes, and affect the performance of the global model \cite{24, FW2_13}. Other methods modify the training data directly, either by injecting new samples \cite{FW1_15}, modifying existing samples to disrupt the model \cite{FW1_15}, or modifying samples to introduce new by-products in the generated samples \cite{FW1_11, FW1_13, FW1_15, FW2_13, FW2_25}.

\textit{Adversarial examples} attack relies on access to the latent code used by the generator and perturbing some input into it to generate a completely different sample from what would have been generated without the perturbation. \textit{Adversarial examples} can be used during training for \textit{adversarial training} \cite{18, FW1_32}, increasing the robustness of the model. Other papers focus on implementing changes to how the \textit{GAN} works in order to be theoretically robust against \textit{adversarial examples} \cite{FW1_8}. Additionally, metrics commonly used to determine sample quality, e.g., \textit{Fréchet Inception Distance} (FID) and \textit{Inseption Score} (IS), are also explored and are determined to be insufficient in discovering \textit{adversarial examples} \cite{FW2_2}.

\textit{Evasion} attacks focus on finding minimal perturbations on the latent codes, which produce unsatisfactory samples, compared to what is produced by the unperturbed latent code \cite{FW1_15}, in order to generate unsatisfactory samples, but still following the statistical distribution of the training samples.

\subsubsection{Attack Knowledge:}
We separate an adversary's knowledge into \textit{white box}, \textit{grey box}, and \textit{black box}, which are covered by 26\%, 26\%, 26\% of the papers on \textit{GAN}, respectively. In contrast, 45\% of the papers cover \textit{differential privacy} implementations \cite{BW1_4, BW1_5, BW1_6, BW1_11, BW1_14, BW1_15, FW1_17, FW1_29, FW1_38, FW1_51, BW2_3, BW2_7, FW2_3, FW2_8, FW2_11, FW2_19, FW2_30}, other privacy guarantees \cite{FW1_40}, and general privacy challenges \cite{FW2_29}, and would likely be applicable in scenarios of limited knowledge, where the adversary does not have access to training data, i.e., \textit{black box} or \textit{grey box}.

\textit{White box} attacks, like \textit{membership inference} \cite{02, BW1_9, BW1_10, FW1_15}, \textit{fidelity extraction} \cite{FW1_27}, and \textit{poisoning} \cite{FW1_11, FW1_15, FW2_25}, enable full knowledge of the parameters of the models which allow the attacker to fine-tune the attack towards the specific model. Additionally, for \textit{membership inference} attacks, full knowledge allows for querying the discriminator to extract information on the training data \cite{BW1_8, BW1_9, FW1_2, FW1_15, FW1_16, FW1_27, FW2_17}. Against GANs, the access to internal information on the model allows for more successful attacks compared to more restricted knowledge scenarios \cite{02, BW1_9, BW1_10, FW1_27}.

\textit{Grey box} attacks, e.g., \textit{membership inference} \cite{02}, mean that adversaries know the input-output mapping of the model, i.e., the latent code and the generated samples, but not the internal parameters of the discriminator or generator models. Other \textit{membership inference} attacks \cite{BW1_9, FW1_15, FW1_54}, as well as \textit{fidelity extraction} attacks \cite{FW1_15, FW1_27}, require limited knowledge of the training and test data, representing potential public databases that an adversary would have access to. In comparison, \textit{poisoning} \cite{24, FW2_13} attacks on \textit{federated GAN} allow for knowing the training model's structure but not the training data of the other participants. Other \textit{poisoning} \cite{FW1_13, FW1_15} attacks, \textit{evasion} \cite{FW1_15} attacks, \textit{adversarial examples} \cite{18, FW1_8, FW1_32, FW2_2}, and \textit{property inference} attacks \cite{FW1_54} instead rely entirely on the ability of the adversary to control the input to the model, i.e., training data or latent code. In general, the more knowledge an adversary has, the better performance of the attack \cite{02, FW1_27, FW1_54}.

\textit{Black box} attacks, like in the case of \textit{membership inference}, mean that the adversary only has access to the generated samples from the model, where attacks rely on inferring if samples are part of the training based on distance calculations between the sample and the generated set \cite{02, FW1_15}, or by training classifiers on the generated set \cite{03, FW1_20, FW1_54}. Other approaches for \textit{membership inference} \cite{FW1_42, FW1_43, FW1_54, BW1_9, BW1_10} and \textit{model extraction} \cite{FW1_15, FW1_27} attacks rely on recreating a substitute model and perform \textit{white-box} attacks on this model,  later used for an attack on the original model.

\subsubsection{Attack Goal:}
We separate an adversary's goal into \textit{privacy} and \textit{integrity}, which are covered by 79\%, and 24\% of the papers on \textit{GAN}, respectively.

\textit{Privacy} attacks, like \textit{membership inference} \cite{02, 03, BW1_8, BW1_9, BW1_10, FW1_2, FW1_15, FW1_16, FW1_20, FW1_42, FW1_43, FW1_54, FW2_17}, have a goal to determine whether the GAN model used a specific sample for training, often measured by the \textit{success rate} \cite{03}, \textit{accuracy} \cite{BW1_8, BW1_9, BW1_10, FW1_2}, or \textit{Area Under the Curve of the Receiver Operator Characteristic curve} (AUCROC) \cite{02}. Other attacks like \textit{attribute inference} \cite{FW1_15}, and \textit{model inference} \cite{FW1_15, FW1_27} attacks extract data distribution, for the former, and model information, for the latter, using metrics like \textit{accuracy} \cite{FW1_27} to determine the success of the attack. Many papers also cover the application of \textit{differential privacy}  \cite{BW1_4, BW1_5, BW1_6, BW1_11, BW1_14, BW1_15, FW1_17, FW1_29, FW1_38, FW1_51, BW2_3, BW2_7, FW2_3, FW2_8, FW2_11, FW2_19, FW2_30}, or other privacy guarantees \cite{FW1_40} in training to ensure the privacy of the training samples, as a way to prevent private data from being inferred from released datasets or models. One paper also covers the privacy challenges in \textit{GANs} \cite{FW2_29}.

\textit{Integrity} attacks like \textit{poisoning} \cite{24, FW1_11, FW1_13, FW1_15, FW2_13, FW2_25} focus on decreasing the quality of generated samples or altering the mapping between latent code and generated samples, while \textit{evasion} attacks \cite{FW1_15} and \textit{adversarial examples} \cite{18, FW1_8, FW1_32} force the model to output unsatisfactory output with small perturbations in the input.

\subsubsection{Metrics to measure attack success:}
We separate the metrics used by an adversary to determine attack success into \textit{true/false positive/negative}, and \textit{distance metrics} based, which are covered by 38\%, and 19\% of the papers on \textit{GAN}, respectively, with 45\% of the papers covering protection against general privacy attacks, like, \textit{differential privacy} \cite{BW1_4, BW1_5, BW1_6, BW1_11, BW1_14, BW1_15, FW1_17, FW1_29, FW1_38, FW1_51, BW2_3, BW2_7, FW2_3, FW2_8, FW2_11, FW2_19, FW2_30}, context-dependent implementations of privacy guarantees \cite{FW1_40}, and general explorations of privacy challenges in \textit{GANs} \cite{FW2_29}, and do not outline any metrics for measuring attack success.

\textit{True/false positive/negative} metrics in \textit{membership inference} attacks measure how well the adversary can determine if the query sample belongs to the training set, measured by \textit{AUCROC} \cite{02, FW1_42, FW1_54, FW2_17}, \textit{Area Under Precision-Recall Curve} (AUPRC) \cite{FW1_20, FW1_42}, \textit{F1-score} \cite{FW1_42}, \textit{accuracy} \cite{BW1_8, BW1_9, BW1_10, FW1_2, FW1_16, FW1_27, FW1_42}, \textit{precision} \cite{FW1_20}, \textit{recall} \cite{FW1_20}, \textit{true positive} \cite{BW1_8, FW1_43, FW2_17}, \textit{false positive} \cite{BW1_8, FW1_43, FW2_17}, and \textit{success rate} \cite{03, FW1_15}. For \textit{poisoning} attacks, \textit{accuracy} \cite{24} is used to determine both if the poisoned task and the main task of the model perform well. For \textit{adversarial examples}, \textit{accuracy} of the discriminator determines how well the attack can influence the generated samples \cite{18}.

\textit{Distance metrics} in \textit{membership inference} attacks, consists of \textit{reconstruction distance} \cite{FW1_15} and \textit{$\epsilon$-ball distance} \cite{FW1_15}, which determine the distance between the query sample, and the set of generated samples, to determine if the training used the sample. In \textit{poisoning} attacks, \textit{SSIM} \cite{FW1_11}, \textit{PSNR} \cite{FW1_11}, \textit{IS} \cite{FW1_13, FW2_13}, \textit{FID} \cite{FW1_13, FW2_13, FW2_25}, and \textit{Kernel Inception Distance} (KID) \cite{FW2_13} are used for evaluating the performance of the model in non-adversarial and adversarial scenarios. With \textit{adversarial examples}, \textit{SSIM} and \textit{FID} are used for evaluating the performance of the generated samples under adversarial noise \cite{FW1_8}. Additionally, distance metrics measure the perturbations applied to data in \textit{evasion} attacks \cite{FW1_15} and \textit{adversarial examples} \cite{FW1_32}. \citet{FW2_2} evaluate the \textit{FID} and \textit{IS} metrics in terms of the metrics' ability to discover \textit{adversarial examples}, which the metrics are not able to.

\subsubsection{Attack focus:}
We separate the adversaries' focus into \textit{production} and \textit{training}, which are covered by 88\%, and 12\% of the papers on \textit{GAN}, respectively.

\textit{Production} systems under \textit{membership inference} \cite{02, 03, BW1_8, BW1_9, BW1_10, FW1_2, FW1_15, FW1_16, FW1_20, FW1_42, FW1_43, FW1_54, FW2_17}, \textit{attribute inference}, \cite{FW1_15, FW1_42} and \textit{model extraction} \cite{FW1_15, FW1_27} attacks rely on using the generated samples from the model to infer information. 

\textit{Training} systems under \textit{poisoning} attacks are mainly attacked by altering the training data of the model \cite{FW1_11, FW1_13, FW2_25}. In contrast, in \textit{federated GAN} works by poisoning the global model as a participant in the federated training \cite{24, FW2_13}.

\begin{tcolorbox}[colback=blue!5!white,colframe=blue!50!black,
  colbacktitle=blue!75!black,title=Summary of attacks against \textit{GANs}:]
The primary attacks against \textit{GANs} revolve around extracting information from the model based on the generated samples they produce, including: \textit{membership inference}, where the adversary tries to find if the training used a specific sample, \textit{attribute inference} to extract the features of the training data, and \textit{model extraction}, where one clone the model, in turn giving the adversary access to optimize attacks on the cloned model to attack the original model subsequently. \textit{Poisoning} attacks are also used, where the adversary crafts samples to be used in training, either to decrease the utility of the generated models or to introduce backdoors into the generated samples. Knowledge of the model is also essential, as more knowledge enables optimized attacks. However, \textit{membership inference} attacks are made with only knowledge of the generated samples. \textit{Distance metrics} are used for determining the model's performance by analyzing the clean and attacked generated samples. In contrast, \textit{true/false positive/negative} determines the degree to which the attack was successful.  
\end{tcolorbox}

\subsection{Super Resolution}
\label{sec:attributes_super_res}
We found three papers (3\%) covering super-resolution models.

\subsubsection{Attack Method:}
Attacks against \textit{super resolution} only include \textit{adversarial examples}.

\textit{Adversarial example} attacks perturb the input to the model to alter the result significantly \cite{FW1_39, FW2_7, FW2_37}. One paper also produces general perturbations for super-resolution \cite{FW1_39}, i.e., the perturbation is independent of the model's training.

\subsubsection{Attack Knowledge:}
The adversary's knowledge is only covered by \textit{white box} attacks.

\textit{White-box} attacks revolve around knowing a mapping between input and output, and, with limited perturbations on the input, being able to produce large perturbations on the output \cite{FW1_39, FW2_7, FW2_37}. For this attack, an adversary would require access to the model to know the perturbed output. The perturbation is iteratively fine-tuned, which also applies to the general perturbation \cite{FW1_39}.

\subsubsection{Attack Goal:}
\textit{Integrity} attacks are the only goal covered for \textit{super-resolution}.

\textit{Integrity} attacks, like \textit{adversarial examples}, have as a goal to insert imperceptible perturbation on the input to the model, in turn causing an extensive deterioration of the output from the model \cite{FW1_39, FW2_7, FW2_37}.

\subsubsection{Metrics to measure attack success:}
\textit{distance metrics} are the only metrics an adversary uses to determine attack success for \textit{super-resolution}.

\textit{Distance metrics} include the difference between perturbed and unperturbed input and output measurements. The metrics used for this include \textit{PSNR} \cite{FW1_39, FW2_7, FW2_37} measuring noise, \textit{SSIM} \cite{FW2_7, FW2_37} measuring perceptual differences, and \textit{Learned Perceptual Image Patch Similarity} (LPIPS) \cite{FW2_7} measuring the difference in activation in the model on two inputs.

\subsubsection{Target Focus:}
The adversary only focuses their attack on \textit{production} systems.

\textit{Production} systems under \textit{adversarial examples} mean an adversary makes perturbations to the input to the model \cite{FW1_39, FW2_7, FW2_37}, in order to cause large perturbations in the output.

\begin{tcolorbox}[colback=blue!5!white,colframe=blue!50!black,
  colbacktitle=blue!75!black,title=Summary of attacks against \textit{super-resolution}:]
Against \textit{super resolution}, \textit{adversarial examples} is the only attack, focusing on disrupting the model's output by perturbing the input, requiring complete knowledge of the model to perform the attack. The attack focuses mainly on using \textit{distance metrics} to measure the difference between clean and adversarial input and output.
  \end{tcolorbox}

\section{Results of RQ2: defenses methods against attacks}
\begin{figure}[ht]
    \centering
    \begin{minipage}{0.48\textwidth}
        \centering
        \includegraphics[width=\textwidth]{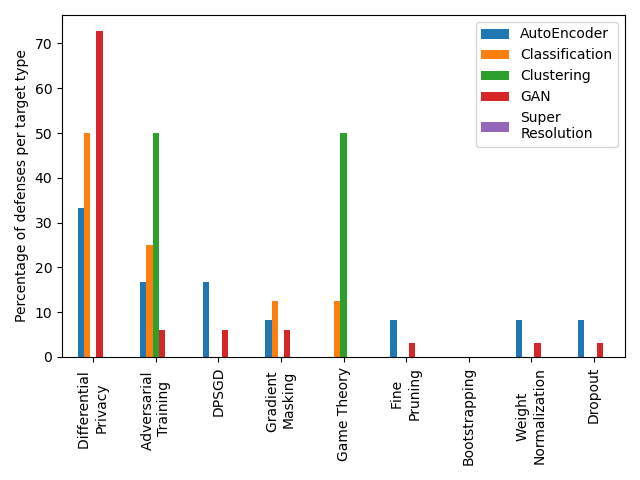}
        \caption{Distribution of the different types of defenses found in the review}
        \label{fig:defenses_percent}
    \end{minipage}
    \hfill
    \begin{minipage}{0.48\textwidth}
        \centering
        \includegraphics[width=\textwidth]{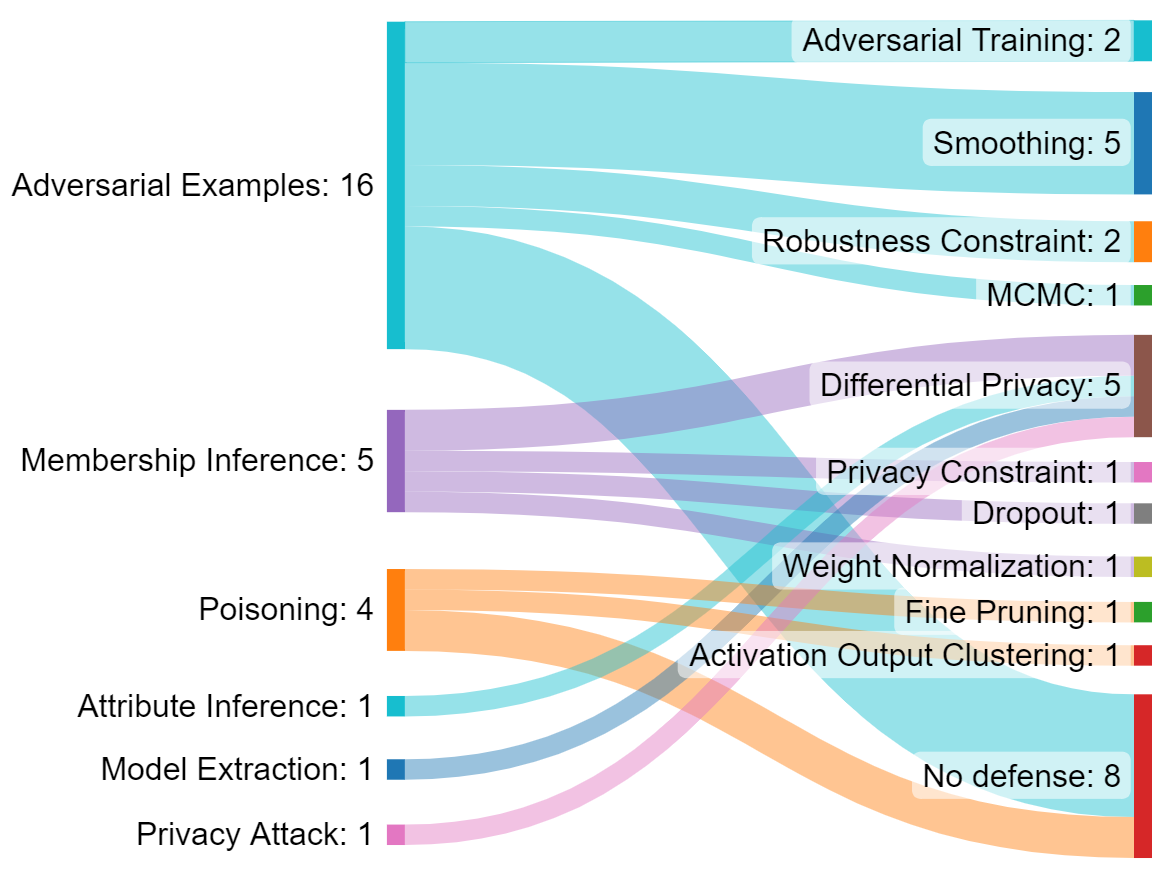}
        \caption{Sankey diagram showing the relative appearance between attack types and defenses for autoencoders}
        \label{fig:autoencoder_attacktype_defenses_sankey}
    \end{minipage}
\end{figure}

A summary of the percentage of papers with defenses used by different target types is covered in Figure \ref{fig:defenses_percent}, which shows the proportion of defense types for each target type, e.g., \textit{differential privacy} contributes ~33\%, 50\%, and ~72\%, to the total number of defenses of \textit{AutoEncoder}, \textit{classification}, and \textit{GAN}, respectively. Details on the different \textit{defenses} can be found in Appendix \ref{sec:description_defenses}.

\subsection{Defenses against attacks targeting AutoEncoder}
A summary of the relationship between the attack type and defensive measure for autoencoders is shown in Figure \ref{fig:autoencoder_attacktype_defenses_sankey}.

\textit{Adversarial training} consists of using adversaries' strategy to generate samples to be learned by the model, making the model more robust against the specific attack used to generate the samples. This defense has certain drawbacks, like reducing performance on clean samples \cite{17, 21} or lacking generalizability on other perturbations \cite{21}.

\textit{Smoothing}, also referred to as \textit{gradient masking}, works by preventing adversaries from inferring information on the internal gradients of the model \cite{17, FW1_15, FW1_45, FW1_46, FW1_50, FW2_26}, providing noticeable robustness increase against \textit{adversarial examples} in \textit{white-box} settings \cite{17}, and marginal robustness increase in \textit{black-box} settings \cite{17}. 

\textit{Robustness constraints} work by enforcing a guarantee on the lower bounds of when perturbations in the input would affect the output. In particular, this is targeted as a general defense against \textit{adversarial examples} \cite{FW1_30, FW1_52}. 

\textit{Markov Chain Monte Carlo} (MCMC) is employed on the latent code from the encoder, where the variables are updated before being used by the decoder \cite{FW2_16}. This method provides robustness against \textit{adversarial examples} and higher quality data than \textit{autoencoder} under attack. 

\textit{Differential privacy} ensures that, under certain constraints, an adversary cannot determine if a specific sample was used in training, providing slight \cite{02, FW1_25} to large \cite{FW1_53} amount of privacy protection, depending on the privacy budget \cite{FW1_53}. Attacks like \textit{membership inference} \cite{02, FW1_15, FW1_53}, \textit{attribute inference} \cite{FW1_15}, and \textit{model extraction} \cite{FW1_15} can be mitigated by using this method. Differential privacy is also used as a way of ensuring the privacy of synthetic data \cite{FW1_51}. However, this method requires more data resources and causes a decrease in the quality of the data generated \cite{02, FW1_25, FW1_51, FW1_53}.

\textit{Privacy Constraints} is another method for ensuring the privacy of training data when releasing synthetic data \cite{FW1_25}. Here though, differential privacy is implemented with a causal model, where the utility of the generated samples is higher than non-causal differential privacy implementations for the same privacy budget. 

\textit{Dropout} works on neural networks by dropping random neurons during training to prevent the model from overfitting the data. While the method is effective against \textit{membership inference} \cite{BW1_10, FW1_15}, it also results in a quality decrease of generated samples and requires more computer resources \cite{BW1_10, FW1_15}.

\textit{Weight normalization} works by periodically re-parameterizing the neural network weights. In turn, the re-parametrizing can increase the model's generalizability and provide some protection against \textit{membership inference} attacks \cite{FW1_15}.

\textit{Fine pruning} works in neural networks by removing specific neurons that are not improving the performance on clean samples. The method is not, however, proven effective for autoencoders, as "fine-pruning does not remove the by-product task" introduced by \textit{poisoning} attacks \cite{FW1_11, FW1_15}. Additionally, fine pruning also decreases the quality of generated samples and requires more computing resources \cite{FW1_11, FW1_15}.

\textit{Activation Output Clustering} works by analyzing a hidden layer of a neural network to discover different distributions between clean and adversarial input. The defense does not work for generative models. However, the strategy struggles to differentiate between real and adversarial input \cite{FW1_11, FW1_15}.

Lastly, several papers do not describe defensive measures against their respective adversarial attack, like \textit{adversarial examples} \cite{10, 16, BW1_2, FW1_3, FW2_14, FW2_14, FW2_15}, and \textit{poisoning} \cite{FW1_33, FW2_24}.

\subsection{Defenses against attacks targeting classification}
\textit{Adversarial training} for defending UL-supported classification is done by adversarially training an \textit{autoencoder} and the classifier \cite{17}, with adversarial robustness increasing or decreasing depending on the training set. \citet{17} conclude that adversarial robustness using an \textit{autoencoder} and \textit{adversarial training} provide the same effect, differing from the robustness provided by \textit{gradient masking}.

\textit{Relabeling} entails calculating the influence a specific sample would have on the training in case of \textit{poisoning} \cite{FW1_35}, and re-labeling the sample, in turn decreasing the performance of the \textit{poisoning attack}.

\subsection{Defenses against attacks targeting clustering}
\begin{figure}[ht]
    \centering
    \begin{minipage}{0.48\textwidth}
        \centering
        \includegraphics[width=\textwidth]{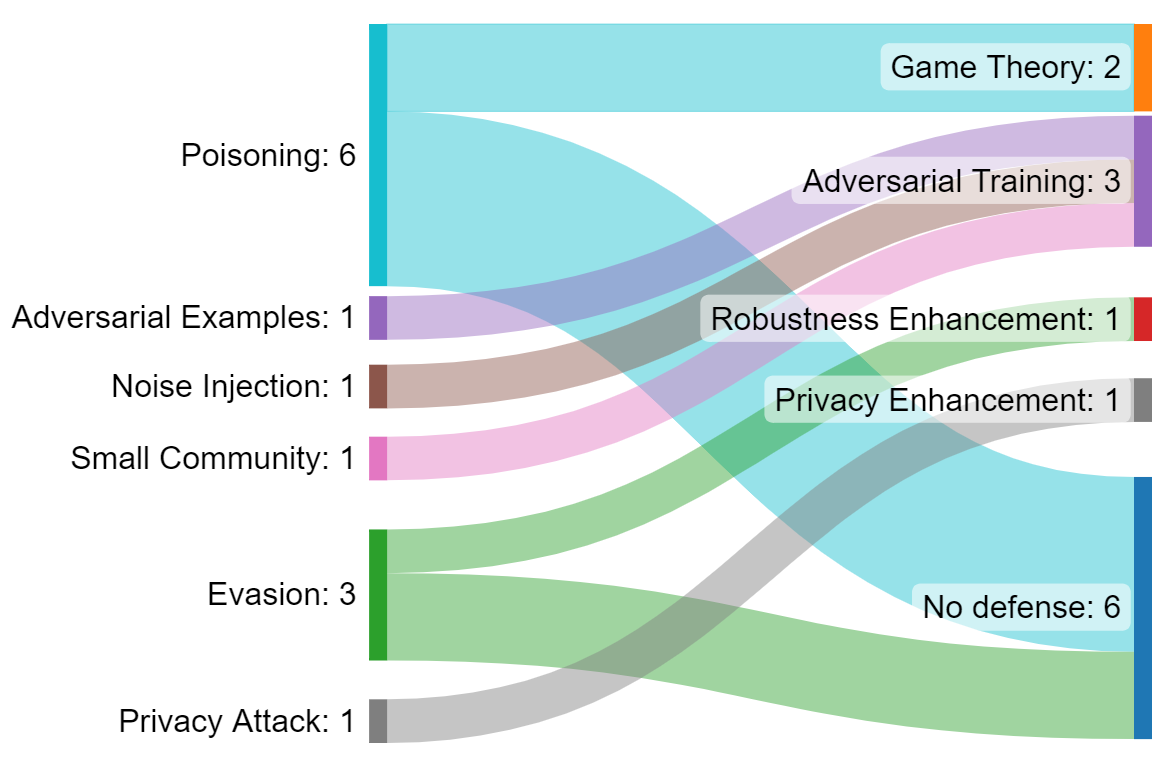}
        \caption{Sankey diagram showing the relative appearance between attack types and defenses for clustering}
        \label{fig:clustering_attacktype_defenses_sankey}
    \end{minipage}
    \hfill
    \begin{minipage}{0.48\textwidth}
        \centering
        \includegraphics[width=\textwidth]{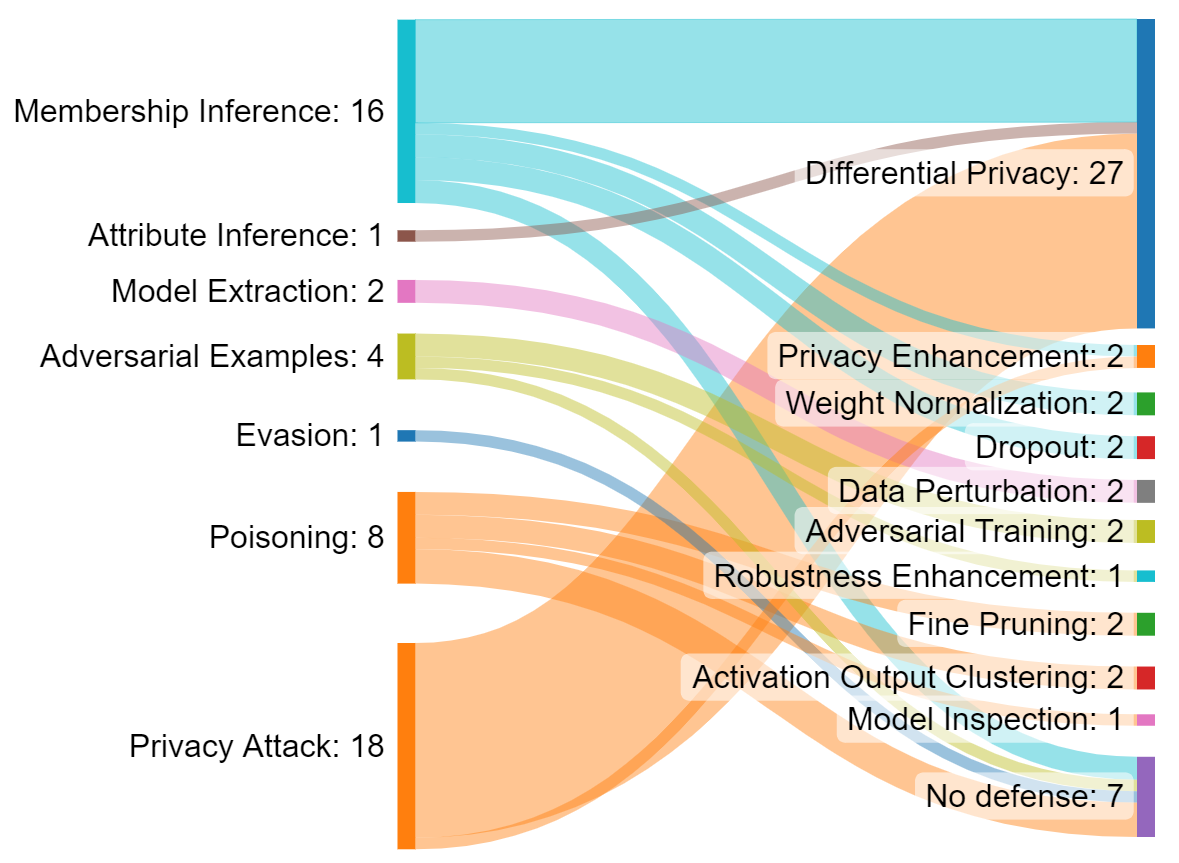}
        \caption{Sankey diagram showing the relative appearance between attack types and defenses for GANs}
        \label{fig:gan_attacktype_defenses_sankey}
    \end{minipage}
\end{figure}

A summary of the relationship between the attack type and defensive measure for clustering is shown in Figure \ref{fig:clustering_attacktype_defenses_sankey}.

\textit{Game theory} is proposed as a general defensive strategy for clustering. The general strategy involves the target employing defensive measures depending on specified conditions \cite{12} or modeling the attacker-target system as an adversarial game to determine optimal defenses \cite{13} for a variety of attacks.

\textit{Adversarial training} is employed with adversarial strategies like \textit{adversarial examples} \cite{FW1_10}, \textit{noise injection} \cite{FW1_23}, and \textit{small community} \cite{FW1_23}, which ensures the clustering trains to be more robust against these specific attacks. The drawback of this method, however, is the potential reduction of the overall \textit{accuracy} of the models \cite{FW1_23}, though some results indicate that \textit{adversarial training} does improve the performance \cite{FW1_10}.

\textit{Robustness enhancement} constructs feature sets for the clustering consisting of binaries for important features and numerical values for secondary features \cite{FW1_9}, preventing adversarial attacks from merging clusters. 

\textit{Privacy enhancement} applies privacy constraints to the clustering process for the anonymity of samples belonging to a specific cluster to be ensured \cite{BW2_9}.

Lastly, several papers do not describe defensive measures against adversarial attacks, like \textit{poisoning} \cite{14, 15, FW1_6} and \textit{evasion} \cite{13, FW1_4}.

\subsection{Defenses against attacks targeting GAN}
A summary of the relationship between the attack type and defensive measure for GANs is shown in Figure \ref{fig:gan_attacktype_defenses_sankey}.

\textit{Differential privacy} is used as a defensive method in 27 papers. The research in \cite{BW1_4, BW1_5, BW1_6, BW1_11, BW1_14, BW1_15, FW1_17, FW1_29, FW1_38, FW1_51, BW2_3, BW2_7, FW2_3, FW2_8, FW2_11, FW2_19, FW2_29, FW2_30} all cover theoretical applications of differential privacy. The defense is also applied specifically to counter \textit{membership inference} in \cite{02, 03, BW1_8, BW1_9, FW1_2, FW1_15, FW1_20, FW1_42, FW1_43}, providing slight \cite{FW1_42, 02, FW1_20} to large \cite{BW1_8, FW1_2, FW1_20} privacy protection, depending on the privacy budget \cite{BW1_9, FW1_2, FW1_20} and training set size \cite{BW1_8}. Differential privacy can also cause quality decreases in generated samples \cite{BW1_9, FW1_2} and cause groups of data to become over-represented by the generator \cite{FW1_2}.

\textit{Privacy Enchancement} are methods for ensuring the privacy of training samples without implementing differential privacy. In \cite{FW1_16}, the model trains to generate samples independent of the set of samples used for training to protect the model from \textit{membership inference} attacks. In \cite{FW1_40}, the model trains the discriminator to determine whether two samples belong to the same identity, which, through the adversarial min-max game between the generator and the discriminator, forces the generator to produce private samples. 

\textit{Weight normalization} recalculates the parameters of the weight in the model in order to improve generalizability \cite{FW1_15}. While the method can reduce the accuracy of \textit{memberhsip inference} attacks \cite{BW1_9, FW1_15}, it also can result in instability between the discriminator and the generator \cite{BW1_9, FW1_15}.

\textit{Dropout} works by removing random neurons in a neural network to prevent overfitting by the model and might be viable for GANs, as it decreases the accuracy of \textit{membership inference} attacks \cite{BW1_9, BW1_10}, but at the same time reduces the quality of the generated samples \cite{BW1_9, BW1_10, FW1_15}.

\textit{Data perturbation} is a method to perturb the output data produced by the model in order to protect against \textit{model extraction} attacks \cite{FW1_15, FW1_27}, by adding different kinds of noise, e.g., Gaussian Filtering and JPEG compression.

\textit{Adversarial training} is used by \citet{18}, with an additional classifier to increase the generator's robustness and generalizability. In contrast, \citet{FW1_32} find other research covering adversarial training for the discriminator and how adversarially trained GAN can be robust against perturbations.

\textit{Robustness enhancement} is a method for forcing the generator to generate samples within the target space, in particular samples under adversarial and non-adversarial noise \cite{FW1_8}, in order to avoid lousy quality data from being generated under noise.

\textit{Fine pruning} works by removing neurons in a neural network that does not provide good quality generated samples but did not remove the introduced by-products from \textit{poisoning} attacks and decreased the performance of the model \cite{FW1_15}.

\textit{Activation Output Clustering} analyzes one of the layers of a neural network and clusters the values from these, with the hope of differentiating between actual and malicious data from a \textit{poisoning} attack \cite{FW1_11}.

\textit{Inspection} relies on analyzing the model parameters, i.e., \textit{static model inspection} \cite{FW1_13}; analyzing the model behavior in action, i.e., \textit{dynamic model inspection} \cite{FW1_13}; or analyzing the output of the model to compare with known adversarial data \cite{FW1_13}. The defensive measure appears unreliable in analyzing the output, and model inspection methods require knowledge of the adversary to be effective.

Lastly, some papers cover adversarial attacks that are not covered by any defensive measures, like \textit{membership inference} \cite{FW1_54, FW2_17}, \textit{poisoning} \cite{24, FW2_13,  FW2_25}, and \textit{adversarial examples} \cite{FW2_2}.

\subsection{Defenses against attacks targeting super-resolution:}
As shown in Section \ref{sec:attributes_super_res}, all attacks on \textit{super-resolution} consists of \textit{adversarial examples}. 

\textit{Data pre-processing} is one such defense, covered by \citet{FW1_39}, where the data size is reduced by one pixel and subsequently resized to the original size, resulting in an increase of the measured \textit{PSNR} on the adversarial input.

\textit{Ensamble} methods covered by \citet{FW1_39} use a \textit{geometric self-ensemble} method, where the original images are geometrically transformed and passed through the model, subsequently inversely transformed and combined to create the final upscaled image. This method increases the adversarial inputs measured by \textit{PSNR}.

\textit{Adversarial training} is performed by including \textit{adversarial examples} in the re-training of the super-resolution model, which increases model generalizability, with the model performing better against \textit{adversarial examples} \cite{FW2_7, FW2_37}.

\section{Results of RQ3: Challenges to secure UL-based systems}

\subsection{Challenges to secure systems using AutoEncoders}
As autoencoders are applicable for several different purposes, there are also different challenges depending on the use. For generative autoencoders, \textit{privacy} \cite{02, BW1_10, FW1_25, FW1_53} attack is a problem, particularly when an adversary has access to the model parameters \cite{02}. Though there are defenses like \textit{differential privacy} \cite{02, FW1_51, FW1_53}, \textit{fine pruning} \cite{FW1_11}, \textit{activation output clustering} \cite{FW1_11}, the impact on generated samples is notable. Autoencoder systems for encoding/decoding, or noise reduction, are also vulnerable, particularly against \textit{adversarial examples} in the input \cite{BW1_2, FW1_2}, or latent space \cite{10, FW1_33}. Defensive measures for these kinds of attacks on autoencoders are also not very well researched \cite{21}.

\subsection{Challenges to secure UL supported classifications}
We can divide the adversarial robustness of classification into two groups. Firstly is the use of UL systems to provide adversarial robustness of the classifier, e.g., using an \textit{autoencoder} in front of the classifier \cite{17}. Secondly is the exploration of adversarial robustness in systems employing UL in tandem with classification, e.g., using \textit{clustering} to separate malicious and legitimate data for classification \cite{12}, training classifiers on generated samples \cite{24, FW1_29, FW1_40, FW1_42, FW1_53, FW2_8, FW2_11}, or employing a classifier in addition to discriminator in a \textit{GAN} \cite{18}. In this case, while there exist several attacks, like \textit{poisoning} \cite{19, 24, FW1_35, FW1_47}, defensive measures in the literature is mainly theoretical, like \textit{game theory} \cite{12}.

\subsection{Challenges to secure clustering}
While clustering as a concept relies on finding optimal groups of samples based on some distance metric, how the distance is calculated impacts the vulnerabilities and adversarial robustness of clustering. It is speculated that the specific \textit{poisoning} attacks by \citet{13} might only apply to \textit{single-linkage hierarchical} clustering, while other clustering methods would not. Additionally, known vulnerabilities of clustering are still not explored enough in the literature, as most attacks focus on specific clustering methods \cite{13, 14, 15, BW2_9}. Defending clustering methods is also not very well explored \cite{13}, with methods like \textit{adversarial simulation} \cite{13} and general privacy constraints \cite{BW2_9} left as future work in the surveyed papers. 

\subsection{Challenges to secure GAN-based systems}
Adversarial robustness for GAN models is covered by various papers, with many limitations. One of the major limitations is the compromise between GAN utility and the implementation of defensive measures. \textit{Differential privacy} provides privacy for training samples \cite{02, 03, BW1_8, BW1_9, BW1_14, FW1_15, FW1_17, FW1_29, FW1_38, FW1_42, FW1_51, BW2_3, BW2_7, FW2_8, FW2_11, FW2_19, FW2_30}; and \textit{dropout} \cite{BW1_9, BW1_10, FW1_15} and \textit{weight normalization} \cite{BW1_9, FW1_15} reduce the effectiveness of \textit{membership inference attacks}. Simultaneously, the defenses increase computational cost \cite{02, BW1_8, BW1_9, FW1_15, FW2_11}, lower the quality of the generated data \cite{02, 03, BW1_4, BW1_8, BW1_9, BW1_10, BW1_14, FW1_15, FW1_16, FW1_27, FW1_38, FW1_40, FW1_42, FW1_51, BW2_3, FW2_11, FW2_19}, lower the stability of the model \cite{BW1_4, BW1_5, BW1_14, FW1_15, BW2_3, FW2_2, FW2_8}, and is not effective for other kinds of inference attacks \cite{03, FW1_20}. When it comes to \textit{poisoning} attacks, defensive measures that work with discriminatory models, e.g., \textit{fine pruning} and \textit{activation output clustering}, do not work to prevent or detect \textit{poisoning} attacks against generative models like GAN \cite{FW1_11}. Other solutions like \textit{output inspection} \cite{FW1_13} rely on knowledge of the \textit{poisoning} attack and would be impractical to implement outside particular situations. Additionally, the notion of generalization \cite{02, BW1_8, BW1_9, BW1_10, FW1_2, FW1_15, FW1_16, FW1_20, FW1_38, FW1_43, FW2_17} and diversity \cite{BW1_5, BW1_9, FW1_2, FW1_17, FW1_20, FW2_17} in generated data is discussed as underlying properties of GAN that influence the effectiveness of attacks, e.g., when the GAN produces diverse but non-generalized samples, the model is vulnerable to \textit{membership inference} \cite{02, BW1_8, BW1_9, BW1_10, FW1_2, FW1_15, FW1_16, FW1_20, FW1_43, FW2_17}. In the literature, attacks like \textit{data injection} \cite{FW1_15} and defenses like \textit{data augmentation} \cite{FW1_11, FW1_15} are not explored but mentioned as potential future work. No single defensive method can defend against several attacks simultaneously \cite{FW1_32}.

\subsection{Challenges to secure super-resolution-based systems}
In the papers covering adversarial robustness of super-resolution models \cite{FW1_39, FW2_7, FW2_37}, the papers do not cover the limitations of the work, other than "more advanced defense methods can be investigated in the future work" \cite{FW1_39} about defenses against \textit{adversarial examples} attacks.

\section{Discussion}
\subsection{Comparison with related work}
\citet{liu2018survey}, \citet{guan2018machine}, and \citet{zhang2020privacy} do not outline their method of discovering the list of papers used for their analysis, and \citet{martins2020adversarial} only outline their search engines and keywords. This study follows a more SLR systematic approach, and the papers covered by this study are, therefore, more complete and representative of the state of the field.

Table \ref{tab:comparison} compares the papers regarding the number of papers and which defenses were covered. Several of the defenses are covered both by the other surveys and this one, like \textit{adversarial training}, \textit{smoothing/gradient masking}, and \textit{differential privacy}. However, this paper does cover a multitude of papers related to each defense, while most of the other papers cover only one or a few papers on the same defensive measure. The exception is \citet{zhang2020privacy}, which covers \textit{differential privacy} with several papers.
The most notable difference between this paper, and the other papers, is that this paper has not found any defenses using \textit{homomorphic encryption}, which can be explainable by 1) this SLR did not craft its search well enough to include \textit{homomorphic encryption}, 2) unsupervised learning does not work well with \textit{homomorphic encryption}, or 3) \textit{homomorphic encryption} has not yet been widely used as a defensive measure for unsupervised learning. Doing a surface search on \textit{Google Scholar} on \textit{homomorphic encryption} with unsupervised learning does not reveal a large number of papers on this subject, which leads us to believe that number three is the most likely answer as to why it does not show up in this SLR. With this, research into defensive measures for UL using \textit{homomorphic encryption} might be a viable future endeavor.
Other defensive measures that are present in the other papers but not in ours, like \textit{reject on negative impact}, \textit{defensive distillation}, \textit{universal perturbation defense method}, do not appear to be applicable in an unsupervised scenario, as they are only applicable for classification.
Lastly, defenses like \textit{transferability block}, \textit{MagNet}, \textit{reverse poisoning}, \textit{query distribution detection}, \textit{hierarchical model stacking}, \textit{alternating training}, \textit{defensive perturbation}, and \textit{layer separation} appear on the surface to be applicable in an unsupervised scenario, and in future research should be evaluated to find out whether or not they work for UL.

\begin{table}[ht]
    \caption{Comparison between surveys on adversarial robustness in machine learning. Cross (✗) shows that the column label is not applicable, Tick (✓) shows that the column label does apply, and Asterisk ($\ast$) shows that the column label partially applies.}
    \centering
    \begin{tabular}{|p{0.04\textwidth}|p{0.15\textwidth}|p{0.07\textwidth}|p{0.64\textwidth}|}
    \hline
    \textbf{Ref.} & Short description & \textbf{Num. papers} & \textbf{Defenses Covered} \\
    \hline
    \cite{liu2018survey} & Adversarial Examples attacks against Machine Learning & 42 & 
    Reject on Negative Impact \cite{nelson2009misleading}, Adversarial Training \cite{goodfellow2014explaining}, Defense Distillation \cite{papernot2016distillation, carlini2016defensive}, Ensemble method \cite{sengupta2018mtdeep, tramer2017ensemble, abbasi2017robustness}, Differential Privacy \cite{dwork2006differential, abadi2016deep, kusner2015differentially}, Homomorphic Encryption \cite{damgaard2012multiparty, yao2017investigation} \\
    \hline
    \cite{martins2020adversarial}& Adversarial Examples against detection techniques & 20 & 
    Adversarial Training \cite{goodfellow2014explaining, sankaranarayanan2018regularizing}, Gradient Masking \cite{duddu2018survey, papernot2016towards, papernot2017practical}, Defensive Distillation \cite{papernot2016distillation}, Feature Squeezing \cite{xu2017feature}, Transferability Block \cite{hosseini2017blocking}, Universal Perturbation Defense Method \cite{akhtar2018defense}, MagNet \cite{meng2017magnet} \\
    \hline
    \cite{guan2018machine} & Impact of ML on security, and robustness of AI systems & 14 & 
    Reject on Negative Impact \cite{nelson2009misleading}, Learning Algorithm Improvement \cite{liu2017robust, baracaldo2017mitigating}, Adversarial Retraining, Output Smoothness \cite{gu2014towards}, Defensive Distillation \cite{papernot2016distillation, carlini2016defensive}, Differential Privacy \cite{bittner2017differentially}, Homomorphic Encryption \cite{damgaard2012multiparty} \\
    \hline
    \cite{zhang2020privacy} & Privacy threats and mitigations within ML & 45 & 
    Reverse Poisoning \cite{orekondy2020prediction},  Query Distribution Detection \cite{juuti2019prada}, Hierarchical Model Stacking \cite{salem2018ml}, Alternating Training \cite{nasr2018machine}, Defensive Perturbation \cite{jia2019memguard, jia2018attriguard}, Collaborative Training \cite{shokri2015privacy, mcmahan2017communication}, Homomorphic Encryption \cite{rivest1978data, gentry2009fully, bansal2011privacy, zhang2015privacy, gilad2016cryptonets, chabanne2017privacy, liu2017oblivious} ,  Differential Privacy \cite{dwork2006calibrating, bae2018security, shokri2015privacy, abadi2016deep, phan2016differential, phan2017adaptive, papernot2016semi} , Layer Separation \cite{osia2020hybrid, chi2018privacy} \\
    \hline
    This work & A systematic review on adversarial robustness of unsupervised machine learning & 86 & 
    Adversarial training \cite{17, 18, 21, FW1_10, FW1_23, FW1_32, FW2_7, FW2_37}, Differential Privacy \cite{02, 03, BW1_4, BW1_5, BW1_6, BW1_8, BW1_9, BW1_11, BW1_14, BW1_15, FW1_2, FW1_15, FW1_17, FW1_20, FW1_25, FW1_29, FW1_38, FW1_42, FW1_43, FW1_51, FW1_53, BW2_3, BW2_7, FW2_3, FW2_8, FW2_11, FW2_19, FW2_29, FW2_30}, Dropout \cite{BW1_9, BW1_10, FW1_15}, Fine Pruning \cite{FW1_15}, Smoothing \cite{17, FW1_15, FW1_45, FW1_46, FW1_50, FW2_26}, Weight Normalization \cite{BW1_9, FW1_15},  Robustness Constraints \cite{FW1_30}, Activation Output Clustering \cite{FW1_11, FW1_15},  MCMC \cite{FW2_16}, Relabeling \cite{FW1_35}, Game Theory \cite{12}, Robustness Enhancement \cite{FW1_9}, Privacy Enhancement \cite{BW2_9}, Inspection \cite{FW1_13}, Data pre-processing \cite{FW1_39} , Ensamble \cite{FW1_39} \\
    \hline
    \end{tabular}
    
    \label{tab:comparison}
\end{table}

\subsection{Implications to academia}
Our study gives a structured and comprehensive summary of the state-of-the-art adversarial robustness of unsupervised learning and how this can be used for further research. \textit{Attack knowledge} is one of the core parts of the attack, and while specific attacks like \textit{adversarial examples} and \textit{membership inference} can work with minimal knowledge, all attacks perform noticeably better with an increase in knowledge of the system. In this regard, our paper aims at structuring and summarizing the performance of attacks and defenses depending on the amount of knowledge, which could be used in future research to make comparisons easier. Specific attacks also employ substitute models trained to mimic the actual model and perform full-knowledge attacks on these, which can transfer to the actual model. Future research should therefore consider a broader range of knowledge scenarios in their analysis, as a \textit{black box} attack could be turned into a \textit{white box} attack using such methods.

Another critical attack aspect is the \textit{attack goal}. On the one hand, our paper shows that most attacks focus on infringing on the \textit{privacy} of the target, which in turn motivates future research into finding methods for ensuring privacy since this is the most popular attack vector. Additionally, we believe our findings that \textit{differential privacy} impact model performance should motivate future research into providing privacy guarantees with less impact on performance. On the other hand, our paper also shows that \textit{integrity} attacks are effective. However, their defenses are not given as much focus in the research, which is something we wish to motivate further research in doing.

Future research would also benefit from this paper concerning the kinds of attacks leveraged against specific technologies. We can, for example, see that \textit{poisoning} attacks target most technologies; or that there are no \textit{privacy} attacks on \textit{clustering}, which future research could use as motivation for studying these topics.

We also believe the summaries on \textit{metrics} can enable future research with both motivations for selecting specific metrics and categorizing which metrics are used by which papers, making comparisons between papers easier.

Regarding \textit{defenses}, we believe our paper provides future research with several significant findings. We categorize how each defensive measure is used, for which technology, and against which attack they are leveraged. We hope to motivate future research to utilize this survey in finding defenses and make for easier comparisons when researching defensive measures. We have also presented how each defense performs in terms of effectiveness against specific attacks and, where applicable, how they impact the model's performance. In turn, we hope to motivate future research into less performance-reducing defenses. Lastly, while many papers cover complete knowledge scenarios to defend against a worst-case attack, only focusing on one level of knowledge might also be a problem, as defensive measures in different attacker knowledge scenarios are not necessarily transferable. Therefore, as mentioned in regards to \textit{attack knowledge}, we believe future research should consider how the defense is impacted when employed against attacks in different knowledge scenarios.

One of the main challenges facing adversarially robust UL is the general need for more focus on constructing defensive measures against attacks. Much research focuses primarily on the vulnerability of a specific model without formulating defenses. Additionally, when research does focus on defensive measures, it is mostly against specific attacks. While exploring specific defenses is good, there is a noticeable lack of more general defensive measures, with the big exception of \textit{differential privacy}, which has seen the most focus out of any defense in UL. We, therefore, believe that future research would benefit from focusing on more generalizable defenses.\\

We believe our model in Figure \ref{fig:model_3part} of analyzing adversarial robustness, i.e., looking at \textit{target type}, \textit{attack method}, \textit{attack knowledge}, \textit{attack goal}, \textit{attack focus}, \textit{metrics}, and \textit{defenses} can be used in future research as a guide for what is essential to cover in this kind of research, in order to structure knowledge and make comparisons of results between papers easier. Future research also needs to be performed in a manner that enables easy comparisons between results by explicitly defining these different categories.

\subsection{Implication to industry}
Those in the industry who utilize any of the technologies presented, e.g., clustering for anomaly detection or synthetic data generation, can use the results in this paper to better evaluate their use of this technology, e.g., in risk assessment. Additionally, since a significant part of this paper explicitly outlines the state-of-the-art concerning defensive measures for these technologies, this paper would be a good resource for finding and implementing defensive measures to protect their UL models. Lastly, our focus on finding vulnerabilities in these technologies could motivate the industry to consider the security of their models and not just use unsupervised models for their utility.

\subsection{Threats to validity}
We strictly followed the SLR guideline to ensure the method was unambiguous and considered biases that could influence the results' quality. However, we could still identify some possible validity threats related to this study. 

The researchers' knowledge of adversarial robustness in unsupervised learning would inevitably impact the search query. It would impact the specific words and the decision for when the query is good enough. The researchers have tried to mitigate this impact by 1) Documenting the specific papers that served as a basis for what the search should produce and 2) The initial search query would only be accepted if it produced between 1000 and 10 000 papers. 

Another possible threat to validity is the research databases used for finding papers. This study covered as many relevant databases as possible, including IEEE Xplore, ACM Digital Library, arXiv, and Scopus.  

When filtering the papers, for the first round of filtration, where papers were excluded based on the title alone, the researchers had to decide to exclude instead of include a paper explicitly. The reasoning is that deciding to exclude requires more thought than an inclusive decision, making the researcher less likely to exclude relevant papers. From the second round onward, the researchers documented their reasoning for exclusion or inclusion and produced documentation that could be scrutinized. 

Regarding the quality assurance of the papers, there are also several possible biases to consider. First, there is the bias in choosing which categories contribute to the overall quality score. For this, we considered four different papers \cite{keele2007guidelines, kitchenham2015evidence, dybaa2008strength, ivarsson2011method} for constructing a list of questions. Of these, \cite{keele2007guidelines} and \cite{kitchenham2015evidence} appeared most relevant. We compared these two and concluded that neither paper was better than the other, ultimately producing the 11 questions for the QA by excluding one irrelevant question from \cite{kitchenham2015evidence}. 
Secondly, there is a bias in how the papers were graded based on these questions. To combat this, after the QA of the remaining 28 papers in the initial round, we conducted a kappa analysis on the scoring between the two authors to indicate whether the scoring was externally valid. The two authors scored a random sample of eight papers, and a weighted kappa score was calculated, producing a weighted kappa score of 0.68.

\section{Conclusion and future work}
The existing literature on adversarial robustness in machine learning needs more coverage on unsupervised machine learning technologies. We have therefore performed a systematic literature review to summarize adversarial attacks, defenses, and research gaps related to unsupervised machine learning. From an initial list of 9129 papers, we have identified a list of 86 primary studies. We summarized the attack types, properties of targeted systems, metrics used for attack success, and the impact of the attacks. For each \textit{target type}, we summarized and analyzed the papers in relation to 1) \textit{attack type}, \textit{attack goal}, \textit{attack focus}, \textit{attack knowledge}, and \textit{metrics}, 2) summarized and analyzed the papers relating to \textit{defense}, and 3) summarized and analyzed the current challenges in research into adversarial robustness of UL.

One of the notable findings is that, except for \textit{differential privacy}, the field of unsupervised learning has a noticeable lack of focus on effective defensive methods, as several papers also conclude that defensive methods used in other fields of machine learning do not work in the context of unsupervised learning. Additionally, despite the fact that unsupervised learning systems are fairly diverse in function and form, the metrics used to determine whether attacks are successful are primarily shared, and many adversaries use distance metrics to determine the strength of the attack to avoid detection. Knowing this, crafting more sophisticated detection methods that would not be affected by distance-metric limitations might be worthwhile.

As mentioned in the discussion, further research into poisoning attacks is a worthwhile direction due to both the applicability of poisoning attacks against, seemingly, any unsupervised machine learning (and perhaps all types of machine learning) and due to the amount of data that is used in UL, which could contain poisoned samples since the data could be sourced from public databases or data that anyone can contribute to.

\bibliographystyle{ACM-Reference-Format}
\bibliography{samples/references}


\begin{thebibliography}{159}


\ifx \showCODEN    \undefined \def \showCODEN     #1{\unskip}     \fi
\ifx \showDOI      \undefined \def \showDOI       #1{#1}\fi
\ifx \showISBNx    \undefined \def \showISBNx     #1{\unskip}     \fi
\ifx \showISBNxiii \undefined \def \showISBNxiii  #1{\unskip}     \fi
\ifx \showISSN     \undefined \def \showISSN      #1{\unskip}     \fi
\ifx \showLCCN     \undefined \def \showLCCN      #1{\unskip}     \fi
\ifx \shownote     \undefined \def \shownote      #1{#1}          \fi
\ifx \showarticletitle \undefined \def \showarticletitle #1{#1}   \fi
\ifx \showURL      \undefined \def \showURL       {\relax}        \fi
\providecommand\bibfield[2]{#2}
\providecommand\bibinfo[2]{#2}
\providecommand\natexlab[1]{#1}
\providecommand\showeprint[2][]{arXiv:#2}

\bibitem[Abadi et~al\mbox{.}(2016)]%
        {abadi2016deep}
\bibfield{author}{\bibinfo{person}{Martin Abadi}, \bibinfo{person}{Andy Chu},
  \bibinfo{person}{Ian Goodfellow}, \bibinfo{person}{H~Brendan McMahan},
  \bibinfo{person}{Ilya Mironov}, \bibinfo{person}{Kunal Talwar}, {and}
  \bibinfo{person}{Li Zhang}.} \bibinfo{year}{2016}\natexlab{}.
\newblock \showarticletitle{Deep learning with differential privacy}. In
  \bibinfo{booktitle}{\emph{Proceedings of the 2016 ACM SIGSAC conference on
  computer and communications security}}. \bibinfo{pages}{308--318}.
\newblock


\bibitem[Abbasi and Gagn{\'e}(2017)]%
        {abbasi2017robustness}
\bibfield{author}{\bibinfo{person}{Mahdieh Abbasi} {and}
  \bibinfo{person}{Christian Gagn{\'e}}.} \bibinfo{year}{2017}\natexlab{}.
\newblock \showarticletitle{Robustness to adversarial examples through an
  ensemble of specialists}.
\newblock \bibinfo{journal}{\emph{arXiv preprint arXiv:1702.06856}}
  (\bibinfo{year}{2017}).
\newblock


\bibitem[Akhtar et~al\mbox{.}(2018)]%
        {akhtar2018defense}
\bibfield{author}{\bibinfo{person}{Naveed Akhtar}, \bibinfo{person}{Jian Liu},
  {and} \bibinfo{person}{Ajmal Mian}.} \bibinfo{year}{2018}\natexlab{}.
\newblock \showarticletitle{Defense against universal adversarial
  perturbations}. In \bibinfo{booktitle}{\emph{Proceedings of the IEEE
  Conference on Computer Vision and Pattern Recognition}}.
  \bibinfo{pages}{3389--3398}.
\newblock


\bibitem[Albaseer et~al\mbox{.}(2020)]%
        {10}
\bibfield{author}{\bibinfo{person}{Abdullatif Albaseer},
  \bibinfo{person}{Bekir~Sait Ciftler}, {and} \bibinfo{person}{Mohamed~M
  Abdallah}.} \bibinfo{year}{2020}\natexlab{}.
\newblock \showarticletitle{Performance evaluation of physical attacks against
  e2e autoencoder over rayleigh fading channel}. In
  \bibinfo{booktitle}{\emph{2020 IEEE International Conference on Informatics,
  IoT, and Enabling Technologies (ICIoT)}}. IEEE, \bibinfo{pages}{177--182}.
\newblock


\bibitem[Alfarra et~al\mbox{.}(2022)]%
        {FW2_2}
\bibfield{author}{\bibinfo{person}{Motasem Alfarra}, \bibinfo{person}{Juan~C
  P{\'e}rez}, \bibinfo{person}{Anna Fr{\"u}hst{\"u}ck},
  \bibinfo{person}{Philip~HS Torr}, \bibinfo{person}{Peter Wonka}, {and}
  \bibinfo{person}{Bernard Ghanem}.} \bibinfo{year}{2022}\natexlab{}.
\newblock \showarticletitle{On the Robustness of Quality Measures for GANs}.
\newblock \bibinfo{journal}{\emph{arXiv preprint arXiv:2201.13019}}
  (\bibinfo{year}{2022}).
\newblock


\bibitem[Ali et~al\mbox{.}(2021)]%
        {ali2021voice}
\bibfield{author}{\bibinfo{person}{A~Tahseen Ali}, \bibinfo{person}{Hasanen~S
  Abdullah}, {and} \bibinfo{person}{Mohammad~N Fadhil}.}
  \bibinfo{year}{2021}\natexlab{}.
\newblock \showarticletitle{Voice recognition system using machine learning
  techniques}.
\newblock \bibinfo{journal}{\emph{Materials Today: Proceedings}}
  (\bibinfo{year}{2021}), \bibinfo{pages}{1--7}.
\newblock


\bibitem[Ali and Shah(2005)]%
        {ali2005supervised}
\bibfield{author}{\bibinfo{person}{Saad Ali} {and} \bibinfo{person}{Mubarak
  Shah}.} \bibinfo{year}{2005}\natexlab{}.
\newblock \showarticletitle{A supervised learning framework for generic object
  detection in images}. In \bibinfo{booktitle}{\emph{Tenth IEEE International
  Conference on Computer Vision (ICCV'05) Volume 1}}, Vol.~\bibinfo{volume}{2}.
  IEEE, \bibinfo{pages}{1347--1354}.
\newblock


\bibitem[Anindya and Kantarcioglu(2018)]%
        {FW1_4}
\bibfield{author}{\bibinfo{person}{Imrul~Chowdhury Anindya} {and}
  \bibinfo{person}{Murat Kantarcioglu}.} \bibinfo{year}{2018}\natexlab{}.
\newblock \showarticletitle{Adversarial anomaly detection using centroid-based
  clustering}. In \bibinfo{booktitle}{\emph{2018 IEEE International Conference
  on Information Reuse and Integration (IRI)}}. IEEE, \bibinfo{pages}{1--8}.
\newblock


\bibitem[Anwar et~al\mbox{.}(2020)]%
        {anwar2020deep}
\bibfield{author}{\bibinfo{person}{Saeed Anwar}, \bibinfo{person}{Salman Khan},
  {and} \bibinfo{person}{Nick Barnes}.} \bibinfo{year}{2020}\natexlab{}.
\newblock \showarticletitle{A deep journey into super-resolution: A survey}.
\newblock \bibinfo{journal}{\emph{ACM Computing Surveys (CSUR)}}
  \bibinfo{volume}{53}, \bibinfo{number}{3} (\bibinfo{year}{2020}),
  \bibinfo{pages}{1--34}.
\newblock


\bibitem[Badu-Marfo et~al\mbox{.}(2020)]%
        {FW2_3}
\bibfield{author}{\bibinfo{person}{Godwin Badu-Marfo}, \bibinfo{person}{Bilal
  Farooq}, {and} \bibinfo{person}{Zachary Patterson}.}
  \bibinfo{year}{2020}\natexlab{}.
\newblock \showarticletitle{A Differentially Private Multi-Output Deep
  Generative Networks Approach For Activity Diary Synthesis}.
\newblock \bibinfo{journal}{\emph{arXiv preprint arXiv:2012.14574}}
  (\bibinfo{year}{2020}).
\newblock


\bibitem[Bae et~al\mbox{.}(2018)]%
        {bae2018security}
\bibfield{author}{\bibinfo{person}{Ho Bae}, \bibinfo{person}{Jaehee Jang},
  \bibinfo{person}{Dahuin Jung}, \bibinfo{person}{Hyemi Jang},
  \bibinfo{person}{Heonseok Ha}, \bibinfo{person}{Hyungyu Lee}, {and}
  \bibinfo{person}{Sungroh Yoon}.} \bibinfo{year}{2018}\natexlab{}.
\newblock \showarticletitle{Security and privacy issues in deep learning}.
\newblock \bibinfo{journal}{\emph{arXiv preprint arXiv:1807.11655}}
  (\bibinfo{year}{2018}).
\newblock


\bibitem[Banerjee et~al\mbox{.}(2018)]%
        {12}
\bibfield{author}{\bibinfo{person}{Nikhil Banerjee}, \bibinfo{person}{Thanassis
  Giannetsos}, \bibinfo{person}{Emmanouil Panaousis}, {and}
  \bibinfo{person}{Clive~Cheong Took}.} \bibinfo{year}{2018}\natexlab{}.
\newblock \showarticletitle{Unsupervised learning for trustworthy IoT}. In
  \bibinfo{booktitle}{\emph{2018 IEEE international conference on fuzzy systems
  (FUZZ-IEEE)}}. IEEE, \bibinfo{pages}{1--8}.
\newblock


\bibitem[Bank et~al\mbox{.}(2020)]%
        {bank2020autoencoders}
\bibfield{author}{\bibinfo{person}{Dor Bank}, \bibinfo{person}{Noam
  Koenigstein}, {and} \bibinfo{person}{Raja Giryes}.}
  \bibinfo{year}{2020}\natexlab{}.
\newblock \showarticletitle{Autoencoders}.
\newblock \bibinfo{journal}{\emph{arXiv preprint arXiv:2003.05991}}
  (\bibinfo{year}{2020}).
\newblock


\bibitem[Bansal et~al\mbox{.}(2011)]%
        {bansal2011privacy}
\bibfield{author}{\bibinfo{person}{Ankur Bansal}, \bibinfo{person}{Tingting
  Chen}, {and} \bibinfo{person}{Sheng Zhong}.} \bibinfo{year}{2011}\natexlab{}.
\newblock \showarticletitle{Privacy preserving back-propagation neural network
  learning over arbitrarily partitioned data}.
\newblock \bibinfo{journal}{\emph{Neural Computing and Applications}}
  \bibinfo{volume}{20} (\bibinfo{year}{2011}), \bibinfo{pages}{143--150}.
\newblock


\bibitem[Baracaldo et~al\mbox{.}(2017)]%
        {baracaldo2017mitigating}
\bibfield{author}{\bibinfo{person}{Nathalie Baracaldo}, \bibinfo{person}{Bryant
  Chen}, \bibinfo{person}{Heiko Ludwig}, {and} \bibinfo{person}{Jaehoon~Amir
  Safavi}.} \bibinfo{year}{2017}\natexlab{}.
\newblock \showarticletitle{Mitigating poisoning attacks on machine learning
  models: A data provenance based approach}. In
  \bibinfo{booktitle}{\emph{Proceedings of the 10th ACM workshop on artificial
  intelligence and security}}. \bibinfo{pages}{103--110}.
\newblock


\bibitem[Barrett et~al\mbox{.}(2022)]%
        {FW1_50}
\bibfield{author}{\bibinfo{person}{Ben Barrett}, \bibinfo{person}{Alexander
  Camuto}, \bibinfo{person}{Matthew Willetts}, {and} \bibinfo{person}{Tom
  Rainforth}.} \bibinfo{year}{2022}\natexlab{}.
\newblock \showarticletitle{Certifiably robust variational autoencoders}. In
  \bibinfo{booktitle}{\emph{International Conference on Artificial Intelligence
  and Statistics}}. PMLR, \bibinfo{pages}{3663--3683}.
\newblock


\bibitem[Beaulieu-Jones et~al\mbox{.}(2019)]%
        {BW1_6}
\bibfield{author}{\bibinfo{person}{Brett~K Beaulieu-Jones},
  \bibinfo{person}{Zhiwei~Steven Wu}, \bibinfo{person}{Chris Williams},
  \bibinfo{person}{Ran Lee}, \bibinfo{person}{Sanjeev~P Bhavnani},
  \bibinfo{person}{James~Brian Byrd}, {and} \bibinfo{person}{Casey~S Greene}.}
  \bibinfo{year}{2019}\natexlab{}.
\newblock \showarticletitle{Privacy-preserving generative deep neural networks
  support clinical data sharing}.
\newblock \bibinfo{journal}{\emph{Circulation: Cardiovascular Quality and
  Outcomes}} \bibinfo{volume}{12}, \bibinfo{number}{7} (\bibinfo{year}{2019}),
  \bibinfo{pages}{e005122}.
\newblock


\bibitem[Bernau et~al\mbox{.}(2022)]%
        {FW1_53}
\bibfield{author}{\bibinfo{person}{Daniel Bernau}, \bibinfo{person}{Jonas
  Robl}, {and} \bibinfo{person}{Florian Kerschbaum}.}
  \bibinfo{year}{2022}\natexlab{}.
\newblock \showarticletitle{Assessing Differentially Private Variational
  Autoencoders under Membership Inference}.
\newblock \bibinfo{journal}{\emph{arXiv preprint arXiv:2204.07877}}
  (\bibinfo{year}{2022}).
\newblock


\bibitem[Biggio et~al\mbox{.}(2014a)]%
        {14}
\bibfield{author}{\bibinfo{person}{Battista Biggio},
  \bibinfo{person}{Samuel~Rota Bul{\`o}}, \bibinfo{person}{Ignazio Pillai},
  \bibinfo{person}{Michele Mura}, \bibinfo{person}{Eyasu~Zemene Mequanint},
  \bibinfo{person}{Marcello Pelillo}, {and} \bibinfo{person}{Fabio Roli}.}
  \bibinfo{year}{2014}\natexlab{a}.
\newblock \showarticletitle{Poisoning complete-linkage hierarchical
  clustering}. In \bibinfo{booktitle}{\emph{Joint IAPR International Workshops
  on Statistical Techniques in Pattern Recognition (SPR) and Structural and
  Syntactic Pattern Recognition (SSPR)}}. Springer, \bibinfo{pages}{42--52}.
\newblock


\bibitem[Biggio et~al\mbox{.}(2013)]%
        {13}
\bibfield{author}{\bibinfo{person}{Battista Biggio}, \bibinfo{person}{Ignazio
  Pillai}, \bibinfo{person}{Samuel Rota~Bul{\`o}}, \bibinfo{person}{Davide
  Ariu}, \bibinfo{person}{Marcello Pelillo}, {and} \bibinfo{person}{Fabio
  Roli}.} \bibinfo{year}{2013}\natexlab{}.
\newblock \showarticletitle{Is data clustering in adversarial settings
  secure?}. In \bibinfo{booktitle}{\emph{Proceedings of the 2013 ACM workshop
  on Artificial intelligence and security}}. \bibinfo{pages}{87--98}.
\newblock


\bibitem[Biggio et~al\mbox{.}(2014b)]%
        {15}
\bibfield{author}{\bibinfo{person}{Battista Biggio}, \bibinfo{person}{Konrad
  Rieck}, \bibinfo{person}{Davide Ariu}, \bibinfo{person}{Christian
  Wressnegger}, \bibinfo{person}{Igino Corona}, \bibinfo{person}{Giorgio
  Giacinto}, {and} \bibinfo{person}{Fabio Roli}.}
  \bibinfo{year}{2014}\natexlab{b}.
\newblock \showarticletitle{Poisoning behavioral malware clustering}. In
  \bibinfo{booktitle}{\emph{Proceedings of the 2014 workshop on artificial
  intelligent and security workshop}}. \bibinfo{pages}{27--36}.
\newblock


\bibitem[Bittner et~al\mbox{.}(2017)]%
        {bittner2017differentially}
\bibfield{author}{\bibinfo{person}{Daniel~M Bittner}, \bibinfo{person}{Anand~D
  Sarwate}, {and} \bibinfo{person}{Rebecca~N Wright}.}
  \bibinfo{year}{2017}\natexlab{}.
\newblock \showarticletitle{Differentially Private Noisy Search with
  Applications to Anomaly Detection}. In \bibinfo{booktitle}{\emph{Proceedings
  of the 10th ACM Workshop on Artificial Intelligence and Security}}.
  \bibinfo{pages}{53--53}.
\newblock


\bibitem[Braun and Clarke(2006)]%
        {braun2006using}
\bibfield{author}{\bibinfo{person}{Virginia Braun} {and}
  \bibinfo{person}{Victoria Clarke}.} \bibinfo{year}{2006}\natexlab{}.
\newblock \showarticletitle{Using thematic analysis in psychology}.
\newblock \bibinfo{journal}{\emph{Qualitative research in psychology}}
  \bibinfo{volume}{3}, \bibinfo{number}{2} (\bibinfo{year}{2006}),
  \bibinfo{pages}{77--101}.
\newblock


\bibitem[Camuto et~al\mbox{.}(2021)]%
        {FW1_52}
\bibfield{author}{\bibinfo{person}{Alexander Camuto}, \bibinfo{person}{Matthew
  Willetts}, \bibinfo{person}{Stephen Roberts}, \bibinfo{person}{Chris Holmes},
  {and} \bibinfo{person}{Tom Rainforth}.} \bibinfo{year}{2021}\natexlab{}.
\newblock \showarticletitle{Towards a theoretical understanding of the
  robustness of variational autoencoders}. In
  \bibinfo{booktitle}{\emph{International Conference on Artificial Intelligence
  and Statistics}}. PMLR, \bibinfo{pages}{3565--3573}.
\newblock


\bibitem[Carlini and Wagner(2016)]%
        {carlini2016defensive}
\bibfield{author}{\bibinfo{person}{Nicholas Carlini} {and}
  \bibinfo{person}{David Wagner}.} \bibinfo{year}{2016}\natexlab{}.
\newblock \showarticletitle{Defensive distillation is not robust to adversarial
  examples}.
\newblock \bibinfo{journal}{\emph{arXiv preprint arXiv:1607.04311}}
  (\bibinfo{year}{2016}).
\newblock


\bibitem[Castillo et~al\mbox{.}(2021)]%
        {FW2_7}
\bibfield{author}{\bibinfo{person}{Angela Castillo},
  \bibinfo{person}{Mar{\'\i}a Escobar}, \bibinfo{person}{Juan~C P{\'e}rez},
  \bibinfo{person}{Andr{\'e}s Romero}, \bibinfo{person}{Radu Timofte},
  \bibinfo{person}{Luc Van~Gool}, {and} \bibinfo{person}{Pablo Arbelaez}.}
  \bibinfo{year}{2021}\natexlab{}.
\newblock \showarticletitle{Generalized Real-World Super-Resolution through
  Adversarial Robustness}. In \bibinfo{booktitle}{\emph{Proceedings of the
  IEEE/CVF International Conference on Computer Vision}}.
  \bibinfo{pages}{1855--1865}.
\newblock


\bibitem[Cemgil et~al\mbox{.}(2019)]%
        {FW1_46}
\bibfield{author}{\bibinfo{person}{Taylan Cemgil}, \bibinfo{person}{Sumedh
  Ghaisas}, \bibinfo{person}{Krishnamurthy~Dj Dvijotham}, {and}
  \bibinfo{person}{Pushmeet Kohli}.} \bibinfo{year}{2019}\natexlab{}.
\newblock \showarticletitle{Adversarially robust representations with smooth
  encoders}. In \bibinfo{booktitle}{\emph{International Conference on Learning
  Representations}}.
\newblock


\bibitem[Chabanne et~al\mbox{.}(2017)]%
        {chabanne2017privacy}
\bibfield{author}{\bibinfo{person}{Herv{\'e} Chabanne}, \bibinfo{person}{Amaury
  De~Wargny}, \bibinfo{person}{Jonathan Milgram}, \bibinfo{person}{Constance
  Morel}, {and} \bibinfo{person}{Emmanuel Prouff}.}
  \bibinfo{year}{2017}\natexlab{}.
\newblock \showarticletitle{Privacy-preserving classification on deep neural
  network}.
\newblock \bibinfo{journal}{\emph{Cryptology ePrint Archive}}
  (\bibinfo{year}{2017}).
\newblock


\bibitem[Chander and Vijaya(2021)]%
        {chander2021unsupervised}
\bibfield{author}{\bibinfo{person}{Satish Chander} {and} \bibinfo{person}{P
  Vijaya}.} \bibinfo{year}{2021}\natexlab{}.
\newblock \showarticletitle{Unsupervised learning methods for data clustering}.
\newblock In \bibinfo{booktitle}{\emph{Artificial Intelligence in Data
  Mining}}. \bibinfo{publisher}{Elsevier}, \bibinfo{pages}{41--64}.
\newblock


\bibitem[Chandrasekaran et~al\mbox{.}(2021)]%
        {FW1_25}
\bibfield{author}{\bibinfo{person}{Varun Chandrasekaran},
  \bibinfo{person}{Darren Edge}, \bibinfo{person}{Somesh Jha},
  \bibinfo{person}{Amit Sharma}, \bibinfo{person}{Cheng Zhang}, {and}
  \bibinfo{person}{Shruti Tople}.} \bibinfo{year}{2021}\natexlab{}.
\newblock \showarticletitle{Causally Constrained Data Synthesis for Private
  Data Release}.
\newblock \bibinfo{journal}{\emph{arXiv preprint arXiv:2105.13144}}
  (\bibinfo{year}{2021}).
\newblock


\bibitem[Chen et~al\mbox{.}(2020a)]%
        {BW1_15}
\bibfield{author}{\bibinfo{person}{Dingfan Chen}, \bibinfo{person}{Tribhuvanesh
  Orekondy}, {and} \bibinfo{person}{Mario Fritz}.}
  \bibinfo{year}{2020}\natexlab{a}.
\newblock \showarticletitle{Gs-wgan: A gradient-sanitized approach for learning
  differentially private generators}.
\newblock \bibinfo{journal}{\emph{Advances in Neural Information Processing
  Systems}}  \bibinfo{volume}{33} (\bibinfo{year}{2020}),
  \bibinfo{pages}{12673--12684}.
\newblock


\bibitem[Chen et~al\mbox{.}(2020b)]%
        {02}
\bibfield{author}{\bibinfo{person}{Dingfan Chen}, \bibinfo{person}{Ning Yu},
  \bibinfo{person}{Yang Zhang}, {and} \bibinfo{person}{Mario Fritz}.}
  \bibinfo{year}{2020}\natexlab{b}.
\newblock \showarticletitle{Gan-leaks: A taxonomy of membership inference
  attacks against generative models}. In \bibinfo{booktitle}{\emph{Proceedings
  of the 2020 ACM SIGSAC conference on computer and communications security}}.
  \bibinfo{pages}{343--362}.
\newblock


\bibitem[Chen et~al\mbox{.}(2022)]%
        {FW2_8}
\bibfield{author}{\bibinfo{person}{Jia-Wei Chen}, \bibinfo{person}{Chia-Mu Yu},
  \bibinfo{person}{Ching-Chia Kao}, \bibinfo{person}{Tzai-Wei Pang}, {and}
  \bibinfo{person}{Chun-Shien Lu}.} \bibinfo{year}{2022}\natexlab{}.
\newblock \showarticletitle{DPGEN: Differentially Private Generative
  Energy-Guided Network for Natural Image Synthesis}. In
  \bibinfo{booktitle}{\emph{Proceedings of the IEEE/CVF Conference on Computer
  Vision and Pattern Recognition}}. \bibinfo{pages}{8387--8396}.
\newblock


\bibitem[Chen et~al\mbox{.}(2017)]%
        {FW1_23}
\bibfield{author}{\bibinfo{person}{Yizheng Chen}, \bibinfo{person}{Yacin
  Nadji}, \bibinfo{person}{Athanasios Kountouras}, \bibinfo{person}{Fabian
  Monrose}, \bibinfo{person}{Roberto Perdisci}, \bibinfo{person}{Manos
  Antonakakis}, {and} \bibinfo{person}{Nikolaos Vasiloglou}.}
  \bibinfo{year}{2017}\natexlab{}.
\newblock \showarticletitle{Practical attacks against graph-based clustering}.
  In \bibinfo{booktitle}{\emph{Proceedings of the 2017 ACM SIGSAC conference on
  computer and communications security}}. \bibinfo{pages}{1125--1142}.
\newblock


\bibitem[Chi et~al\mbox{.}(2018)]%
        {chi2018privacy}
\bibfield{author}{\bibinfo{person}{Jianfeng Chi}, \bibinfo{person}{Emmanuel
  Owusu}, \bibinfo{person}{Xuwang Yin}, \bibinfo{person}{Tong Yu},
  \bibinfo{person}{William Chan}, \bibinfo{person}{Patrick Tague}, {and}
  \bibinfo{person}{Yuan Tian}.} \bibinfo{year}{2018}\natexlab{}.
\newblock \showarticletitle{Privacy partitioning: Protecting user data during
  the deep learning inference phase}.
\newblock \bibinfo{journal}{\emph{arXiv preprint arXiv:1812.02863}}
  (\bibinfo{year}{2018}).
\newblock


\bibitem[Choi et~al\mbox{.}(2019)]%
        {FW1_39}
\bibfield{author}{\bibinfo{person}{Jun-Ho Choi}, \bibinfo{person}{Huan Zhang},
  \bibinfo{person}{Jun-Hyuk Kim}, \bibinfo{person}{Cho-Jui Hsieh}, {and}
  \bibinfo{person}{Jong-Seok Lee}.} \bibinfo{year}{2019}\natexlab{}.
\newblock \showarticletitle{Evaluating robustness of deep image
  super-resolution against adversarial attacks}. In
  \bibinfo{booktitle}{\emph{Proceedings of the IEEE/CVF International
  Conference on Computer Vision}}. \bibinfo{pages}{303--311}.
\newblock


\bibitem[Chrysos et~al\mbox{.}(2020)]%
        {FW1_8}
\bibfield{author}{\bibinfo{person}{Grigorios~G Chrysos}, \bibinfo{person}{Jean
  Kossaifi}, {and} \bibinfo{person}{Stefanos Zafeiriou}.}
  \bibinfo{year}{2020}\natexlab{}.
\newblock \showarticletitle{Rocgan: Robust conditional gan}.
\newblock \bibinfo{journal}{\emph{International Journal of Computer Vision}}
  \bibinfo{volume}{128}, \bibinfo{number}{10} (\bibinfo{year}{2020}),
  \bibinfo{pages}{2665--2683}.
\newblock


\bibitem[Cin{\`a} et~al\mbox{.}(2022)]%
        {FW1_6}
\bibfield{author}{\bibinfo{person}{Antonio~Emanuele Cin{\`a}},
  \bibinfo{person}{Alessandro Torcinovich}, {and} \bibinfo{person}{Marcello
  Pelillo}.} \bibinfo{year}{2022}\natexlab{}.
\newblock \showarticletitle{A black-box adversarial attack for poisoning
  clustering}.
\newblock \bibinfo{journal}{\emph{Pattern Recognition}}  \bibinfo{volume}{122}
  (\bibinfo{year}{2022}), \bibinfo{pages}{108306}.
\newblock


\bibitem[Condessa and Kolter(2020)]%
        {FW1_30}
\bibfield{author}{\bibinfo{person}{Filipe Condessa} {and} \bibinfo{person}{Zico
  Kolter}.} \bibinfo{year}{2020}\natexlab{}.
\newblock \showarticletitle{Provably robust deep generative models}.
\newblock \bibinfo{journal}{\emph{arXiv preprint arXiv:2004.10608}}
  (\bibinfo{year}{2020}).
\newblock


\bibitem[Creswell et~al\mbox{.}(2017)]%
        {FW1_33}
\bibfield{author}{\bibinfo{person}{Antonia Creswell}, \bibinfo{person}{Anil~A
  Bharath}, {and} \bibinfo{person}{Biswa Sengupta}.}
  \bibinfo{year}{2017}\natexlab{}.
\newblock \showarticletitle{Latentpoison-adversarial attacks on the latent
  space}.
\newblock \bibinfo{journal}{\emph{arXiv preprint arXiv:1711.02879}}
  (\bibinfo{year}{2017}).
\newblock


\bibitem[Creswell et~al\mbox{.}(2018)]%
        {creswell2018generative}
\bibfield{author}{\bibinfo{person}{Antonia Creswell}, \bibinfo{person}{Tom
  White}, \bibinfo{person}{Vincent Dumoulin}, \bibinfo{person}{Kai
  Arulkumaran}, \bibinfo{person}{Biswa Sengupta}, {and} \bibinfo{person}{Anil~A
  Bharath}.} \bibinfo{year}{2018}\natexlab{}.
\newblock \showarticletitle{Generative adversarial networks: An overview}.
\newblock \bibinfo{journal}{\emph{IEEE signal processing magazine}}
  \bibinfo{volume}{35}, \bibinfo{number}{1} (\bibinfo{year}{2018}),
  \bibinfo{pages}{53--65}.
\newblock


\bibitem[Damg{\aa}rd et~al\mbox{.}(2012)]%
        {damgaard2012multiparty}
\bibfield{author}{\bibinfo{person}{Ivan Damg{\aa}rd}, \bibinfo{person}{Valerio
  Pastro}, \bibinfo{person}{Nigel Smart}, {and} \bibinfo{person}{Sarah
  Zakarias}.} \bibinfo{year}{2012}\natexlab{}.
\newblock \showarticletitle{Multiparty computation from somewhat homomorphic
  encryption}. In \bibinfo{booktitle}{\emph{Advances in Cryptology--CRYPTO
  2012: 32nd Annual Cryptology Conference, Santa Barbara, CA, USA, August
  19-23, 2012. Proceedings}}. Springer, \bibinfo{pages}{643--662}.
\newblock


\bibitem[Deng and Sang(2020)]%
        {16}
\bibfield{author}{\bibinfo{person}{Zhixiang Deng} {and} \bibinfo{person}{Qian
  Sang}.} \bibinfo{year}{2020}\natexlab{}.
\newblock \showarticletitle{Harnessing the Adversarial Perturbation to Enhance
  Security in the Autoencoder-Based Communication System}.
\newblock \bibinfo{journal}{\emph{Electronics}} \bibinfo{volume}{9},
  \bibinfo{number}{2} (\bibinfo{year}{2020}), \bibinfo{pages}{294}.
\newblock


\bibitem[Denyer et~al\mbox{.}(2009)]%
        {denyer2009sage}
\bibfield{author}{\bibinfo{person}{David Denyer}, \bibinfo{person}{David
  Tranfield}, \bibinfo{person}{D Buchanan}, {and} \bibinfo{person}{A Bryman}.}
  \bibinfo{year}{2009}\natexlab{}.
\newblock \showarticletitle{The Sage handbook of organizational research
  methods}.
\newblock \bibinfo{journal}{\emph{Reference and Research Book News}}
  \bibinfo{volume}{24}, \bibinfo{number}{3} (\bibinfo{year}{2009}),
  \bibinfo{pages}{776}.
\newblock


\bibitem[Devaranjan et~al\mbox{.}(2020)]%
        {devaranjan2020meta}
\bibfield{author}{\bibinfo{person}{Jeevan Devaranjan}, \bibinfo{person}{Amlan
  Kar}, {and} \bibinfo{person}{Sanja Fidler}.} \bibinfo{year}{2020}\natexlab{}.
\newblock \showarticletitle{Meta-sim2: Unsupervised learning of scene structure
  for synthetic data generation}. In \bibinfo{booktitle}{\emph{Computer
  Vision--ECCV 2020: 16th European Conference, Glasgow, UK, August 23--28,
  2020, Proceedings, Part XVII 16}}. Springer, \bibinfo{pages}{715--733}.
\newblock


\bibitem[Dihuni(2020)]%
        {bytesnum}
\bibfield{author}{\bibinfo{person}{Dihuni}.} \bibinfo{year}{2020}\natexlab{}.
\newblock \bibinfo{booktitle}{\emph{Every Day Big Data Statistics – 2.5
  Quintillion Bytes of Data Created Daily}}.
\newblock
\urldef\tempurl%
\url{https://www.dihuni.com/2020/04/10/every-day-big-data-statistics-2-5-quintillion-bytes-of-data-created-daily/}
\showURL{%
\tempurl}


\bibitem[Ding et~al\mbox{.}(2019)]%
        {FW1_11}
\bibfield{author}{\bibinfo{person}{Shaohua Ding}, \bibinfo{person}{Yulong
  Tian}, \bibinfo{person}{Fengyuan Xu}, \bibinfo{person}{Qun Li}, {and}
  \bibinfo{person}{Sheng Zhong}.} \bibinfo{year}{2019}\natexlab{}.
\newblock \showarticletitle{Trojan attack on deep generative models in
  autonomous driving}. In \bibinfo{booktitle}{\emph{International Conference on
  Security and Privacy in Communication Systems}}. Springer,
  \bibinfo{pages}{299--318}.
\newblock


\bibitem[Duddu(2018)]%
        {duddu2018survey}
\bibfield{author}{\bibinfo{person}{Vasisht Duddu}.}
  \bibinfo{year}{2018}\natexlab{}.
\newblock \showarticletitle{A survey of adversarial machine learning in cyber
  warfare}.
\newblock \bibinfo{journal}{\emph{Defence Science Journal}}
  \bibinfo{volume}{68}, \bibinfo{number}{4} (\bibinfo{year}{2018}),
  \bibinfo{pages}{356}.
\newblock


\bibitem[Dwork(2006)]%
        {dwork2006differential}
\bibfield{author}{\bibinfo{person}{Cynthia Dwork}.}
  \bibinfo{year}{2006}\natexlab{}.
\newblock \showarticletitle{Differential privacy}. In
  \bibinfo{booktitle}{\emph{Automata, Languages and Programming: 33rd
  International Colloquium, ICALP 2006, Venice, Italy, July 10-14, 2006,
  Proceedings, Part II 33}}. Springer, \bibinfo{pages}{1--12}.
\newblock


\bibitem[Dwork et~al\mbox{.}(2006)]%
        {dwork2006calibrating}
\bibfield{author}{\bibinfo{person}{Cynthia Dwork}, \bibinfo{person}{Frank
  McSherry}, \bibinfo{person}{Kobbi Nissim}, {and} \bibinfo{person}{Adam
  Smith}.} \bibinfo{year}{2006}\natexlab{}.
\newblock \showarticletitle{Calibrating noise to sensitivity in private data
  analysis}. In \bibinfo{booktitle}{\emph{Theory of Cryptography: Third Theory
  of Cryptography Conference, TCC 2006, New York, NY, USA, March 4-7, 2006.
  Proceedings 3}}. Springer, \bibinfo{pages}{265--284}.
\newblock


\bibitem[Dyb{\aa} and Dings{\o}yr(2008)]%
        {dybaa2008strength}
\bibfield{author}{\bibinfo{person}{Tore Dyb{\aa}} {and}
  \bibinfo{person}{Torgeir Dings{\o}yr}.} \bibinfo{year}{2008}\natexlab{}.
\newblock \showarticletitle{Strength of evidence in systematic reviews in
  software engineering}. In \bibinfo{booktitle}{\emph{Proceedings of the Second
  ACM-IEEE international symposium on Empirical software engineering and
  measurement}}. \bibinfo{pages}{178--187}.
\newblock


\bibitem[Fan(2020)]%
        {BW2_3}
\bibfield{author}{\bibinfo{person}{Liyue Fan}.}
  \bibinfo{year}{2020}\natexlab{}.
\newblock \showarticletitle{A survey of differentially private generative
  adversarial networks}. In \bibinfo{booktitle}{\emph{The AAAI Workshop on
  Privacy-Preserving Artificial Intelligence}}. \bibinfo{pages}{8}.
\newblock


\bibitem[Franci et~al\mbox{.}(2022)]%
        {FW1_35}
\bibfield{author}{\bibinfo{person}{Adriano Franci}, \bibinfo{person}{Maxime
  Cordy}, \bibinfo{person}{Martin Gubri}, \bibinfo{person}{Mike Papadakis},
  {and} \bibinfo{person}{Yves Le~Traon}.} \bibinfo{year}{2022}\natexlab{}.
\newblock \showarticletitle{Influence-Driven Data Poisoning in Graph-Based
  Semi-Supervised Classifiers}. In \bibinfo{booktitle}{\emph{2022 IEEE/ACM 1st
  International Conference on AI Engineering--Software Engineering for AI
  (CAIN)}}. IEEE, \bibinfo{pages}{77--87}.
\newblock


\bibitem[Fredrikson et~al\mbox{.}(2015)]%
        {fredrikson2015model}
\bibfield{author}{\bibinfo{person}{Matt Fredrikson}, \bibinfo{person}{Somesh
  Jha}, {and} \bibinfo{person}{Thomas Ristenpart}.}
  \bibinfo{year}{2015}\natexlab{}.
\newblock \showarticletitle{Model inversion attacks that exploit confidence
  information and basic countermeasures}. In
  \bibinfo{booktitle}{\emph{Proceedings of the 22nd ACM SIGSAC conference on
  computer and communications security}}. \bibinfo{pages}{1322--1333}.
\newblock


\bibitem[Frigerio et~al\mbox{.}(2019)]%
        {BW1_8}
\bibfield{author}{\bibinfo{person}{Lorenzo Frigerio}, \bibinfo{person}{Anderson
  Santana~de Oliveira}, \bibinfo{person}{Laurent Gomez}, {and}
  \bibinfo{person}{Patrick Duverger}.} \bibinfo{year}{2019}\natexlab{}.
\newblock \showarticletitle{Differentially private generative adversarial
  networks for time series, continuous, and discrete open data}. In
  \bibinfo{booktitle}{\emph{IFIP International Conference on ICT Systems
  Security and Privacy Protection}}. Springer, \bibinfo{pages}{151--164}.
\newblock


\bibitem[Gentry(2009)]%
        {gentry2009fully}
\bibfield{author}{\bibinfo{person}{Craig Gentry}.}
  \bibinfo{year}{2009}\natexlab{}.
\newblock \showarticletitle{Fully homomorphic encryption using ideal lattices}.
  In \bibinfo{booktitle}{\emph{Proceedings of the forty-first annual ACM
  symposium on Theory of computing}}. \bibinfo{pages}{169--178}.
\newblock


\bibitem[Gilad-Bachrach et~al\mbox{.}(2016)]%
        {gilad2016cryptonets}
\bibfield{author}{\bibinfo{person}{Ran Gilad-Bachrach}, \bibinfo{person}{Nathan
  Dowlin}, \bibinfo{person}{Kim Laine}, \bibinfo{person}{Kristin Lauter},
  \bibinfo{person}{Michael Naehrig}, {and} \bibinfo{person}{John Wernsing}.}
  \bibinfo{year}{2016}\natexlab{}.
\newblock \showarticletitle{Cryptonets: Applying neural networks to encrypted
  data with high throughput and accuracy}. In
  \bibinfo{booktitle}{\emph{International conference on machine learning}}.
  PMLR, \bibinfo{pages}{201--210}.
\newblock


\bibitem[Gondim-Ribeiro et~al\mbox{.}(2018)]%
        {FW1_3}
\bibfield{author}{\bibinfo{person}{George Gondim-Ribeiro},
  \bibinfo{person}{Pedro Tabacof}, {and} \bibinfo{person}{Eduardo Valle}.}
  \bibinfo{year}{2018}\natexlab{}.
\newblock \showarticletitle{Adversarial attacks on variational autoencoders}.
\newblock \bibinfo{journal}{\emph{arXiv preprint arXiv:1806.04646}}
  (\bibinfo{year}{2018}).
\newblock


\bibitem[Goodfellow et~al\mbox{.}(2020)]%
        {goodfellow2020generative}
\bibfield{author}{\bibinfo{person}{Ian Goodfellow}, \bibinfo{person}{Jean
  Pouget-Abadie}, \bibinfo{person}{Mehdi Mirza}, \bibinfo{person}{Bing Xu},
  \bibinfo{person}{David Warde-Farley}, \bibinfo{person}{Sherjil Ozair},
  \bibinfo{person}{Aaron Courville}, {and} \bibinfo{person}{Yoshua Bengio}.}
  \bibinfo{year}{2020}\natexlab{}.
\newblock \showarticletitle{Generative adversarial networks}.
\newblock \bibinfo{journal}{\emph{Commun. ACM}} \bibinfo{volume}{63},
  \bibinfo{number}{11} (\bibinfo{year}{2020}), \bibinfo{pages}{139--144}.
\newblock


\bibitem[Goodfellow et~al\mbox{.}(2014)]%
        {goodfellow2014explaining}
\bibfield{author}{\bibinfo{person}{Ian~J Goodfellow}, \bibinfo{person}{Jonathon
  Shlens}, {and} \bibinfo{person}{Christian Szegedy}.}
  \bibinfo{year}{2014}\natexlab{}.
\newblock \showarticletitle{Explaining and harnessing adversarial examples}.
\newblock \bibinfo{journal}{\emph{arXiv preprint arXiv:1412.6572}}
  (\bibinfo{year}{2014}).
\newblock


\bibitem[Grira et~al\mbox{.}(2004)]%
        {grira2004unsupervised}
\bibfield{author}{\bibinfo{person}{Nizar Grira}, \bibinfo{person}{Michel
  Crucianu}, {and} \bibinfo{person}{Nozha Boujemaa}.}
  \bibinfo{year}{2004}\natexlab{}.
\newblock \showarticletitle{Unsupervised and semi-supervised clustering: a
  brief survey}.
\newblock \bibinfo{journal}{\emph{A review of machine learning techniques for
  processing multimedia content}} \bibinfo{volume}{1}, \bibinfo{number}{2004}
  (\bibinfo{year}{2004}), \bibinfo{pages}{9--16}.
\newblock


\bibitem[Gu and Rigazio(2014)]%
        {gu2014towards}
\bibfield{author}{\bibinfo{person}{Shixiang Gu} {and} \bibinfo{person}{Luca
  Rigazio}.} \bibinfo{year}{2014}\natexlab{}.
\newblock \showarticletitle{Towards deep neural network architectures robust to
  adversarial examples}.
\newblock \bibinfo{journal}{\emph{arXiv preprint arXiv:1412.5068}}
  (\bibinfo{year}{2014}).
\newblock


\bibitem[Guan et~al\mbox{.}(2018)]%
        {guan2018machine}
\bibfield{author}{\bibinfo{person}{Zhenyu Guan}, \bibinfo{person}{Liangxu
  Bian}, \bibinfo{person}{Tao Shang}, {and} \bibinfo{person}{Jianwei Liu}.}
  \bibinfo{year}{2018}\natexlab{}.
\newblock \showarticletitle{When machine learning meets security issues: A
  survey}. In \bibinfo{booktitle}{\emph{2018 IEEE international conference on
  intelligence and safety for robotics (ISR)}}. IEEE,
  \bibinfo{pages}{158--165}.
\newblock


\bibitem[Gupta et~al\mbox{.}(2020)]%
        {FW2_11}
\bibfield{author}{\bibinfo{person}{Saurabh Gupta}, \bibinfo{person}{Arun~Balaji
  Buduru}, {and} \bibinfo{person}{Ponnurangam Kumaraguru}.}
  \bibinfo{year}{2020}\natexlab{}.
\newblock \showarticletitle{imdpgan: Generating private and specific data with
  generative adversarial networks}. In \bibinfo{booktitle}{\emph{2020 Second
  IEEE International Conference on Trust, Privacy and Security in Intelligent
  Systems and Applications (TPS-ISA)}}. IEEE, \bibinfo{pages}{64--72}.
\newblock


\bibitem[Hahne et~al\mbox{.}(2008)]%
        {hahne2008unsupervised}
\bibfield{author}{\bibinfo{person}{Florian Hahne}, \bibinfo{person}{Wolfgang
  Huber}, \bibinfo{person}{Robert Gentleman}, \bibinfo{person}{Seth Falcon},
  \bibinfo{person}{R Gentleman}, {and} \bibinfo{person}{VJ Carey}.}
  \bibinfo{year}{2008}\natexlab{}.
\newblock \showarticletitle{Unsupervised machine learning}.
\newblock \bibinfo{journal}{\emph{Bioconductor case studies}}
  (\bibinfo{year}{2008}), \bibinfo{pages}{137--157}.
\newblock


\bibitem[Han et~al\mbox{.}(2018)]%
        {han2018gan}
\bibfield{author}{\bibinfo{person}{Changhee Han}, \bibinfo{person}{Hideaki
  Hayashi}, \bibinfo{person}{Leonardo Rundo}, \bibinfo{person}{Ryosuke Araki},
  \bibinfo{person}{Wataru Shimoda}, \bibinfo{person}{Shinichi Muramatsu},
  \bibinfo{person}{Yujiro Furukawa}, \bibinfo{person}{Giancarlo Mauri}, {and}
  \bibinfo{person}{Hideki Nakayama}.} \bibinfo{year}{2018}\natexlab{}.
\newblock \showarticletitle{GAN-based synthetic brain MR image generation}. In
  \bibinfo{booktitle}{\emph{2018 IEEE 15th international symposium on
  biomedical imaging (ISBI 2018)}}. IEEE, \bibinfo{pages}{734--738}.
\newblock


\bibitem[Hassanzadeh and Tillman(2022)]%
        {FW1_16}
\bibfield{author}{\bibinfo{person}{Parisa Hassanzadeh} {and}
  \bibinfo{person}{Robert~E Tillman}.} \bibinfo{year}{2022}\natexlab{}.
\newblock \showarticletitle{Generative Models with Information-Theoretic
  Protection Against Membership Inference Attacks}.
\newblock \bibinfo{journal}{\emph{arXiv preprint arXiv:2206.00071}}
  (\bibinfo{year}{2022}).
\newblock


\bibitem[Hayes et~al\mbox{.}(2017)]%
        {BW1_9}
\bibfield{author}{\bibinfo{person}{Jamie Hayes}, \bibinfo{person}{Luca Melis},
  \bibinfo{person}{George Danezis}, {and} \bibinfo{person}{Emiliano
  De~Cristofaro}.} \bibinfo{year}{2017}\natexlab{}.
\newblock \showarticletitle{Logan: Membership inference attacks against
  generative models}.
\newblock \bibinfo{journal}{\emph{arXiv preprint arXiv:1705.07663}}
  (\bibinfo{year}{2017}).
\newblock


\bibitem[Hilprecht et~al\mbox{.}(2019)]%
        {BW1_10}
\bibfield{author}{\bibinfo{person}{Benjamin Hilprecht}, \bibinfo{person}{Martin
  H{\"a}rterich}, {and} \bibinfo{person}{Daniel Bernau}.}
  \bibinfo{year}{2019}\natexlab{}.
\newblock \showarticletitle{Monte Carlo and Reconstruction Membership Inference
  Attacks against Generative Models.}
\newblock \bibinfo{journal}{\emph{Proc. Priv. Enhancing Technol.}}
  \bibinfo{volume}{2019}, \bibinfo{number}{4} (\bibinfo{year}{2019}),
  \bibinfo{pages}{232--249}.
\newblock


\bibitem[Hosseini et~al\mbox{.}(2017)]%
        {hosseini2017blocking}
\bibfield{author}{\bibinfo{person}{Hossein Hosseini}, \bibinfo{person}{Yize
  Chen}, \bibinfo{person}{Sreeram Kannan}, \bibinfo{person}{Baosen Zhang},
  {and} \bibinfo{person}{Radha Poovendran}.} \bibinfo{year}{2017}\natexlab{}.
\newblock \showarticletitle{Blocking transferability of adversarial examples in
  black-box learning systems}.
\newblock \bibinfo{journal}{\emph{arXiv preprint arXiv:1703.04318}}
  (\bibinfo{year}{2017}).
\newblock


\bibitem[Hu et~al\mbox{.}(2021)]%
        {FW1_20}
\bibfield{author}{\bibinfo{person}{Aoting Hu}, \bibinfo{person}{Renjie Xie},
  \bibinfo{person}{Zhigang Lu}, \bibinfo{person}{Aiqun Hu}, {and}
  \bibinfo{person}{Minhui Xue}.} \bibinfo{year}{2021}\natexlab{}.
\newblock \showarticletitle{TableGAN-MCA: Evaluating Membership Collisions of
  GAN-Synthesized Tabular Data Releasing}. In
  \bibinfo{booktitle}{\emph{Proceedings of the 2021 ACM SIGSAC Conference on
  Computer and Communications Security}}. \bibinfo{pages}{2096--2112}.
\newblock


\bibitem[Hu and Pang(2021)]%
        {FW1_27}
\bibfield{author}{\bibinfo{person}{Hailong Hu} {and} \bibinfo{person}{Jun
  Pang}.} \bibinfo{year}{2021}\natexlab{}.
\newblock \showarticletitle{Model Extraction and Defenses on Generative
  Adversarial Networks}.
\newblock \bibinfo{journal}{\emph{arXiv preprint arXiv:2101.02069}}
  (\bibinfo{year}{2021}).
\newblock


\bibitem[Ivarsson and Gorschek(2011)]%
        {ivarsson2011method}
\bibfield{author}{\bibinfo{person}{Martin Ivarsson} {and} \bibinfo{person}{Tony
  Gorschek}.} \bibinfo{year}{2011}\natexlab{}.
\newblock \showarticletitle{A method for evaluating rigor and industrial
  relevance of technology evaluations}.
\newblock \bibinfo{journal}{\emph{Empirical Software Engineering}}
  \bibinfo{volume}{16}, \bibinfo{number}{3} (\bibinfo{year}{2011}),
  \bibinfo{pages}{365--395}.
\newblock


\bibitem[Jia and Gong(2018)]%
        {jia2018attriguard}
\bibfield{author}{\bibinfo{person}{Jinyuan Jia} {and}
  \bibinfo{person}{Neil~Zhenqiang Gong}.} \bibinfo{year}{2018}\natexlab{}.
\newblock \showarticletitle{Attriguard: A practical defense against attribute
  inference attacks via adversarial machine learning}. In
  \bibinfo{booktitle}{\emph{27th $\{$USENIX$\}$ security symposium
  ($\{$USENIX$\}$ security 18)}}. \bibinfo{pages}{513--529}.
\newblock


\bibitem[Jia et~al\mbox{.}(2019)]%
        {jia2019memguard}
\bibfield{author}{\bibinfo{person}{Jinyuan Jia}, \bibinfo{person}{Ahmed Salem},
  \bibinfo{person}{Michael Backes}, \bibinfo{person}{Yang Zhang}, {and}
  \bibinfo{person}{Neil~Zhenqiang Gong}.} \bibinfo{year}{2019}\natexlab{}.
\newblock \showarticletitle{Memguard: Defending against black-box membership
  inference attacks via adversarial examples}. In
  \bibinfo{booktitle}{\emph{Proceedings of the 2019 ACM SIGSAC conference on
  computer and communications security}}. \bibinfo{pages}{259--274}.
\newblock


\bibitem[Jin and Li(2022)]%
        {FW2_13}
\bibfield{author}{\bibinfo{person}{Ruinan Jin} {and} \bibinfo{person}{Xiaoxiao
  Li}.} \bibinfo{year}{2022}\natexlab{}.
\newblock \showarticletitle{Backdoor Attack is a Devil in Federated GAN-Based
  Medical Image Synthesis}. In \bibinfo{booktitle}{\emph{International Workshop
  on Simulation and Synthesis in Medical Imaging}}. Springer,
  \bibinfo{pages}{154--165}.
\newblock


\bibitem[Jordon et~al\mbox{.}(2018)]%
        {BW1_11}
\bibfield{author}{\bibinfo{person}{James Jordon}, \bibinfo{person}{Jinsung
  Yoon}, {and} \bibinfo{person}{Mihaela Van Der~Schaar}.}
  \bibinfo{year}{2018}\natexlab{}.
\newblock \showarticletitle{PATE-GAN: Generating synthetic data with
  differential privacy guarantees}. In \bibinfo{booktitle}{\emph{International
  conference on learning representations}}.
\newblock


\bibitem[Juuti et~al\mbox{.}(2019)]%
        {juuti2019prada}
\bibfield{author}{\bibinfo{person}{Mika Juuti}, \bibinfo{person}{Sebastian
  Szyller}, \bibinfo{person}{Samuel Marchal}, {and} \bibinfo{person}{N
  Asokan}.} \bibinfo{year}{2019}\natexlab{}.
\newblock \showarticletitle{PRADA: protecting against DNN model stealing
  attacks}. In \bibinfo{booktitle}{\emph{2019 IEEE European Symposium on
  Security and Privacy (EuroS\&P)}}. IEEE, \bibinfo{pages}{512--527}.
\newblock


\bibitem[Keele et~al\mbox{.}(2007)]%
        {keele2007guidelines}
\bibfield{author}{\bibinfo{person}{Staffs Keele} {et~al\mbox{.}}}
  \bibinfo{year}{2007}\natexlab{}.
\newblock \bibinfo{booktitle}{\emph{Guidelines for performing systematic
  literature reviews in software engineering}}.
\newblock \bibinfo{type}{{T}echnical {R}eport}. \bibinfo{institution}{Technical
  report, ver. 2.3 ebse technical report. ebse}.
\newblock


\bibitem[Kim et~al\mbox{.}(2020)]%
        {17}
\bibfield{author}{\bibinfo{person}{Byeong~Cheon Kim}, \bibinfo{person}{Jung~Uk
  Kim}, \bibinfo{person}{Hakmin Lee}, {and} \bibinfo{person}{Yong~Man Ro}.}
  \bibinfo{year}{2020}\natexlab{}.
\newblock \showarticletitle{Revisiting role of autoencoders in adversarial
  settings}. In \bibinfo{booktitle}{\emph{2020 IEEE International Conference on
  Image Processing (ICIP)}}. IEEE, \bibinfo{pages}{1856--1860}.
\newblock


\bibitem[Kingma et~al\mbox{.}(2019)]%
        {kingma2019introduction}
\bibfield{author}{\bibinfo{person}{Diederik~P Kingma}, \bibinfo{person}{Max
  Welling}, {et~al\mbox{.}}} \bibinfo{year}{2019}\natexlab{}.
\newblock \showarticletitle{An introduction to variational autoencoders}.
\newblock \bibinfo{journal}{\emph{Foundations and Trends{\textregistered} in
  Machine Learning}} \bibinfo{volume}{12}, \bibinfo{number}{4}
  (\bibinfo{year}{2019}), \bibinfo{pages}{307--392}.
\newblock


\bibitem[Kitchenham et~al\mbox{.}(2015)]%
        {kitchenham2015evidence}
\bibfield{author}{\bibinfo{person}{Barbara~Ann Kitchenham},
  \bibinfo{person}{David Budgen}, {and} \bibinfo{person}{Pearl Brereton}.}
  \bibinfo{year}{2015}\natexlab{}.
\newblock \bibinfo{booktitle}{\emph{Evidence-based software engineering and
  systematic reviews}}. Vol.~\bibinfo{volume}{4}.
\newblock \bibinfo{publisher}{CRC press}.
\newblock


\bibitem[Kos et~al\mbox{.}(2018)]%
        {FW2_14}
\bibfield{author}{\bibinfo{person}{Jernej Kos}, \bibinfo{person}{Ian Fischer},
  {and} \bibinfo{person}{Dawn Song}.} \bibinfo{year}{2018}\natexlab{}.
\newblock \showarticletitle{Adversarial examples for generative models}. In
  \bibinfo{booktitle}{\emph{2018 ieee security and privacy workshops (spw)}}.
  IEEE, \bibinfo{pages}{36--42}.
\newblock


\bibitem[Kossen et~al\mbox{.}(2022)]%
        {FW1_38}
\bibfield{author}{\bibinfo{person}{Tabea Kossen}, \bibinfo{person}{Manuel~A
  Hirzel}, \bibinfo{person}{Vince~I Madai}, \bibinfo{person}{Franziska
  Boenisch}, \bibinfo{person}{Anja Hennemuth}, \bibinfo{person}{Kristian
  Hildebrand}, \bibinfo{person}{Sebastian Pokutta}, \bibinfo{person}{Kartikey
  Sharma}, \bibinfo{person}{Adam Hilbert}, \bibinfo{person}{Jan Sobesky},
  {et~al\mbox{.}}} \bibinfo{year}{2022}\natexlab{}.
\newblock \showarticletitle{Toward Sharing Brain Images: Differentially Private
  TOF-MRA Images With Segmentation Labels Using Generative Adversarial
  Networks}.
\newblock \bibinfo{journal}{\emph{Frontiers in artificial intelligence}}
  \bibinfo{volume}{5} (\bibinfo{year}{2022}).
\newblock


\bibitem[Kunar et~al\mbox{.}(2021)]%
        {FW1_42}
\bibfield{author}{\bibinfo{person}{Aditya Kunar}, \bibinfo{person}{Robert
  Birke}, \bibinfo{person}{Zilong Zhao}, {and} \bibinfo{person}{Lydia Chen}.}
  \bibinfo{year}{2021}\natexlab{}.
\newblock \showarticletitle{DTGAN: Differential Private Training for Tabular
  GANs}.
\newblock \bibinfo{journal}{\emph{arXiv preprint arXiv:2107.02521}}
  (\bibinfo{year}{2021}).
\newblock


\bibitem[Kusner et~al\mbox{.}(2015)]%
        {kusner2015differentially}
\bibfield{author}{\bibinfo{person}{Matt Kusner}, \bibinfo{person}{Jacob
  Gardner}, \bibinfo{person}{Roman Garnett}, {and} \bibinfo{person}{Kilian
  Weinberger}.} \bibinfo{year}{2015}\natexlab{}.
\newblock \showarticletitle{Differentially private Bayesian optimization}. In
  \bibinfo{booktitle}{\emph{International conference on machine learning}}.
  PMLR, \bibinfo{pages}{918--927}.
\newblock


\bibitem[Kuzina et~al\mbox{.}(2021)]%
        {FW2_15}
\bibfield{author}{\bibinfo{person}{Anna Kuzina}, \bibinfo{person}{Max Welling},
  {and} \bibinfo{person}{Jakub~M Tomczak}.} \bibinfo{year}{2021}\natexlab{}.
\newblock \showarticletitle{Diagnosing vulnerability of variational
  auto-encoders to adversarial attacks}.
\newblock \bibinfo{journal}{\emph{arXiv preprint arXiv:2103.06701}}
  (\bibinfo{year}{2021}).
\newblock


\bibitem[Kuzina et~al\mbox{.}(2022)]%
        {FW2_16}
\bibfield{author}{\bibinfo{person}{Anna Kuzina}, \bibinfo{person}{Max Welling},
  {and} \bibinfo{person}{Jakub~M Tomczak}.} \bibinfo{year}{2022}\natexlab{}.
\newblock \showarticletitle{Defending Variational Autoencoders from Adversarial
  Attacks with MCMC}.
\newblock \bibinfo{journal}{\emph{arXiv preprint arXiv:2203.09940}}
  (\bibinfo{year}{2022}).
\newblock


\bibitem[Lin et~al\mbox{.}(2020)]%
        {03}
\bibfield{author}{\bibinfo{person}{Zinan Lin}, \bibinfo{person}{Alankar Jain},
  \bibinfo{person}{Chen Wang}, \bibinfo{person}{Giulia Fanti}, {and}
  \bibinfo{person}{Vyas Sekar}.} \bibinfo{year}{2020}\natexlab{}.
\newblock \showarticletitle{Using GANs for sharing networked time series data:
  Challenges, initial promise, and open questions}. In
  \bibinfo{booktitle}{\emph{Proceedings of the ACM Internet Measurement
  Conference}}. \bibinfo{pages}{464--483}.
\newblock


\bibitem[Lin et~al\mbox{.}(2021)]%
        {FW1_43}
\bibfield{author}{\bibinfo{person}{Zinan Lin}, \bibinfo{person}{Vyas Sekar},
  {and} \bibinfo{person}{Giulia Fanti}.} \bibinfo{year}{2021}\natexlab{}.
\newblock \showarticletitle{On the privacy properties of gan-generated
  samples}. In \bibinfo{booktitle}{\emph{International Conference on Artificial
  Intelligence and Statistics}}. PMLR, \bibinfo{pages}{1522--1530}.
\newblock


\bibitem[Liu et~al\mbox{.}(2017b)]%
        {liu2017robust}
\bibfield{author}{\bibinfo{person}{Chang Liu}, \bibinfo{person}{Bo Li},
  \bibinfo{person}{Yevgeniy Vorobeychik}, {and} \bibinfo{person}{Alina Oprea}.}
  \bibinfo{year}{2017}\natexlab{b}.
\newblock \showarticletitle{Robust linear regression against training data
  poisoning}. In \bibinfo{booktitle}{\emph{Proceedings of the 10th ACM workshop
  on artificial intelligence and security}}. \bibinfo{pages}{91--102}.
\newblock


\bibitem[Liu et~al\mbox{.}(2017a)]%
        {liu2017oblivious}
\bibfield{author}{\bibinfo{person}{Jian Liu}, \bibinfo{person}{Mika Juuti},
  \bibinfo{person}{Yao Lu}, {and} \bibinfo{person}{Nadarajah Asokan}.}
  \bibinfo{year}{2017}\natexlab{a}.
\newblock \showarticletitle{Oblivious neural network predictions via minionn
  transformations}. In \bibinfo{booktitle}{\emph{Proceedings of the 2017 ACM
  SIGSAC conference on computer and communications security}}.
  \bibinfo{pages}{619--631}.
\newblock


\bibitem[Liu et~al\mbox{.}(2018)]%
        {liu2018survey}
\bibfield{author}{\bibinfo{person}{Qiang Liu}, \bibinfo{person}{Pan Li},
  \bibinfo{person}{Wentao Zhao}, \bibinfo{person}{Wei Cai},
  \bibinfo{person}{Shui Yu}, {and} \bibinfo{person}{Victor~CM Leung}.}
  \bibinfo{year}{2018}\natexlab{}.
\newblock \showarticletitle{A survey on security threats and defensive
  techniques of machine learning: A data driven view}.
\newblock \bibinfo{journal}{\emph{IEEE access}}  \bibinfo{volume}{6}
  (\bibinfo{year}{2018}), \bibinfo{pages}{12103--12117}.
\newblock


\bibitem[Liu and Hsieh(2019)]%
        {18}
\bibfield{author}{\bibinfo{person}{Xuanqing Liu} {and} \bibinfo{person}{Cho-Jui
  Hsieh}.} \bibinfo{year}{2019}\natexlab{}.
\newblock \showarticletitle{Rob-gan: Generator, discriminator, and adversarial
  attacker}. In \bibinfo{booktitle}{\emph{Proceedings of the IEEE/CVF
  Conference on Computer Vision and Pattern Recognition}}.
  \bibinfo{pages}{11234--11243}.
\newblock


\bibitem[Liu et~al\mbox{.}(2019b)]%
        {19}
\bibfield{author}{\bibinfo{person}{Xuanqing Liu}, \bibinfo{person}{Si Si},
  \bibinfo{person}{Xiaojin Zhu}, \bibinfo{person}{Yang Li}, {and}
  \bibinfo{person}{Cho-Jui Hsieh}.} \bibinfo{year}{2019}\natexlab{b}.
\newblock \showarticletitle{A unified framework for data poisoning attack to
  graph-based semi-supervised learning}.
\newblock \bibinfo{journal}{\emph{arXiv preprint arXiv:1910.14147}}
  (\bibinfo{year}{2019}).
\newblock


\bibitem[Liu et~al\mbox{.}(2019c)]%
        {liu2019reinforcement}
\bibfield{author}{\bibinfo{person}{Xing Liu}, \bibinfo{person}{Hansong Xu},
  \bibinfo{person}{Weixian Liao}, {and} \bibinfo{person}{Wei Yu}.}
  \bibinfo{year}{2019}\natexlab{c}.
\newblock \showarticletitle{Reinforcement learning for cyber-physical systems}.
  In \bibinfo{booktitle}{\emph{2019 IEEE International Conference on Industrial
  Internet (ICII)}}. IEEE, \bibinfo{pages}{318--327}.
\newblock


\bibitem[Liu et~al\mbox{.}(2019a)]%
        {FW2_19}
\bibfield{author}{\bibinfo{person}{Yi Liu}, \bibinfo{person}{Jialiang Peng},
  \bibinfo{person}{JQ James}, {and} \bibinfo{person}{Yi Wu}.}
  \bibinfo{year}{2019}\natexlab{a}.
\newblock \showarticletitle{PPGAN: Privacy-preserving generative adversarial
  network}. In \bibinfo{booktitle}{\emph{2019 IEEE 25Th international
  conference on parallel and distributed systems (ICPADS)}}. IEEE,
  \bibinfo{pages}{985--989}.
\newblock


\bibitem[Long et~al\mbox{.}(2021)]%
        {BW2_7}
\bibfield{author}{\bibinfo{person}{Yunhui Long}, \bibinfo{person}{Boxin Wang},
  \bibinfo{person}{Zhuolin Yang}, \bibinfo{person}{Bhavya Kailkhura},
  \bibinfo{person}{Aston Zhang}, \bibinfo{person}{Carl Gunter}, {and}
  \bibinfo{person}{Bo Li}.} \bibinfo{year}{2021}\natexlab{}.
\newblock \showarticletitle{G-PATE: Scalable Differentially Private Data
  Generator via Private Aggregation of Teacher Discriminators}.
\newblock \bibinfo{journal}{\emph{Advances in Neural Information Processing
  Systems}}  \bibinfo{volume}{34} (\bibinfo{year}{2021}),
  \bibinfo{pages}{2965--2977}.
\newblock


\bibitem[Martins et~al\mbox{.}(2020)]%
        {martins2020adversarial}
\bibfield{author}{\bibinfo{person}{Nuno Martins},
  \bibinfo{person}{Jos{\'e}~Magalh{\~a}es Cruz}, \bibinfo{person}{Tiago Cruz},
  {and} \bibinfo{person}{Pedro~Henriques Abreu}.}
  \bibinfo{year}{2020}\natexlab{}.
\newblock \showarticletitle{Adversarial machine learning applied to intrusion
  and malware scenarios: a systematic review}.
\newblock \bibinfo{journal}{\emph{IEEE Access}}  \bibinfo{volume}{8}
  (\bibinfo{year}{2020}), \bibinfo{pages}{35403--35419}.
\newblock


\bibitem[McMahan et~al\mbox{.}(2017)]%
        {mcmahan2017communication}
\bibfield{author}{\bibinfo{person}{Brendan McMahan}, \bibinfo{person}{Eider
  Moore}, \bibinfo{person}{Daniel Ramage}, \bibinfo{person}{Seth Hampson},
  {and} \bibinfo{person}{Blaise~Aguera y Arcas}.}
  \bibinfo{year}{2017}\natexlab{}.
\newblock \showarticletitle{Communication-efficient learning of deep networks
  from decentralized data}. In \bibinfo{booktitle}{\emph{Artificial
  intelligence and statistics}}. PMLR, \bibinfo{pages}{1273--1282}.
\newblock


\bibitem[Meng and Chen(2017)]%
        {meng2017magnet}
\bibfield{author}{\bibinfo{person}{Dongyu Meng} {and} \bibinfo{person}{Hao
  Chen}.} \bibinfo{year}{2017}\natexlab{}.
\newblock \showarticletitle{Magnet: a two-pronged defense against adversarial
  examples}. In \bibinfo{booktitle}{\emph{Proceedings of the 2017 ACM SIGSAC
  conference on computer and communications security}}.
  \bibinfo{pages}{135--147}.
\newblock


\bibitem[Mukherjee et~al\mbox{.}(2021)]%
        {FW1_2}
\bibfield{author}{\bibinfo{person}{Sumit Mukherjee}, \bibinfo{person}{Yixi Xu},
  \bibinfo{person}{Anusua Trivedi}, \bibinfo{person}{Nabajyoti Patowary}, {and}
  \bibinfo{person}{Juan~Lavista Ferres}.} \bibinfo{year}{2021}\natexlab{}.
\newblock \showarticletitle{privGAN: Protecting GANs from membership inference
  attacks at low cost to utility.}
\newblock \bibinfo{journal}{\emph{Proc. Priv. Enhancing Technol.}}
  \bibinfo{volume}{2021}, \bibinfo{number}{3} (\bibinfo{year}{2021}),
  \bibinfo{pages}{142--163}.
\newblock


\bibitem[Nasr et~al\mbox{.}(2018)]%
        {nasr2018machine}
\bibfield{author}{\bibinfo{person}{Milad Nasr}, \bibinfo{person}{Reza Shokri},
  {and} \bibinfo{person}{Amir Houmansadr}.} \bibinfo{year}{2018}\natexlab{}.
\newblock \showarticletitle{Machine learning with membership privacy using
  adversarial regularization}. In \bibinfo{booktitle}{\emph{Proceedings of the
  2018 ACM SIGSAC conference on computer and communications security}}.
  \bibinfo{pages}{634--646}.
\newblock


\bibitem[Nelson et~al\mbox{.}(2009)]%
        {nelson2009misleading}
\bibfield{author}{\bibinfo{person}{Blaine Nelson}, \bibinfo{person}{Marco
  Barreno}, \bibinfo{person}{Fuching Jack~Chi}, \bibinfo{person}{Anthony~D
  Joseph}, \bibinfo{person}{Benjamin~IP Rubinstein}, \bibinfo{person}{Udam
  Saini}, \bibinfo{person}{Charles Sutton}, \bibinfo{person}{JD Tygar}, {and}
  \bibinfo{person}{Kai Xia}.} \bibinfo{year}{2009}\natexlab{}.
\newblock \showarticletitle{Misleading learners: Co-opting your spam filter}.
\newblock \bibinfo{journal}{\emph{Machine learning in cyber trust: Security,
  privacy, and reliability}} (\bibinfo{year}{2009}), \bibinfo{pages}{17--51}.
\newblock


\bibitem[Nkashama et~al\mbox{.}(2022)]%
        {FW2_24}
\bibfield{author}{\bibinfo{person}{D Nkashama}, \bibinfo{person}{Arian
  Soltani}, \bibinfo{person}{Jean-Charles Verdier}, \bibinfo{person}{Marc
  Frappier}, \bibinfo{person}{Pierre-Marting Tardif}, {and}
  \bibinfo{person}{Froduald Kabanza}.} \bibinfo{year}{2022}\natexlab{}.
\newblock \showarticletitle{Robustness Evaluation of Deep Unsupervised Learning
  Algorithms for Intrusion Detection Systems}.
\newblock \bibinfo{journal}{\emph{arXiv preprint arXiv:2207.03576}}
  (\bibinfo{year}{2022}).
\newblock


\bibitem[Oleszkiewicz et~al\mbox{.}(2018)]%
        {FW1_40}
\bibfield{author}{\bibinfo{person}{Witold Oleszkiewicz}, \bibinfo{person}{Peter
  Kairouz}, \bibinfo{person}{Karol Piczak}, \bibinfo{person}{Ram Rajagopal},
  {and} \bibinfo{person}{Tomasz Trzci{\'n}ski}.}
  \bibinfo{year}{2018}\natexlab{}.
\newblock \showarticletitle{Siamese generative adversarial privatizer for
  biometric data}. In \bibinfo{booktitle}{\emph{Asian Conference on Computer
  Vision}}. Springer, \bibinfo{pages}{482--497}.
\newblock


\bibitem[Orekondy et~al\mbox{.}(2020)]%
        {orekondy2020prediction}
\bibfield{author}{\bibinfo{person}{Tribhuvanesh Orekondy},
  \bibinfo{person}{Bernt Schiele}, {and} \bibinfo{person}{Mario Fritz}.}
  \bibinfo{year}{2020}\natexlab{}.
\newblock \showarticletitle{Prediction poisoning: Utility-constrained defenses
  against model stealing attacks}. In \bibinfo{booktitle}{\emph{International
  Conference on Representation Learning (ICLR) 2020}}.
\newblock


\bibitem[Osia et~al\mbox{.}(2020)]%
        {osia2020hybrid}
\bibfield{author}{\bibinfo{person}{Seyed~Ali Osia}, \bibinfo{person}{Ali~Shahin
  Shamsabadi}, \bibinfo{person}{Sina Sajadmanesh}, \bibinfo{person}{Ali
  Taheri}, \bibinfo{person}{Kleomenis Katevas}, \bibinfo{person}{Hamid~R
  Rabiee}, \bibinfo{person}{Nicholas~D Lane}, {and} \bibinfo{person}{Hamed
  Haddadi}.} \bibinfo{year}{2020}\natexlab{}.
\newblock \showarticletitle{A hybrid deep learning architecture for
  privacy-preserving mobile analytics}.
\newblock \bibinfo{journal}{\emph{IEEE Internet of Things Journal}}
  \bibinfo{volume}{7}, \bibinfo{number}{5} (\bibinfo{year}{2020}),
  \bibinfo{pages}{4505--4518}.
\newblock


\bibitem[Papernot et~al\mbox{.}(2016a)]%
        {papernot2016semi}
\bibfield{author}{\bibinfo{person}{Nicolas Papernot},
  \bibinfo{person}{Mart{\'\i}n Abadi}, \bibinfo{person}{Ulfar Erlingsson},
  \bibinfo{person}{Ian Goodfellow}, {and} \bibinfo{person}{Kunal Talwar}.}
  \bibinfo{year}{2016}\natexlab{a}.
\newblock \showarticletitle{Semi-supervised knowledge transfer for deep
  learning from private training data}.
\newblock \bibinfo{journal}{\emph{arXiv preprint arXiv:1610.05755}}
  (\bibinfo{year}{2016}).
\newblock


\bibitem[Papernot et~al\mbox{.}(2017)]%
        {papernot2017practical}
\bibfield{author}{\bibinfo{person}{Nicolas Papernot}, \bibinfo{person}{Patrick
  McDaniel}, \bibinfo{person}{Ian Goodfellow}, \bibinfo{person}{Somesh Jha},
  \bibinfo{person}{Z~Berkay Celik}, {and} \bibinfo{person}{Ananthram Swami}.}
  \bibinfo{year}{2017}\natexlab{}.
\newblock \showarticletitle{Practical black-box attacks against machine
  learning}. In \bibinfo{booktitle}{\emph{Proceedings of the 2017 ACM on Asia
  conference on computer and communications security}}.
  \bibinfo{pages}{506--519}.
\newblock


\bibitem[Papernot et~al\mbox{.}(2016b)]%
        {papernot2016towards}
\bibfield{author}{\bibinfo{person}{Nicolas Papernot}, \bibinfo{person}{Patrick
  McDaniel}, \bibinfo{person}{Arunesh Sinha}, {and} \bibinfo{person}{Michael
  Wellman}.} \bibinfo{year}{2016}\natexlab{b}.
\newblock \showarticletitle{Towards the science of security and privacy in
  machine learning}.
\newblock \bibinfo{journal}{\emph{arXiv preprint arXiv:1611.03814}}
  (\bibinfo{year}{2016}).
\newblock


\bibitem[Papernot et~al\mbox{.}(2016c)]%
        {papernot2016distillation}
\bibfield{author}{\bibinfo{person}{Nicolas Papernot}, \bibinfo{person}{Patrick
  McDaniel}, \bibinfo{person}{Xi Wu}, \bibinfo{person}{Somesh Jha}, {and}
  \bibinfo{person}{Ananthram Swami}.} \bibinfo{year}{2016}\natexlab{c}.
\newblock \showarticletitle{Distillation as a defense to adversarial
  perturbations against deep neural networks}. In
  \bibinfo{booktitle}{\emph{2016 IEEE symposium on security and privacy (SP)}}.
  IEEE, \bibinfo{pages}{582--597}.
\newblock


\bibitem[Pe{\~n}a et~al\mbox{.}(2014)]%
        {pena2014object}
\bibfield{author}{\bibinfo{person}{Jos{\'e}~M Pe{\~n}a},
  \bibinfo{person}{Pedro~A Guti{\'e}rrez}, \bibinfo{person}{C{\'e}sar
  Herv{\'a}s-Mart{\'\i}nez}, \bibinfo{person}{Johan Six},
  \bibinfo{person}{Richard~E Plant}, {and} \bibinfo{person}{Francisca
  L{\'o}pez-Granados}.} \bibinfo{year}{2014}\natexlab{}.
\newblock \showarticletitle{Object-based image classification of summer crops
  with machine learning methods}.
\newblock \bibinfo{journal}{\emph{Remote sensing}} \bibinfo{volume}{6},
  \bibinfo{number}{6} (\bibinfo{year}{2014}), \bibinfo{pages}{5019--5041}.
\newblock


\bibitem[Petticrew and Roberts(2008)]%
        {petticrew2008systematic}
\bibfield{author}{\bibinfo{person}{Mark Petticrew} {and} \bibinfo{person}{Helen
  Roberts}.} \bibinfo{year}{2008}\natexlab{}.
\newblock \bibinfo{booktitle}{\emph{Systematic reviews in the social sciences:
  A practical guide}}.
\newblock \bibinfo{publisher}{John Wiley \& Sons}.
\newblock


\bibitem[Phan et~al\mbox{.}(2016)]%
        {phan2016differential}
\bibfield{author}{\bibinfo{person}{NhatHai Phan}, \bibinfo{person}{Yue Wang},
  \bibinfo{person}{Xintao Wu}, {and} \bibinfo{person}{Dejing Dou}.}
  \bibinfo{year}{2016}\natexlab{}.
\newblock \showarticletitle{Differential privacy preservation for deep
  auto-encoders: an application of human behavior prediction}. In
  \bibinfo{booktitle}{\emph{Thirtieth AAAI Conference on Artificial
  Intelligence}}.
\newblock


\bibitem[Phan et~al\mbox{.}(2017)]%
        {phan2017adaptive}
\bibfield{author}{\bibinfo{person}{NhatHai Phan}, \bibinfo{person}{Xintao Wu},
  \bibinfo{person}{Han Hu}, {and} \bibinfo{person}{Dejing Dou}.}
  \bibinfo{year}{2017}\natexlab{}.
\newblock \showarticletitle{Adaptive laplace mechanism: Differential privacy
  preservation in deep learning}. In \bibinfo{booktitle}{\emph{2017 IEEE
  international conference on data mining (ICDM)}}. IEEE,
  \bibinfo{pages}{385--394}.
\newblock


\bibitem[Pinaya et~al\mbox{.}(2020)]%
        {pinaya2020autoencoders}
\bibfield{author}{\bibinfo{person}{Walter Hugo~Lopez Pinaya},
  \bibinfo{person}{Sandra Vieira}, \bibinfo{person}{Rafael Garcia-Dias}, {and}
  \bibinfo{person}{Andrea Mechelli}.} \bibinfo{year}{2020}\natexlab{}.
\newblock \showarticletitle{Autoencoders}.
\newblock In \bibinfo{booktitle}{\emph{Machine learning}}.
  \bibinfo{publisher}{Elsevier}, \bibinfo{pages}{193--208}.
\newblock


\bibitem[Rajotte and Ng(2020)]%
        {FW1_29}
\bibfield{author}{\bibinfo{person}{Jean-Francois Rajotte} {and}
  \bibinfo{person}{Raymond~T Ng}.} \bibinfo{year}{2020}\natexlab{}.
\newblock \showarticletitle{Private data sharing between decentralized users
  through the privGAN architecture}. In \bibinfo{booktitle}{\emph{2020 IEEE
  24th International Enterprise Distributed Object Computing Workshop
  (EDOCW)}}. IEEE, \bibinfo{pages}{37--42}.
\newblock


\bibitem[Rani et~al\mbox{.}(2023)]%
        {rani2023deep}
\bibfield{author}{\bibinfo{person}{Geeta Rani}, \bibinfo{person}{Upasana
  Pandey}, \bibinfo{person}{Aniket~Anil Wagde}, {and}
  \bibinfo{person}{Vijaypal~Singh Dhaka}.} \bibinfo{year}{2023}\natexlab{}.
\newblock \showarticletitle{A deep reinforcement learning technique for bug
  detection in video games}.
\newblock \bibinfo{journal}{\emph{International Journal of Information
  Technology}} \bibinfo{volume}{15}, \bibinfo{number}{1}
  (\bibinfo{year}{2023}), \bibinfo{pages}{355--367}.
\newblock


\bibitem[Rawat et~al\mbox{.}(2021)]%
        {FW1_13}
\bibfield{author}{\bibinfo{person}{Ambrish Rawat}, \bibinfo{person}{Killian
  Levacher}, {and} \bibinfo{person}{Mathieu Sinn}.}
  \bibinfo{year}{2021}\natexlab{}.
\newblock \showarticletitle{The Devil is in the GAN: Defending Deep Generative
  Models Against Backdoor Attacks}.
\newblock \bibinfo{journal}{\emph{arXiv preprint arXiv:2108.01644}}
  (\bibinfo{year}{2021}).
\newblock


\bibitem[Rivest et~al\mbox{.}(1978)]%
        {rivest1978data}
\bibfield{author}{\bibinfo{person}{Ronald~L Rivest}, \bibinfo{person}{Len
  Adleman}, \bibinfo{person}{Michael~L Dertouzos}, {et~al\mbox{.}}}
  \bibinfo{year}{1978}\natexlab{}.
\newblock \showarticletitle{On data banks and privacy homomorphisms}.
\newblock \bibinfo{journal}{\emph{Foundations of secure computation}}
  \bibinfo{volume}{4}, \bibinfo{number}{11} (\bibinfo{year}{1978}),
  \bibinfo{pages}{169--180}.
\newblock


\bibitem[Rocco(2021)]%
        {rocco2021deepfake}
\bibfield{author}{\bibinfo{person}{Fabrizio Rocco}.}
  \bibinfo{year}{2021}\natexlab{}.
\newblock \showarticletitle{Deepfake generation: evaluating state of the art
  generative networks}.
\newblock  (\bibinfo{year}{2021}).
\newblock


\bibitem[R{\"o}sner and Schmidt(2018)]%
        {BW2_9}
\bibfield{author}{\bibinfo{person}{Clemens R{\"o}sner} {and}
  \bibinfo{person}{Melanie Schmidt}.} \bibinfo{year}{2018}\natexlab{}.
\newblock \showarticletitle{Privacy preserving clustering with constraints}.
\newblock \bibinfo{journal}{\emph{arXiv preprint arXiv:1802.02497}}
  (\bibinfo{year}{2018}).
\newblock


\bibitem[Sabottke et~al\mbox{.}(2019)]%
        {FW1_9}
\bibfield{author}{\bibinfo{person}{Carl Sabottke}, \bibinfo{person}{Daniel
  Chen}, \bibinfo{person}{Lucas Layman}, {and} \bibinfo{person}{Tudor
  Dumitra{\c{s}}}.} \bibinfo{year}{2019}\natexlab{}.
\newblock \showarticletitle{How to trick the Borg: threat models against manual
  and automated techniques for detecting network attacks}.
\newblock \bibinfo{journal}{\emph{Computers \& Security}}  \bibinfo{volume}{81}
  (\bibinfo{year}{2019}), \bibinfo{pages}{25--40}.
\newblock


\bibitem[Sadeghi and Larsson(2019)]%
        {21}
\bibfield{author}{\bibinfo{person}{Meysam Sadeghi} {and}
  \bibinfo{person}{Erik~G Larsson}.} \bibinfo{year}{2019}\natexlab{}.
\newblock \showarticletitle{Physical adversarial attacks against end-to-end
  autoencoder communication systems}.
\newblock \bibinfo{journal}{\emph{IEEE Communications Letters}}
  \bibinfo{volume}{23}, \bibinfo{number}{5} (\bibinfo{year}{2019}),
  \bibinfo{pages}{847--850}.
\newblock


\bibitem[Sajeeda and Hossain(2022)]%
        {FW1_32}
\bibfield{author}{\bibinfo{person}{Afia Sajeeda} {and}
  \bibinfo{person}{BM~Mainul Hossain}.} \bibinfo{year}{2022}\natexlab{}.
\newblock \showarticletitle{Exploring Generative Adversarial Networks and
  Adversarial Training}.
\newblock \bibinfo{journal}{\emph{International Journal of Cognitive Computing
  in Engineering}} (\bibinfo{year}{2022}).
\newblock


\bibitem[Salem et~al\mbox{.}(2020)]%
        {FW2_25}
\bibfield{author}{\bibinfo{person}{Ahmed Salem}, \bibinfo{person}{Yannick
  Sautter}, \bibinfo{person}{Michael Backes}, \bibinfo{person}{Mathias
  Humbert}, {and} \bibinfo{person}{Yang Zhang}.}
  \bibinfo{year}{2020}\natexlab{}.
\newblock \showarticletitle{Baaan: Backdoor attacks against autoencoder and
  gan-based machine learning models}.
\newblock \bibinfo{journal}{\emph{arXiv preprint arXiv:2010.03007}}
  (\bibinfo{year}{2020}).
\newblock


\bibitem[Salem et~al\mbox{.}(2018)]%
        {salem2018ml}
\bibfield{author}{\bibinfo{person}{Ahmed Salem}, \bibinfo{person}{Yang Zhang},
  \bibinfo{person}{Mathias Humbert}, \bibinfo{person}{Pascal Berrang},
  \bibinfo{person}{Mario Fritz}, {and} \bibinfo{person}{Michael Backes}.}
  \bibinfo{year}{2018}\natexlab{}.
\newblock \showarticletitle{Ml-leaks: Model and data independent membership
  inference attacks and defenses on machine learning models}.
\newblock \bibinfo{journal}{\emph{arXiv preprint arXiv:1806.01246}}
  (\bibinfo{year}{2018}).
\newblock


\bibitem[Sankaranarayanan et~al\mbox{.}(2018)]%
        {sankaranarayanan2018regularizing}
\bibfield{author}{\bibinfo{person}{Swami Sankaranarayanan},
  \bibinfo{person}{Arpit Jain}, \bibinfo{person}{Rama Chellappa}, {and}
  \bibinfo{person}{Ser~Nam Lim}.} \bibinfo{year}{2018}\natexlab{}.
\newblock \showarticletitle{Regularizing deep networks using efficient
  layerwise adversarial training}. In \bibinfo{booktitle}{\emph{Proceedings of
  the AAAI Conference on Artificial Intelligence}}, Vol.~\bibinfo{volume}{32}.
\newblock


\bibitem[Sengupta et~al\mbox{.}(2018)]%
        {sengupta2018mtdeep}
\bibfield{author}{\bibinfo{person}{Sailik Sengupta}, \bibinfo{person}{Tathagata
  Chakraborti}, {and} \bibinfo{person}{Subbarao Kambhampati}.}
  \bibinfo{year}{2018}\natexlab{}.
\newblock \showarticletitle{Mtdeep: boosting the security of deep neural nets
  against adversarial attacks with moving target defense}. In
  \bibinfo{booktitle}{\emph{Workshops at the thirty-second AAAI conference on
  artificial intelligence}}.
\newblock


\bibitem[Shi et~al\mbox{.}(2020)]%
        {shi2020reinforcement}
\bibfield{author}{\bibinfo{person}{Yi Shi}, \bibinfo{person}{Yalin~E Sagduyu},
  {and} \bibinfo{person}{Tugba Erpek}.} \bibinfo{year}{2020}\natexlab{}.
\newblock \showarticletitle{Reinforcement learning for dynamic resource
  optimization in 5G radio access network slicing}. In
  \bibinfo{booktitle}{\emph{2020 IEEE 25th international workshop on computer
  aided modeling and design of communication links and networks (CAMAD)}}.
  IEEE, \bibinfo{pages}{1--6}.
\newblock


\bibitem[Shokri and Shmatikov(2015)]%
        {shokri2015privacy}
\bibfield{author}{\bibinfo{person}{Reza Shokri} {and} \bibinfo{person}{Vitaly
  Shmatikov}.} \bibinfo{year}{2015}\natexlab{}.
\newblock \showarticletitle{Privacy-preserving deep learning}. In
  \bibinfo{booktitle}{\emph{Proceedings of the 22nd ACM SIGSAC conference on
  computer and communications security}}. \bibinfo{pages}{1310--1321}.
\newblock


\bibitem[Srivastava et~al\mbox{.}(2015)]%
        {srivastava2015unsupervised}
\bibfield{author}{\bibinfo{person}{Nitish Srivastava}, \bibinfo{person}{Elman
  Mansimov}, {and} \bibinfo{person}{Ruslan Salakhudinov}.}
  \bibinfo{year}{2015}\natexlab{}.
\newblock \showarticletitle{Unsupervised learning of video representations
  using lstms}. In \bibinfo{booktitle}{\emph{International conference on
  machine learning}}. PMLR, \bibinfo{pages}{843--852}.
\newblock


\bibitem[Sum~Liu et~al\mbox{.}(2018)]%
        {FW2_17}
\bibfield{author}{\bibinfo{person}{Kin Sum~Liu}, \bibinfo{person}{Chaowei
  Xiao}, \bibinfo{person}{Bo Li}, {and} \bibinfo{person}{Jie Gao}.}
  \bibinfo{year}{2018}\natexlab{}.
\newblock \showarticletitle{Performing Co-Membership Attacks Against Deep
  Generative Models}.
\newblock \bibinfo{journal}{\emph{arXiv e-prints}} (\bibinfo{year}{2018}),
  \bibinfo{pages}{arXiv--1805}.
\newblock


\bibitem[Sumedha and Weigt(2008)]%
        {sumedha2008unsupervised}
\bibfield{author}{\bibinfo{person}{M~Leone Sumedha} {and}
  \bibinfo{person}{Martin Weigt}.} \bibinfo{year}{2008}\natexlab{}.
\newblock \showarticletitle{Unsupervised and semi-supervised clustering by
  message passing: soft-constraint affinity propagation}.
\newblock \bibinfo{journal}{\emph{The European Physical Journal B}}
  \bibinfo{volume}{66} (\bibinfo{year}{2008}), \bibinfo{pages}{125--135}.
\newblock


\bibitem[Sun et~al\mbox{.}(2020)]%
        {FW2_26}
\bibfield{author}{\bibinfo{person}{Chengjin Sun}, \bibinfo{person}{Sizhe Chen},
  {and} \bibinfo{person}{Xiaolin Huang}.} \bibinfo{year}{2020}\natexlab{}.
\newblock \showarticletitle{Double backpropagation for training autoencoders
  against adversarial attack}.
\newblock \bibinfo{journal}{\emph{arXiv preprint arXiv:2003.01895}}
  (\bibinfo{year}{2020}).
\newblock


\bibitem[Sun et~al\mbox{.}(2021)]%
        {FW1_15}
\bibfield{author}{\bibinfo{person}{Hui Sun}, \bibinfo{person}{Tianqing Zhu},
  \bibinfo{person}{Zhiqiu Zhang}, \bibinfo{person}{Dawei~Jin Xiong},
  \bibinfo{person}{Wanlei Zhou}, {et~al\mbox{.}}}
  \bibinfo{year}{2021}\natexlab{}.
\newblock \showarticletitle{Adversarial Attacks Against Deep Generative Models
  on Data: A Survey}.
\newblock \bibinfo{journal}{\emph{arXiv preprint arXiv:2112.00247}}
  (\bibinfo{year}{2021}).
\newblock


\bibitem[Tabacof et~al\mbox{.}(2016)]%
        {BW1_2}
\bibfield{author}{\bibinfo{person}{Pedro Tabacof}, \bibinfo{person}{Julia
  Tavares}, {and} \bibinfo{person}{Eduardo Valle}.}
  \bibinfo{year}{2016}\natexlab{}.
\newblock \showarticletitle{Adversarial images for variational autoencoders}.
\newblock \bibinfo{journal}{\emph{arXiv preprint arXiv:1612.00155}}
  (\bibinfo{year}{2016}).
\newblock


\bibitem[Tantipongpipat et~al\mbox{.}(2021)]%
        {FW1_51}
\bibfield{author}{\bibinfo{person}{Uthaipon~Tao Tantipongpipat},
  \bibinfo{person}{Chris Waites}, \bibinfo{person}{Digvijay Boob},
  \bibinfo{person}{Amaresh~Ankit Siva}, {and} \bibinfo{person}{Rachel
  Cummings}.} \bibinfo{year}{2021}\natexlab{}.
\newblock \showarticletitle{Differentially private synthetic mixed-type data
  generation for unsupervised learning}. In \bibinfo{booktitle}{\emph{2021 12th
  International Conference on Information, Intelligence, Systems \&
  Applications (IISA)}}. IEEE, \bibinfo{pages}{1--9}.
\newblock


\bibitem[Tinsley et~al\mbox{.}(2021)]%
        {FW2_29}
\bibfield{author}{\bibinfo{person}{Patrick Tinsley}, \bibinfo{person}{Adam
  Czajka}, {and} \bibinfo{person}{Patrick Flynn}.}
  \bibinfo{year}{2021}\natexlab{}.
\newblock \showarticletitle{This face does not exist... but it might be yours!
  identity leakage in generative models}. In
  \bibinfo{booktitle}{\emph{Proceedings of the IEEE/CVF Winter Conference on
  Applications of Computer Vision}}. \bibinfo{pages}{1320--1328}.
\newblock


\bibitem[Torfi et~al\mbox{.}(2022)]%
        {FW2_30}
\bibfield{author}{\bibinfo{person}{Amirsina Torfi}, \bibinfo{person}{Edward~A
  Fox}, {and} \bibinfo{person}{Chandan~K Reddy}.}
  \bibinfo{year}{2022}\natexlab{}.
\newblock \showarticletitle{Differentially private synthetic medical data
  generation using convolutional gans}.
\newblock \bibinfo{journal}{\emph{Information Sciences}}  \bibinfo{volume}{586}
  (\bibinfo{year}{2022}), \bibinfo{pages}{485--500}.
\newblock


\bibitem[Tram{\`e}r et~al\mbox{.}(2017)]%
        {tramer2017ensemble}
\bibfield{author}{\bibinfo{person}{Florian Tram{\`e}r}, \bibinfo{person}{Alexey
  Kurakin}, \bibinfo{person}{Nicolas Papernot}, \bibinfo{person}{Ian
  Goodfellow}, \bibinfo{person}{Dan Boneh}, {and} \bibinfo{person}{Patrick
  McDaniel}.} \bibinfo{year}{2017}\natexlab{}.
\newblock \showarticletitle{Ensemble adversarial training: Attacks and
  defenses}.
\newblock \bibinfo{journal}{\emph{arXiv preprint arXiv:1705.07204}}
  (\bibinfo{year}{2017}).
\newblock


\bibitem[Walker and Myrick(2006)]%
        {walker2006grounded}
\bibfield{author}{\bibinfo{person}{Diane Walker} {and}
  \bibinfo{person}{Florence Myrick}.} \bibinfo{year}{2006}\natexlab{}.
\newblock \showarticletitle{Grounded theory: An exploration of process and
  procedure}.
\newblock \bibinfo{journal}{\emph{Qualitative health research}}
  \bibinfo{volume}{16}, \bibinfo{number}{4} (\bibinfo{year}{2006}),
  \bibinfo{pages}{547--559}.
\newblock


\bibitem[Wang et~al\mbox{.}(2017)]%
        {wang2017generative}
\bibfield{author}{\bibinfo{person}{Kunfeng Wang}, \bibinfo{person}{Chao Gou},
  \bibinfo{person}{Yanjie Duan}, \bibinfo{person}{Yilun Lin},
  \bibinfo{person}{Xinhu Zheng}, {and} \bibinfo{person}{Fei-Yue Wang}.}
  \bibinfo{year}{2017}\natexlab{}.
\newblock \showarticletitle{Generative adversarial networks: introduction and
  outlook}.
\newblock \bibinfo{journal}{\emph{IEEE/CAA Journal of Automatica Sinica}}
  \bibinfo{volume}{4}, \bibinfo{number}{4} (\bibinfo{year}{2017}),
  \bibinfo{pages}{588--598}.
\newblock


\bibitem[Warrens(2015)]%
        {warrens2015five}
\bibfield{author}{\bibinfo{person}{Matthijs~J Warrens}.}
  \bibinfo{year}{2015}\natexlab{}.
\newblock \showarticletitle{Five ways to look at Cohen's kappa}.
\newblock \bibinfo{journal}{\emph{Journal of Psychology \& Psychotherapy}}
  \bibinfo{volume}{5}, \bibinfo{number}{4} (\bibinfo{year}{2015}),
  \bibinfo{pages}{1}.
\newblock


\bibitem[Willetts et~al\mbox{.}(2019)]%
        {FW1_45}
\bibfield{author}{\bibinfo{person}{Matthew Willetts},
  \bibinfo{person}{Alexander Camuto}, \bibinfo{person}{Tom Rainforth},
  \bibinfo{person}{Stephen Roberts}, {and} \bibinfo{person}{Chris Holmes}.}
  \bibinfo{year}{2019}\natexlab{}.
\newblock \showarticletitle{Improving VAEs' Robustness to Adversarial Attack}.
\newblock \bibinfo{journal}{\emph{arXiv preprint arXiv:1906.00230}}
  (\bibinfo{year}{2019}).
\newblock


\bibitem[Wu and He(2021)]%
        {FW1_47}
\bibfield{author}{\bibinfo{person}{Jun Wu} {and} \bibinfo{person}{Jingrui He}.}
  \bibinfo{year}{2021}\natexlab{}.
\newblock \showarticletitle{Indirect Invisible Poisoning Attacks on Domain
  Adaptation}. In \bibinfo{booktitle}{\emph{Proceedings of the 27th ACM SIGKDD
  Conference on Knowledge Discovery \& Data Mining}}.
  \bibinfo{pages}{1852--1862}.
\newblock


\bibitem[Xie et~al\mbox{.}(2018)]%
        {BW1_4}
\bibfield{author}{\bibinfo{person}{Liyang Xie}, \bibinfo{person}{Kaixiang Lin},
  \bibinfo{person}{Shu Wang}, \bibinfo{person}{Fei Wang}, {and}
  \bibinfo{person}{Jiayu Zhou}.} \bibinfo{year}{2018}\natexlab{}.
\newblock \showarticletitle{Differentially private generative adversarial
  network}.
\newblock \bibinfo{journal}{\emph{arXiv preprint arXiv:1802.06739}}
  (\bibinfo{year}{2018}).
\newblock


\bibitem[Xu et~al\mbox{.}(2019)]%
        {BW1_14}
\bibfield{author}{\bibinfo{person}{Chugui Xu}, \bibinfo{person}{Ju Ren},
  \bibinfo{person}{Deyu Zhang}, \bibinfo{person}{Yaoxue Zhang},
  \bibinfo{person}{Zhan Qin}, {and} \bibinfo{person}{Kui Ren}.}
  \bibinfo{year}{2019}\natexlab{}.
\newblock \showarticletitle{GANobfuscator: Mitigating information leakage under
  GAN via differential privacy}.
\newblock \bibinfo{journal}{\emph{IEEE Transactions on Information Forensics
  and Security}} \bibinfo{volume}{14}, \bibinfo{number}{9}
  (\bibinfo{year}{2019}), \bibinfo{pages}{2358--2371}.
\newblock


\bibitem[Xu et~al\mbox{.}(2017)]%
        {xu2017feature}
\bibfield{author}{\bibinfo{person}{Weilin Xu}, \bibinfo{person}{David Evans},
  {and} \bibinfo{person}{Yanjun Qi}.} \bibinfo{year}{2017}\natexlab{}.
\newblock \showarticletitle{Feature squeezing: Detecting adversarial examples
  in deep neural networks}.
\newblock \bibinfo{journal}{\emph{arXiv preprint arXiv:1704.01155}}
  (\bibinfo{year}{2017}).
\newblock


\bibitem[Yang et~al\mbox{.}(2020)]%
        {FW1_10}
\bibfield{author}{\bibinfo{person}{Xu Yang}, \bibinfo{person}{Cheng Deng},
  \bibinfo{person}{Kun Wei}, \bibinfo{person}{Junchi Yan}, {and}
  \bibinfo{person}{Wei Liu}.} \bibinfo{year}{2020}\natexlab{}.
\newblock \showarticletitle{Adversarial learning for robust deep clustering}.
\newblock \bibinfo{journal}{\emph{Advances in Neural Information Processing
  Systems}}  \bibinfo{volume}{33} (\bibinfo{year}{2020}),
  \bibinfo{pages}{9098--9108}.
\newblock


\bibitem[Yao et~al\mbox{.}(2017)]%
        {yao2017investigation}
\bibfield{author}{\bibinfo{person}{YC Yao}, \bibinfo{person}{L Song}, {and}
  \bibinfo{person}{E Chi}.} \bibinfo{year}{2017}\natexlab{}.
\newblock \showarticletitle{Investigation on distributed K-means clustering
  algorithm of homomorphic encryption}.
\newblock \bibinfo{journal}{\emph{Comput. Technol. Develop.}}
  \bibinfo{volume}{2} (\bibinfo{year}{2017}), \bibinfo{pages}{81--85}.
\newblock


\bibitem[Yue et~al\mbox{.}(2021)]%
        {FW2_37}
\bibfield{author}{\bibinfo{person}{Jiutao Yue}, \bibinfo{person}{Haofeng Li},
  \bibinfo{person}{Pengxu Wei}, \bibinfo{person}{Guanbin Li}, {and}
  \bibinfo{person}{Liang Lin}.} \bibinfo{year}{2021}\natexlab{}.
\newblock \showarticletitle{Robust real-world image super-resolution against
  adversarial attacks}. In \bibinfo{booktitle}{\emph{Proceedings of the 29th
  ACM International Conference on Multimedia}}. \bibinfo{pages}{5148--5157}.
\newblock


\bibitem[Zhang et~al\mbox{.}(2019)]%
        {24}
\bibfield{author}{\bibinfo{person}{Jiale Zhang}, \bibinfo{person}{Junjun Chen},
  \bibinfo{person}{Di Wu}, \bibinfo{person}{Bing Chen}, {and}
  \bibinfo{person}{Shui Yu}.} \bibinfo{year}{2019}\natexlab{}.
\newblock \showarticletitle{Poisoning attack in federated learning using
  generative adversarial nets}. In \bibinfo{booktitle}{\emph{2019 18th IEEE
  International Conference On Trust, Security And Privacy In Computing And
  Communications/13th IEEE International Conference On Big Data Science And
  Engineering (TrustCom/BigDataSE)}}. IEEE, \bibinfo{pages}{374--380}.
\newblock


\bibitem[Zhang et~al\mbox{.}(2020)]%
        {zhang2020privacy}
\bibfield{author}{\bibinfo{person}{Jiliang Zhang}, \bibinfo{person}{Chen Li},
  \bibinfo{person}{Jing Ye}, {and} \bibinfo{person}{Gang Qu}.}
  \bibinfo{year}{2020}\natexlab{}.
\newblock \showarticletitle{Privacy threats and protection in machine
  learning}. In \bibinfo{booktitle}{\emph{Proceedings of the 2020 on Great
  Lakes Symposium on VLSI}}. \bibinfo{pages}{531--536}.
\newblock


\bibitem[Zhang et~al\mbox{.}(2021)]%
        {FW1_17}
\bibfield{author}{\bibinfo{person}{Longling Zhang}, \bibinfo{person}{Bochen
  Shen}, \bibinfo{person}{Ahmed Barnawi}, \bibinfo{person}{Shan Xi},
  \bibinfo{person}{Neeraj Kumar}, {and} \bibinfo{person}{Yi Wu}.}
  \bibinfo{year}{2021}\natexlab{}.
\newblock \showarticletitle{FedDPGAN: federated differentially private
  generative adversarial networks framework for the detection of COVID-19
  pneumonia}.
\newblock \bibinfo{journal}{\emph{Information Systems Frontiers}}
  \bibinfo{volume}{23}, \bibinfo{number}{6} (\bibinfo{year}{2021}),
  \bibinfo{pages}{1403--1415}.
\newblock


\bibitem[Zhang et~al\mbox{.}(2015)]%
        {zhang2015privacy}
\bibfield{author}{\bibinfo{person}{Qingchen Zhang}, \bibinfo{person}{Laurence~T
  Yang}, {and} \bibinfo{person}{Zhikui Chen}.} \bibinfo{year}{2015}\natexlab{}.
\newblock \showarticletitle{Privacy preserving deep computation model on cloud
  for big data feature learning}.
\newblock \bibinfo{journal}{\emph{IEEE Trans. Comput.}} \bibinfo{volume}{65},
  \bibinfo{number}{5} (\bibinfo{year}{2015}), \bibinfo{pages}{1351--1362}.
\newblock


\bibitem[Zhang et~al\mbox{.}(2018)]%
        {BW1_5}
\bibfield{author}{\bibinfo{person}{Xinyang Zhang}, \bibinfo{person}{Shouling
  Ji}, {and} \bibinfo{person}{Ting Wang}.} \bibinfo{year}{2018}\natexlab{}.
\newblock \showarticletitle{Differentially private releasing via deep
  generative model (technical report)}.
\newblock \bibinfo{journal}{\emph{arXiv preprint arXiv:1801.01594}}
  (\bibinfo{year}{2018}).
\newblock


\bibitem[Zhou et~al\mbox{.}(2021)]%
        {FW1_54}
\bibfield{author}{\bibinfo{person}{Junhao Zhou}, \bibinfo{person}{Yufei Chen},
  \bibinfo{person}{Chao Shen}, {and} \bibinfo{person}{Yang Zhang}.}
  \bibinfo{year}{2021}\natexlab{}.
\newblock \showarticletitle{Property Inference Attacks Against GANs}.
\newblock \bibinfo{journal}{\emph{arXiv preprint arXiv:2111.07608}}
  (\bibinfo{year}{2021}).
\newblock


\end{thebibliography}

\newpage
\appendix

\section{SLR details}
\label{sec:slr_details}
The questions for the QA:
\begin{itemize}
    \item Are the objectives, research questions, and hypotheses (if applicable), clear and relevant? 
    \item Are the case and its units of analysis well defined? 
    \item Is the suitability of the case to address the research questions clearly motivated? 
    \item Are the data collection procedures sufficient for the purpose of the case study (data sources, collection, validation)? 
    \item Is sufficient raw data presented to provide understanding of the case and the analysis? 
    \item Are the analysis procedures sufficient for the purpose of the case study (repeatable, transparent)? 
    \item Is a clear chain of evidence established from observations to conclusions?  
    \item Are threats to validity analyses conducted in a systematic way, and are countermeasures taken to reduce threats?  
    \item Is triangulation applied (multiple collection and analysis methods, multiple authors, multiple theories)?  
    \item Are ethical issues properly addressed (personal intentions, integrity, confidentiality, consent, review board approval)?  
    \item Are conclusions, implications for practice, and future research, suitably reported for its audience? 
\end{itemize}

The full name for the abbreviated journals and conferences.
\begin{itemize}
    \item ACM workshop on Artificial intelligence and security (AISec)
    \item Circulation
    \item IEEE Communications Letters (IEEE COMML)
    \item IEEE/CVF Conference on Computer Vision and Pattern Recognition (CVPR)
    \item International Conference on Artificial Intelligence and Statistics (AISTATS)
    \item IEEE International Conference on Image Processing (IEEE ICIP)
    \item IEEE International Conference On Trust, Security And Privacy In Computing and Communications (TrustCom)
    \item AAAI Workshop on Privacy-Preserving Artificial Intelligence (PPAI)
    \item ACM Transactions on Internet Technology	(TOIT)
    \item Computers \& Security
    \item Frontiers in Artificial Intelligence, (Front. Artif. Intell.)
    \item EAI SecureComm
    \item IEEE TRANSACTIONS ON INFORMATION FORENSICS AND SECURITY (TIFS)
    \item Privacy Enhancing Technologies (PoPETs)
    \item IEEE International Conference on Informatics, IoT, and Enabling Technologies (ICIoT)
    \item IEEE International Conference on Information Reuse and Integration for Data Science (IEEE IRI)
    \item IEEE Conference on Network Function Virtualization and Software Defined Networks (IEEE NFV-SDN)
    \item IEEE international conference on fuzzy systems (FUZZ-IEEE)
    \item ACM Internet Measurement Conference (ACM IMC)
    \item ACM SIGKDD Conference on Knowledge  Discovery and Data Mining (ACM SIGKDD)
    \item International Conference on learning representations (ICLR)
    \item International Journal of Computer Vision (IJCV)
    \item Pattern Recognition (Pattern Recognit)
    \item IEEE Symposium on Security and Privacy Workshops (IEEE SPW)
    \item Structural, Syntactic, and Statistical Pattern Recognition (S+SSPR)
    \item ACM SIGSAC conference on computer and communications security (CCS)
    \item Conference on Neural Information Processing Systems (NeurIPS)
    \item ACM international conference on Multimedia (ACM MM)
    \item Electronics
    \item Information Systems Frontiers (Inf Syst Front)
    \item International Journal of Cognitive Computing in Engineering (JCCE)
\end{itemize}
\label{list:journal_conference_full_names}

\section{Code descriptions}

\subsection{Target types}
\label{sec:description_target_types}
\begin{multicols}{2}
    \begin{itemize}
        \item \textit{GAN}: ML consisting of two models, a generator, and a discriminator, training by competing against each other
        \item \textit{AutoEncoder}: Neural network which is trained on the input and output being the same value 
        \item \textit{Clustering}: Methods using distance functions to determine the size of clusters of the same groups
        \item \textit{Classification}: Here, classification methods are models which have the output from a UL system as its input
        \item \textit{Super Resolution}: ML method for increasing the dimensionality of a sample
    \end{itemize}
\end{multicols}

\subsection{Attack types}
\label{sec:description_attack_types}
\begin{multicols}{2}
    \begin{itemize}
        \item \textit{Adversarial Examples}: Specific perturbation in input to cause disruption
        \item \textit{Poisoning}: Modification or addition of specifically crafted training samples
        \item \textit{Membership Inference}: Extraction of knowledge about training by using the output of a model
        \item \textit{Evasion}: Attacks crafted to avoid detection techniques
        \item \textit{Obfuscation}: Modification of input to be hidden in existing groups or clusters
        \item \textit{Bridging}: Addition of crafted samples to bridge between groups or clusters, causing disruption
        \item \textit{Saturation}: Flooding the system with an excess of data
        \item \textit{Model Extraction}: Extraction of the model functions and parameters
        \item \textit{Backdoor}: Inclusion of a specific trigger in a model causing disruption
        \item \textit{Attribute Inference}: Extraction of attributes of training data
        \item \textit{Reconstruction}: Extraction of statistical features of the training set for a model
        \item \textit{Accuracy Extraction}: Extraction of a model through comparison to generated samples
        \item \textit{Fidelity Extraction}: Extraction of a model through comparison to training samples
    \end{itemize}
\end{multicols}

\subsection{Metrics}
\label{sec:description_metrics}
\begin{multicols}{2}
\begin{itemize}
    \item \textit{Accuracy/ F1-Score}: Measurement of how many samples were correctly identified
    \item \textit{L2}: Measurement of perturbation, by taking the root of all squares of perturbations in the data
    \item \textit{AUCROC}: Measurement of how well a model distinguishes samples, by measuring true and false positive rates for all thresholds
    \item \textit{L-Infinity}: Measurement of perturbation, by taking the largest values among all perturbations
    \item \textit{False Positive}: The number of values measured as positive, while being negative
    \item \textit{BLER}: The rate of erroneous blocks out of all blocks 
    \item \textit{True Positive}: The number of values measured as positive, while being positive
    \item \textit{Misclassification Score}: Measurement of how many classifications are not correct
    \item \textit{L1}: Measurement of perturbation, by taking the sum of all perturbations
    \item \textit{Precision}: A measurement of the rate of true positives, out of all measured positives
    \item \textit{SSIM}: A quantification of the perceived quality of an image
    \item \textit{AUCPRC}: Measurement of how well a model performs, by measuring precision and recall over all thresholds
    \item \textit{Recall}: A measurement of the rate of true positives, out of all measured true values
    \item \textit{True Negative}: The number of values measured as negative, while being negative
    \item \textit{False Negative}: The number of values measured as negative, while being positive
    \item \textit{Average Precision}: Average value of the precision score for all thresholds
    \item \textit{RMSE}: The Root Mean Square Error is a measurement of average error in measurement
    \item \textit{PSNR}: Peak Signal to Noise Ratio measures the amount of noise relative to real data is in a sample
    \item \textit{MSSSIM}: Multi Scale Structural Similarity Index Measure uses SSIM on different scales
    \item \textit{LPIPS}: Learned Perceptual Image Patch Similarity measures similarities in features between generated and original samples
    \item \textit{Jensen-Shannon Divergence}: A measurement for determining the similarities between probability distributions
    \item \textit{SRMSE}: Standardised RMSE 
    \item \textit{Reconstruction Distance}: A measurement for the distance between the original and generated sample
    \item \textit{Total Variation Distance}: Measurement of the distance between variations of different distributions
    \item \textit{Attack Loss}: Measurement of the distance between the latent space distribution of samples with small distortions
\end{itemize}
\end{multicols}

\subsection{Defenses}
\label{sec:description_defenses}
\begin{itemize}
    \item \textit{Differential Privacy: } Using mathematical proofs in training to force the constraint that a single sample cannot be known to be used for training
    \item \textit{Adversarial Training: } The method of using an adversarial method for constructing samples the model otherwise would be vulnerable to, and subsequently training using these samples
    \item \textit{Gradient Masking: } A method of obfuscating the gradients of the output of a model, to prevent useful information from being obtained
    \item \textit{Game Theory: } The employment of defensive methods according to a strategy using a multitude of parameters
    \item \textit{Differentially Private Stochastic Gradient Descent (DPSGD): } The application of DP to the training of GANs
    \item \textit{Weight Normalization: } Reparametrization of the weight vectors in an already existing NN
    \item \textit{Bootstrapping: } The use of existing robust models as the base of training a model
    \item \textit{Fine Pruning: } The removal of neurons in a NN which does not improve the performance
    \item \textit{Dropout: } Random removal of neurons during training
\end{itemize}

\end{document}